\documentclass[11pt, letterpaper]{article}

\usepackage[margin=1in]{geometry}
\usepackage[utf8]{inputenc} % allow utf-8 input
\usepackage[T1]{fontenc}    % use 8-bit T1 fonts
\usepackage[colorlinks]{hyperref}       % hyperlinks
\usepackage{url}            % simple URL typesetting
\usepackage{booktabs}       % professional-quality tables
\usepackage{array}          % to wrap columns in tables
\usepackage{amsfonts}       % blackboard math symbols
\usepackage{nicefrac}       % compact symbols for 1/2, etc.
\usepackage{microtype}      % microtypography
\usepackage{xcolor}          % colors
\usepackage[inline]{enumitem}
\newlist{inlinelist}{enumerate*}{1}
\setlist[inlinelist]{label=\bfseries(\alph*), itemjoin={{, }}, itemjoin*={{, and }}} % <-------------
\setlist[enumerate]{label=\bfseries\arabic*.}
\usepackage[pdftex]{graphicx}
\usepackage{float}
\usepackage{amsmath}
\usepackage{amssymb}
\usepackage{subcaption}
\usepackage{adjustbox}
\usepackage{pifont}
\usepackage{siunitx}
\usepackage{csquotes}
\usepackage{authblk}
\usepackage{lscape}

\usepackage[backend=biber,style=ieee,natbib=true,sorting=nyt,maxnames=2,minnames=1,maxbibnames=7]{biblatex} 
\usepackage[linesnumbered,ruled,vlined]{algorithm2e}
\newcommand{\algvar}[1]{\emph{#1}}

\usepackage{amsthm}
\theoremstyle{definition}

\def\tmp#1#2#3{%
  \definecolor{Hy#1color}{#2}{#3}%
  \hypersetup{#1color=Hy#1color}}
\tmp{link}{HTML}{800006}
\tmp{cite}{HTML}{2E7E2A}
\tmp{file}{HTML}{131877}
\tmp{url} {HTML}{8A0087}
\tmp{menu}{HTML}{727500}
\tmp{run} {HTML}{137776}
\def\tmp#1#2{%
  \colorlet{Hy#1bordercolor}{Hy#1color#2}%
  \hypersetup{#1bordercolor=Hy#1bordercolor}}
\tmp{link}{!60!white}
\tmp{cite}{!60!white}
\tmp{file}{!60!white}
\tmp{url} {!60!white}
\tmp{menu}{!60!white}
\tmp{run} {!60!white}

\def\eqref#1{equation~\ref{#1}}
\def\1{\bm{1}}

\def\rs{{\textnormal{s}}}

\def\rvu{{\mathbf{i}}}

\def\rvm{{\mathbf{m}}}
\def\rvn{{\mathbf{n}}}

\def\rvs{{\mathbf{s}}}
\def\rvt{{\mathbf{t}}}
\def\rvu{{\mathbf{u}}}

\DeclareMathAlphabet{\mathsfit}{\encodingdefault}{\sfdefault}{m}{sl}
\SetMathAlphabet{\mathsfit}{bold}{\encodingdefault}{\sfdefault}{bx}{n}

\def\sR{{\mathbb{R}}}

\def\sZ{{\mathbb{Z}}}

\usepackage{multirow}

\title{On Using Quasirandom Sequences in Machine Learning for Model Weight Initialization}
\author[]{Andriy Miranskyy}
\author[]{Adam Sorrenti}
\author[]{Viral Thakar\thanks{Authors are listed alphabetically.}}

\affil[]{Department of Computer Science, Toronto Metropolitan University, Toronto, Canada}
\affil[]{{\{avm, adam.sorrenti, vthakar\}@torontomu.ca}}
\date{}

\newcommand{\distr}[2]{\mathcal{#1}_{#2}}
\newcommand{\uniform}[1]{\distr{U}{#1}}
\newcommand{\normal}[1]{\distr{N}{#1}}
\newcommand{\tnormal}[1]{\distr{T}{#1}}
\newcommand{\prng}{\mathcal{P}}
\newcommand{\qrng}{\mathcal{Q}}
\newcommand{\mean}[1]{\bar{#1}}
\newcommand{\Nin}{\mathbf{N}_{\text{in}}}
\newcommand{\Nout}{\mathbf{N}_{\text{out}}}
\newcommand{\mask}[1]{\textsc{#1}}

\begin{document}

\maketitle

\begin{abstract}

The effectiveness of training neural networks directly impacts computational costs, resource allocation, and model development timelines in machine learning applications. An optimizer's ability to train the model adequately (in terms of trained model performance) depends on the model's initial weights. Model weight initialization schemes use pseudorandom number generators (PRNGs) as a source of randomness.

We investigate whether substituting PRNGs for low-discrepancy quasirandom number generators (QRNGs)~--- namely Sobol' sequences~--- as a source of randomness for initializers can improve model performance.
We examine Multi-Layer Perceptrons (MLP), Convolutional Neural Networks (CNN), Long Short-Term Memory (LSTM), and Transformer architectures trained on MNIST, CIFAR-10, and IMDB datasets using SGD and Adam optimizers. Our analysis uses ten initialization schemes: Glorot, He, Lecun (both Uniform and Normal); Orthogonal, Random Normal, Truncated Normal, and Random Uniform. Models with weights set using PRNG- and QRNG-based initializers are compared pairwise for each combination of dataset, architecture, optimizer, and initialization scheme.

Our findings indicate that QRNG-based neural network initializers either reach a higher accuracy or achieve the same accuracy more quickly than PRNG-based initializers in 60\% of the 120 experiments conducted.
Thus, using QRNG-based initializers instead of PRNG-based initializers can speed up and improve model training.

\end{abstract}

\section{Introduction}\label{sec:intro}
The effectiveness of training deep neural networks has a direct impact on computational costs, resource allocation, and model development timelines in machine learning applications~\cite{hanin2018start, prince2023understanding}. The initialization of the neural network's weights plays a critical role in its training efficiency: random initialization methods can introduce variations that slow the training or hinder convergence~\cite{ramasinghe2023much, hanin2018start, skorski2021revisiting}. Pseudorandom number generators (PRNGs) are traditionally used to initialize neural networks~\cite{hopfield1982neural,Goodfellow-et-al-2016,keras_initializers}. However, there are other ways to generate sequences of random numbers. For example, low-discrepancy quasirandom number generators (QRNGs), which provide high uniformity in filling high-dimensional spaces, have proved effective in numerical integration in high-dimensional spaces via Monte Carlo simulations~\cite{jackel2002monte}.

QRNGs are also used for optimization~\cite{kucherenko2005application}. Compared with PRNGs, low-discrepancy QRNGs ensure a more uniform exploration of the search space. Uniformity allows for a thorough examination of the parameter space, potentially resulting in more reliable results for functions with multiple local optima. This leads to the following question.

Can QRNG uniformity properties be applied to neural network weight initialization? Our \textbf{hypothesis} is that QRNG-based neural network weight initializers can accelerate neural network training. The following research questions will help us explore various aspects of this hypothesis.

\begin{enumerate}[label=\textbf{RQ\arabic*.}]
    \item What effect does the selection of optimization algorithms have on deep neural networks' performance when initialized using the QRNG-based weight initialization scheme?

    \item How does a QRNG-based weight initialization scheme affect deep neural networks' performance?

    \item How does the choice of deep neural network architecture affect the model's performance when it is initialized using the QRNG-based weight initialization scheme?

    \item How does the change in the dataset affect the performance of deep neural networks initialized with the QRNG-based weight initialization scheme?
\end{enumerate}

We empirically investigate how a specific type of QRNG, namely Sobol' sequences~\cite{sobol1967distribution}, affects the training of four neural network architectures: Multi-Layer Perceptrons (MLP)~\cite{Goodfellow-et-al-2016}, Convolutional Neural Networks (CNN)~\cite{Goodfellow-et-al-2016}, Long Short-Term Memory (LSTM)~\cite{hochreiter1997long}, and Transformer~\cite{vaswani2017attention}. Models' kernel weights are initialized using ten different initializers: Glorot Uniform~\cite{glorot2010understanding}, Glorot Normal~\cite{glorot2010understanding}, He Uniform~\cite{he2015delving}, He Normal~\cite{he2015delving}, Lecun Uniform~\cite{klambauer2017self}, Lecun Normal~\cite{klambauer2017self}, Orthogonal~\cite{saxe2014exact}, Random Uniform~\cite{keras_initializers}, Random Normal~\cite{keras_initializers}, and Truncated Normal~\cite{keras_initializers}. Models are trained on two types of data:
\begin{inlinelist}
    \item images, represented by Modified National Institute of Standards and Technology (MNIST)~\cite{lecun1998gradient} and Canadian Institute For Advanced Research, 10 classes (CIFAR-10)~\cite{krizhevsky2009learning} datasets
    \item natural language, represented by Internet Movie Database (IMDB)~\cite{maas-EtAl:2011:ACL-HLT2011} dataset. 
The stochastic gradient descent (SGD)~\cite{rosenblatt1958perceptron,bilmes1997using} and Adam~\cite{kingma2015adam} methods optimize the models' weights.
\end{inlinelist}

\paragraph{Summary of contributions} We investigate whether QRNG-based initializers can accelerate training in four neural network architectures. Among our contributions are: 
\begin{inlinelist}
    \item developing QRNG-based sampling mechanisms to implement ten popular initializers
    \item providing a Keras-based implementation of these initializers\footnote{Shared via  \url{https://github.com/miranska/qrng-init}~\cite{source_code}.}
    \item evaluating QRNG's impact on training speed across different datasets and optimization methods.
\end{inlinelist}
Our research confirms the hypothesis by addressing the research questions. 
We showed that QRNG-based initializers improve accuracy or accelerate neural network training in 60\% of the 120 experiments conducted. The top 25\% of the improved maximum median accuracies fall between $\approx 0.0775$ and $0.3550$. The negative side effects of using QRNG are minimal, with the top 25\% of observed losses leading to a decrease in accuracy between $\approx 0.0031$ and $0.0402$. 
Data complexity may contribute to variability of results. 
QRNG significantly improved nine of the ten initializers (except Random Uniform). A generalizability assessment requires further research.

The remainder of the paper is organized as follows. Section~\ref{sec:methods} discusses the mechanisms for implementing initializers. Section~\ref{sec:results} evaluates the initializers empirically, presents results, discusses limitations, and poses open questions. Section~\ref{sec:related-work} reviews related work. Finally, Section~\ref{sec:summary} concludes the paper.

\section{Methods}\label{sec:methods}
Below we provide details of the setup needed to compare the performance of PRNGs and QRNGs for weight initialization of neural networks.

\paragraph{Random number generators} 
Pseudorandom numbers can be generated using various algorithms. As a baseline, we use the Philox PRNG~\cite{salmon2011parallel} (as it is the default PRNG in Keras with Tensorflow backend~\cite{tensorflow2015-whitepaper}, a popular machine learning library, see Appendix~\ref{sec:distr_prng} for details); see Appendix~\ref{sec:prng_uniform} for additional details. Furthermore, we will use the ubiquitous Mersenne Twister PRNG~\cite{makoto1998mersenne} to compare the performance of TensorFlow's Philox implementation.

Various QRNGs exist, e.g., those based on Faure~\cite{faure1982sequence}, Halton~\cite{halton1960efficiency}, Niederreiter~\cite{niederreiter1988low}, and Sobol'~\cite{sobol1967distribution} sequences. They provide a deterministic and evenly distributed set of points in multidimensional space, thereby overcoming some limitations of traditional pseudorandom sequences \cite{jackel2002monte}.  Empirical evidence suggests that Sobol' sequences are generally more homogeneous~\cite{jackel2002monte}. Therefore, we chose to use Sobol' sequences in our study. Further details about Sobol' sequences are given in Appendix~\ref{sec:sobol}.

\paragraph{Distributions}
Detailed information about the implementation of the distributions using PRNGs and QRNGs is provided in Appendix~\ref{sec:distributions}. We provide a brief summary below. 

Three distributions are used in the initializers under study: random uniform, random normal, and truncated normal. Keras implementations of uniform distribution with TensorFlow backend use the Philox PRNG; Box-Muller transforms and sampling with rejection are utilized for drawing from random normal and truncated normal distributions. Further details are provided in Appendix~\ref{sec:distr_prng}. 

In order to implement QRNG-based distributions, we use inverse transform sampling, taking values drawn from the Sobol' sequences as input (see Appendix~\ref{sec:distr_qrng} for details). For Mersenne Twister PRNG-based distributions, we follow a similar approach: doing inverse sampling of random sequences generated by the Mersenne Twister algorithm (see Appendix~\ref{sec:distr_prng_mt} for additional details).

\paragraph{Initializers} 
The ten initializers under study can be classified into three groups: those using random uniform, random normal, or truncated normal distributions. 
\begin{inlinelist}
    \item Random uniform distribution is used by Glorot Uniform, He Uniform, Lecun Uniform, and Random Uniform initializers 
    \item Normal distribution --- by Orthogonal and Random Normal initializers
    \item Truncated normal --- by Glorot Normal, He Normal, Lecun Normal, and Truncated Normal initializers. 
\end{inlinelist}
The initializer details are provided in the Appendix~\ref{sec:initializers}.

Initializers can also be categorized on the basis of how they handle random numbers. This categorization will be helpful during the analysis of the results. We create three groups: shape-agnostic, shape-dependent, and orthogonal.
A \textit{shape-agnostic} initializer, such as Random Normal, Random Uniform, and Truncated Normal, uses random numbers independent of the object's shape whose weights we are initializing. 
A \textit{shape-dependent} initializer\footnote{All shape-dependent initializers under study use variance scaling.}, such as Glorot Uniform, Glorot Normal, He Uniform, He Normal, Lecun Uniform, and Lecun Normal, transforms random values according to the parameters that govern the underlying distributions as well as the shapes of the input tensors. 
Finally, an \textit{orthogonal } initializer deserves its own category: while primarily dependent on the parameters governing the underlying distributions, it also performs a significant transformation of the data (namely QR decomposition), which has a dramatic impact on the underlying random sequences.

\section{Results}\label{sec:results}

\subsection{Experimental setup}\label{sec:exp_design}

\paragraph{Datasets}
Our models are trained on two image classification datasets, MNIST and CIFAR-10, and one text classification dataset, IMDB. The details of the datasets and data preparations are given in Appendix~\ref{sec:datasets}.

\paragraph{Deep Neural Network Architectures}
Our models utilize four fundamental deep neural network architectures: MLP, CNN, LSTM, and Transformer, with detailed configurations provided in Appendix~\ref{sec:models}. The architectures we use are intentionally simplified to see how different random number generators affect kernel weights (and avoid the effects of more sophisticated techniques). Consequently, we have omitted layers (such as extra regularization and dense layers) to minimize their influence on performance. We keep the number of kernel weights small, isolating the effects of random number generators (a larger number of weights gives any model better predictive power, as demonstrated in Appendix~\ref{sec:single_layer_study} (Figure~\ref{fig:1-layer-mlp-results}) and discussed in~\cite{zhang2017understanding}. This simplification will reduce overall model efficacy, but our primary objective is to assess the effects of random generators, not to optimize the performance for any particular architecture.

\paragraph{Random number generators} \label{sec:rng_under_study}
The QRNG under study is based on Sobol' sequences (see Section~\ref{sec:methods}) implemented in SciPy v.1.13.1 software~\cite{2020SciPy-NMeth,scipy_sobol}, which utilizes directional integers computed by~\cite{joe2008constructing}. More details are available in Appendix~\ref{sec:sobol}.

Keras v.3.3.3~\cite{chollet2015keras} with TensorFlow v.2.16.1~\cite{tensorflow2015-whitepaper} backend provides the baseline Philox PRNG~\cite{salmon2011parallel} (see Appendix~\ref{sec:distr_prng}), while NumPy library v.1.24~\cite{numpy2023mt} gives us Mersenne Twister PRNG~\cite{makoto1998mersenne}.

\paragraph{Distributions and Initializers}
We intialize models' kernel weights using ten different initializers listed in Section~\ref{sec:methods}.
Appendices~\ref{sec:initializers} and~\ref{sec:distributions} contain implementation details of initializers and their underlying distributions, respectively. Essentially, we modify the Keras~\cite{chollet2015keras} classes responsible for sampling random numbers and creating initializers. Inverse transform sampling is performed using functions from the SciPy v.1.13.1.

\paragraph{Optimizers}
Our models are trained using Adam and SGD algorithms. It allows us to examine how initialization methods and architectural choices are affected by the choice of optimizer. Appendix~\ref{sec:optimizers} lists the model hyperparameters.

\paragraph{Combinations and repetitions of experiments}
For the main experiments, we explore the following combination of models, optimizers, and datasets. With two image classification datasets, we explore two architectures (MLP and CNN) and two optimizers (Adam and SGD), resulting in eight experiments. Four experiments are conducted on the text classification dataset (IMDB) using two architectures (LSTM and Transformer) and two optimizers (Adam and SGD). The $12$ $(=8+4)$ combinations are tested against ten initializers and two random generators (PRNG and QRNG), bringing the number of combinations to $240$. Since optimization is stochastic, experiments are repeated \num{100} times to assess robustness.

\paragraph{Training}
For each individual experiment run, we adhered to the following settings. Every experiment should have the same epoch count and batch size for consistent comparisons. The models are trained for $30$ epochs, providing a substantial training duration for model convergence and performance assessment. A fixed batch size of $64$ was used, promoting efficient gradient updates and facilitating fair comparisons.

\paragraph{Testbed}\label{sec:testbed}
In order to eliminate the sources of randomness from GPU software stacks, we chose a CPU-only testbed. The experiments were performed on high-performance computing clusters with 
Intel Xeon Gold 6148 Skylake CPUs. 
All experiments were allocated 4~GB of memory and 2~CPU cores.

\subsection{Measure performance and analyze the results}\label{sec:performance_measures}

\subsubsection{Compute the central tendency and variability}\label{sec:measures}

To assess model performance, we evaluate prediction accuracy on a dataset's test segment after $i$ epochs of training. That is, we use an \textit{epoch-to-accuracy} metric to compare models across different experimental settings (described in Section~\ref{sec:testbed}).

Given the stochastic nature of optimization, 100 experimental repetitions can result in varying accuracy values, detailed in Appendix~\ref{sec:accuracy_figures}. Thus, we cannot use the epoch-to-accuracy metric directly. Instead, we will measure its central tendency and variability. To mitigate the influence of outliers and skewness in the data, we use the median rather than the mean to measure central tendency and the interquartile range\footnote{The IQR of a vector $x$ is computed as the difference between the third and first quartile of the data in $x$.} (IQR) rather than the standard deviation to measure variability.

Our analysis compares the performance of two models, each employing a PRNG- or QRNG-based source of randomness, across the same dataset, architecture, optimizer, and initializer algorithm. As discussed in Section~\ref{sec:exp_design}, we have $240$ experiments, resulting in $120$ pairs of models.

Specifically, we calculate the percentage of relative change in \textit{median epoch-to-accuracy} (i.e., the number of epochs required to reach a specific median accuracy threshold): 
\begin{equation}\label{eq:rel_epoch_diff} 
E(A) = \frac{E_{\qrng}(A) + \Delta_{\qrng} - E_{\prng}(A)}{E_{\prng}(A)} \times 100, 
\end{equation} 
where $E(A)$ is an efficiency metric, $E_{\qrng}(A)$ and $E_{\prng}(A)$ are the epoch numbers at which the QRNG- and PRNG-based versions of an initializer first reach or exceed the \textit{median accuracy} $A$, respectively, and $\Delta_{\qrng}=4$ represents additional epochs required so select a good starting seed for QRNG initializer (see Appendix~\ref{sec:qrng_seed_selection} for details). Using this formula, we can determine how much faster one model is compared to another in terms of the median epoch-to-accuracy.

Additionally, we assess the variation in accuracy between the two models by the difference of their IQRs:
\begin{equation}\label{eq:iqr_diff} 
    D(A) = D_{\qrng}(A) - D_{\prng}(A),
\end{equation}
where \(D_{\qrng}(A)\) and \(D_{\prng}(A)\) are the IQR of accuracy measurements for the QRNG-based and PRNG-based versions of the initializer at epochs \(E_{\qrng}(A)\) and \(E_{\prng}(A)\), respectively.

Further details on calculating our performance metrics, $E(A)$ and $D(A)$, are provided with numeric examples in Appendix~\ref{sec:example_performance_calculation}.

\subsubsection{Application of performance metrics}
In order to compare the performance of models, let us construct four simple classification rules denoted by $S_{(\cdot)}$. Note that the maximum accuracy values $A_{(\cdot)}^{\max}$ may differ with experimental configuration. We are not comparing the efficacy of different initializers across all configurations but rather examining a specific pair of models individually, as discussed in Section~\ref{sec:measures}. 

\paragraph{Compare maximum accuracy values}
To assess the correctness of prediction, we will use the values $A_{\prng}^{\max}$ and $A_{\qrng}^{\max}$, which represent the maximum median accuracy attained by the PRNG-based and QRNG-based models, respectively, for each combination of dataset, architecture, optimizer, and initializer algorithm.

Our first step is to determine which of the two models achieved the highest accuracy. To do this, we perform a non-parametric, non-paired Mann-Whitney $U$ test~\cite{mann1947test} implemented in R v.4.4.0~\cite{r_environment}. The null hypothesis is that the distributions of $\zeta$ and $\eta$ differ by a location shift of $0$. The one-sided alternative ``greater'' hypothesis is that $\zeta$ is shifted to the right of $\eta$, and the one-sided alternative ``less'' hypothesis is that $\zeta$ is shifted to the left of $\eta$.

In our study, $\zeta$ represents a distribution of 100 accuracy values achieved by the PRNG-based model at the earliest epoch when $A_\prng^{\max}$ was reached, denoted $E(A_\prng^{\max})$. Similarly, $\eta$ represents a distribution of 100 accuracy values achieved by the QRNG-based model at the earliest epoch when $A_\qrng^{\max}$ was reached, denoted $E(A_\qrng^{\max})$. We set the $p$-value for the $U$ test at $0.05$, resulting in the following classification rule:
\begin{equation}\label{eq:score_max_accuracy}
    S_A\left[A_{\prng}^{\max}, A_{\qrng}^{\max}\right] = 
\begin{cases}
    \text{win}, & \text{if $p$-value for the $U$ test with ``less'' alternative hypothesis} < 0.05 \\
    \text{loss}, & \text{if $p$-value for the $U$ test with ``greater'' alternative hypothesis} < 0.05 \\
    \text{tie}, & \text{otherwise}\\
\end{cases}.
\end{equation}

\paragraph{Compare efficiency and relative variability of achieving higher accuracy} 
We also need a simple but informative mechanism to aggregate the large amount of values returned by $E(A)$ and $D(A)$. Let us examine the maximum median epoch-to-accuracy values reached by various setups. 

To compare the ``speed'' with which one model achieves higher accuracy than another, we define:
\begin{equation}\label{eq:A_m}
A_m = \min\left(A_{\prng}^{\max}, A_{\qrng}^{\max}\right),    
\end{equation}
where $A_{\prng}^{\max}$ and $A_{\qrng}^{\max}$ represent the maximum accuracies attained by the PRNG-based and QRNG-based models, respectively, for each combination of dataset, architecture, optimizer, and initializer algorithm. The value $A_m$ reflects the highest accuracy achieved by the less effective model in each pair\footnote{The formulation of $A_m$ ensures that we do not observe the special case described in Appendix~\ref{sec:special_case}.}. $E(A_m)$ indicates the relative speed at which each model reaches $A_m$, while $D(A_m)$ measures the difference in variability at the epochs where each model reaches $A_m$.

The following are classification rules we use to summarize and compare the performance (in terms of the best accuracy achieved) where QRNG might outperform PRNG. For $E(A_m)$, we simply assess the count of epochs with the following conditions:
\begin{equation}\label{eq:score_epoch_diff}
    S_E[E(A_m)] = 
\begin{cases}
    \text{win}, & \text{if } E(A_m) < 0 \\
    \text{tie}, & \text{if } E(A_m) = 0\\
    \text{loss}, & \text{if } E(A_m) > 0\\
\end{cases}.
\end{equation}
For $D(A_m)$, we use the non-parametric Fligner-Killeen median test\footnote{The test is considered robust against deviations from normality~\cite{conover1981comparative}.}~\cite{fligner1971test} of homogeneity of variances (implemented in R v.4.4.0) to assess whether variability is similar. The null hypothesis is that the variances are the same in each group. We use the following classification rules since the test does not have a one-tailed version:
\begin{equation}\label{eq:score_iqr_diff}
S_D[D(A_m)] = 
\begin{cases}
    \text{win}, & \text{if the Fligner-Killeen test's $p$-value} < 0.05 \text{ and }  D(A_m) < -0.01\\
    \text{loss}, & \text{if the Fligner-Killeen test's $p$-value} < 0.05 \text{ and }  D(A_m) > 0.01\\
    \text{tie}, & \text{if the Fligner-Killeen test's $p$-value} < 0.05 \text{ and } -0.01 \le D(A_m) \le 0.01\\
    \text{tie}, &\text{if the Fligner-Killeen test's $p$-value} \ge 0.05\\
\end{cases}.
\end{equation}
From a practical perspective, a negligible difference is defined as an IQR difference of $\pm 0.01$, and these outcomes are classified as ties.
An example of computing $S_E[E(A_m)]$ and $S_D[D(A_m)]$ is given in Appendix~\ref{sec:example_performance_calculation}.

\paragraph{Joining the classification rules}
To conclude whether a PRNG- or QRNG-based model achieved better results in a given pair, we need to aggregate various permutations of the outputs returned by the three $S_{(\cdot)}$ rules. Table~\ref{tbl:final_outcome} presents the aggregation of permutations observed in our experiments. We refer to this aggregation as the ``final outcome''.  For brevity, we only show a subset of permutations of our experiments' outcomes.

\begin{table}[htb]
\caption{The rules for aggregating the results of the three $S_{(\cdot)}$ classification rules. The symbol~* denotes any outcome.}
\label{tbl:final_outcome}
\resizebox{\textwidth}{!}{%
\begin{tabular}{@{}lllllp{10cm}@{}}
\toprule
Final outcome                                     & $S_A$ & $S_E$ & $S_D$ & Mask        & Description                                                                                                                           \\ \midrule

\multirow{2}{*}{Loss} & loss                & *                        & *                           & \mask{l:l,*,*}  & QRNG-based model achieved a lower level of accuracy.                                                                                  \\
                      & tie                 & loss                     & *                           & \mask{l:t,l,*}  & QRNG-based model achieved comparable accuracy slower.                                                                                 \\
\midrule
\multirow{1}{*}{Tie}                       & tie                 & tie                      & tie                  & \mask{t:t,t,t} & QRNG-based model achieved comparable accuracy simultaneously (with similar variation).                                        \\ \midrule

\multirow{5}{*}{Win}                           & tie                 & win                      & win or tie                  & \mask{w:t,w,wt} & QRNG-based model achieved comparable accuracy faster (with less or similar variation).                                                \\
                                           & win                 & loss                      & *                        & \mask{w:w,l,*}  & QRNG-based model achieved a higher level of accuracy faster and reached $A_m$ values slower.                  \\
                                           & win                 & tie                      & win or tie                  & \mask{w:w,t,wt} & QRNG-based model achieved a higher level of accuracy faster and reached $A_m$ values simultaneously (with less or similar variation). \\
                                           & win                 & win                      & loss                        & \mask{w:w,w,l}  & QRNG-based model achieved a higher level of accuracy faster and reached $A_m$ values faster (with higher variation).                  \\
                                               & win                 & win                      & win or tie                  & \mask{w:w,w,wt} & QRNG-based model achieved a higher level of accuracy faster and reached $A_m$ values faster (with less or similar variation).         \\

\bottomrule
\end{tabular}%
}
\end{table}

\subsection{Answers to the Research Questions}
This section addresses the research questions and evaluates the hypotheses presented in Section~\ref{sec:intro}. As mentioned in Section~\ref{sec:exp_design},  we deliberately simplify the models to isolate the effect of random number generators, so the absolute accuracy of the models is far from best-in-class.  However, we are interested in the differences in performance between a pair of models rather than their absolute accuracy.

The raw data that were used to compute the $E(A)$ and $D(A)$ values for the $120$ model pairs are given in Appendix~\ref{sec:accuracy_figures}. Appendix~\ref{sec:summary_stats} provides a summary of the values of $A_{(\cdot)}^{\max}$, $E_{(\cdot)}(A_m)$, and $D_{(\cdot)}(A_m)$, where $(\cdot)$ denotes either $\prng$ or $\qrng$. 
The average values of $A_{\qrng}^{\max} - A_{\prng}^{\max}$, denoted by $\mean{\alpha}$, and the average values of $E(A_m)$, denoted by $\mean{E}(A_m)$, are also given in Appendix~\ref{sec:summary_stats}.
We will use these data to answer the questions below.

\subsubsection{Overview of the results}\label{sec:results_summary}
Before addressing specific research questions, let us review the overall results. Table~\ref{tbl:final_outcomes_count} shows that the QRNG-based models win 60\% of the 120 experiments, tie 1\%, and lose 39\%.

Drilling down into specific reasons for winning, we can see that 54\% of the 72 winning experiments are all-around wins (\mask{w:w,w,wt}), where QRNG-based models achieve a higher level of accuracy faster. In 22\% of cases, QRNG-based models achieve higher accuracy, but $A_m$ values were reached more slowly. Another 13\% result in the outcome \mask{w:t,w,wt}, where QRNG-based models achieve comparable accuracy faster. Similarly, 10\% result in the outcome \mask{w:w,w,l}, where QRNG-based models achieve a higher level of accuracy faster but with higher variation. Finally, 1\% result in the outcome \mask{w:w,t,wt}, where QRNG-based models achieve a higher level of accuracy and reach the $A_m$ value simultaneously.

We have a single tie, where QRNG- and PRNG-based models achieve a comparable level of accuracy simultaneously.

Losses are divided into two groups: 53\% of the 47 losing experiments are those in which QRNG-based models achieve a similar level of accuracy more slowly (outcome \mask{l:t,l,*}). The remaining 47\% of the experiments result in outcome \mask{l:l,*,*}, where the QRNG-based model achieved a lower level of accuracy than the PRNG-based model.

\begin{figure}[htb]
    \centering
    \includegraphics[width=0.8\textwidth]{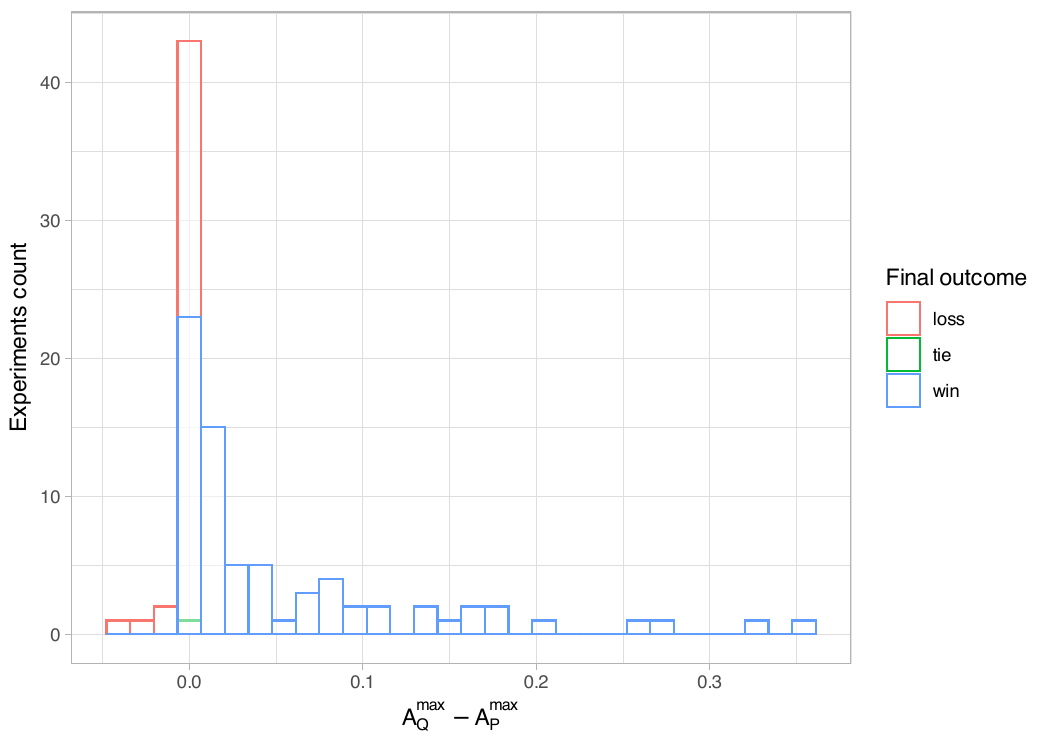}
    \caption{Histogram of the $A_{\qrng}^{\max} - A_{\prng}^{\max}$ values for three final outcomes.}
    \label{fig:max_accuracy_improvement}
\end{figure}

\begin{figure}[htb]
    \centering
    \includegraphics[width=0.8\textwidth]{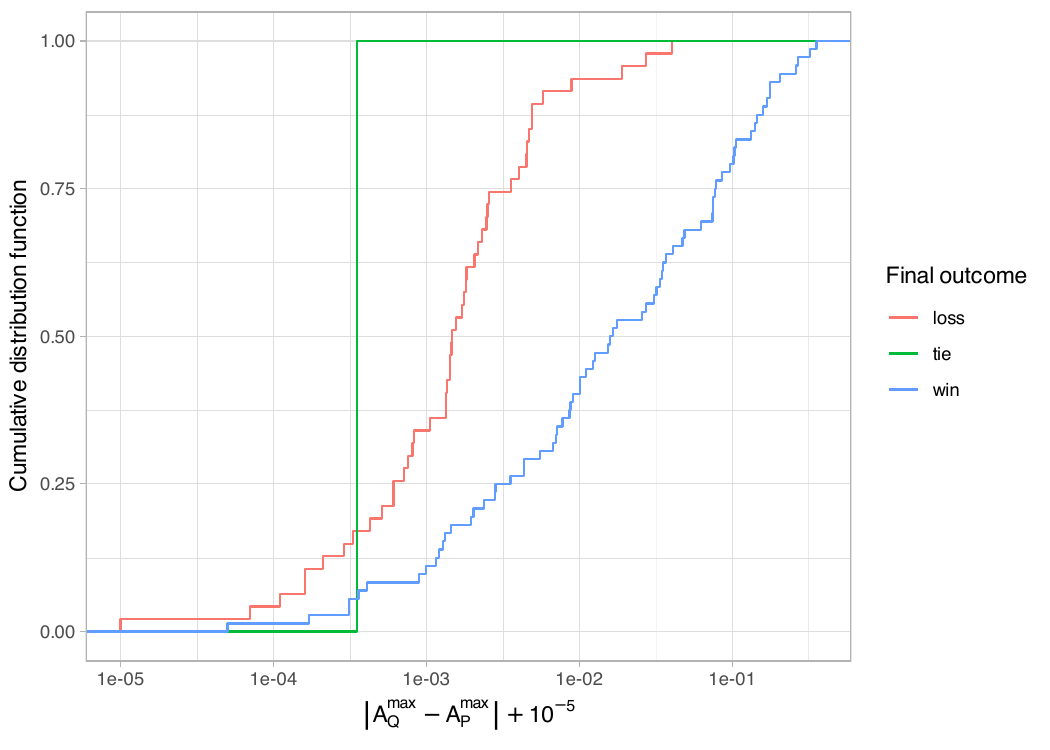}
    \caption{Empirical cumulative distribution function of $|A_{\qrng}^{\max} - A_{\prng}^{\max}| + 10^{-5}$ for three final outcomes. The $10^{-5}$ term is added to enable rendering of the $x$-axis values on a log scale.}
    \label{fig:abs_max_accuracy_improvement_cdf}
\end{figure}

Furthermore, Figure~\ref{fig:max_accuracy_improvement} indicates that when QRNG-based models lose, the decrease in the maximum median accuracy ($A_{\qrng}^{\max} - A_{\prng}^{\max}$) is minimal. Conversely, when QRNG-based models win, the improvement in maximum median accuracy is substantial. This is further illustrated in the empirical cumulative distribution function in Figure~\ref{fig:abs_max_accuracy_improvement_cdf}. The top 25\% of losses range between $\approx 0.0031$ and $0.0402$, while the top 25\% of wins range between $\approx 0.0775$ and $0.3550$. Therefore, it may be beneficial to try QRNG-based models.

Now, let us explore the answers to the specific research questions.

\begin{table}[htb]
\caption{Count of the final outcomes grouped by optimizer, dataset, and model. }
\label{tbl:final_outcomes_count}
\resizebox{\textwidth}{!}{%
\begin{tabular}{@{}l|rrrrrrrrrrrr|r@{}}
\toprule
            & \multicolumn{6}{c}{Adam}                                                            & \multicolumn{6}{c|}{SGD}                                                             & Grand \\ \cmidrule(lr){2-7} \cmidrule(l){8-13} 
Outcome     & \multicolumn{2}{c}{CIFAR-10} & \multicolumn{2}{c}{IMDB} & \multicolumn{2}{c}{MNIST} & \multicolumn{2}{c}{CIFAR-10} & \multicolumn{2}{c}{IMDB} & \multicolumn{2}{c|}{MNIST} & Total \\ \cmidrule(lr){2-3} \cmidrule(lr){4-5} \cmidrule(lr){6-7} \cmidrule(lr){8-9} \cmidrule(lr){10-11} \cmidrule(lr){12-13}
            & CNN           & MLP          & LSTM    & Transf.    & CNN         & MLP         & CNN           & MLP          & LSTM    & Transf.    & CNN         & MLP         &       \\ \midrule
\textsc{l:l,*,*}    & 4             & 1            & 1       & 2              & 5           & 5           & 1             &              &         & 2              & 1           &             & 22    \\
\textsc{l:t,l,*}    & 2             &              & 6       & 1              & 3           & 1           &               & 2            & 5       & 4              &             & 1           & 25    \\  \midrule
Loss Total  & 6             & 1            & 7       & 3              & 8           & 6           & 1             & 2            & 5       & 6              & 1           & 1           & 47    \\  \midrule
\textsc{t:t,t,t}    &               &              & 1       &                &             &             &               &              &         &                &             &             & 1     \\  \midrule
Tie Total   &               &              & 1       &                &             &             &               &              &         &                &             &             & 1     \\  \midrule
\textsc{w:t,w,wt}   &               &              & 2       & 2              &             &             &               &              & 5       &                &             &             & 9     \\
\textsc{w:w,l,*}    &               & 5            &         & 1              & 2           &             & 2             & 3            &         & 3              &             &             & 16    \\
\textsc{w:w,t,wt}   &               &              &         &                &             &             &               &              &         &                & 1           &             & 1     \\
\textsc{w:w,w,l}    &               & 2            &         &                &             &             & 1             & 3            &         & 1              &             &             & 7     \\
\textsc{w:w,w,wt}   & 4             & 2            &         & 4              &             & 4           & 6             & 2            &         &                & 8           & 9           & 39    \\  \midrule
Win Total   & 4             & 9            & 2       & 7              & 2           & 4           & 9             & 8            & 5       & 4              & 9           & 9           & 72    \\ \midrule
Grand Total & 10            & 10           & 10      & 10             & 10          & 10          & 10            & 10           & 10      & 10             & 10          & 10          & 120  \\
\bottomrule
\end{tabular}%
}
\end{table}

\subsubsection{RQ1. What effect does the selection of optimization algorithms have on deep neural networks' performance when initialized using the QRNG-based weight initialization scheme?}
Table~\ref{tbl:final_outcomes_count} shows significant performance differences in models based on the optimizer used. Models using QRNG-based initializers perform better with the SGD optimizer compared to Adam. Specifically, with SGD, models that use QRNG-based initializers outperform PRNG-based ones in 73\% of experiments without ties ($\mean{\alpha} \approx 0.06$). With Adam, they win in 47\% of 60 cases ($\mean{\alpha} \approx 0.05$) and tie in 2\%. 
This superior performance with SGD is likely due to SGD’s inability to dynamically adjust the learning rate, while Adam’s dynamic learning rate can compensate for the poorer starting conditions of the PRNG initializer. For ties and losses, $\mean{\alpha} \approx 0.00$, reinforcing that QRNG-based initialization may be beneficial.

\subsubsection{RQ2. How does a QRNG-based weight initialization scheme affect deep neural networks' performance?}\label{sec:rq_initializers}

Table~\ref{tbl:final_outcomes_count_initializer} indicates that QRNG initializers generally outperform PRNG initializers. Shape-agnostic initializers win in 47\% of 36 experiments without ties; shape-dependent initializers win in 67\% of 72 cases and tie in 1\%; orthogonal initializers win in 58\% of 12 experiments without ties. The average improvement in the maximum median accuracy for shape-agnostic, shape-dependent, and orthogonal initializers is $\mean{\alpha} \approx 0.02\%$, $0.07\%$, and $0.06\%$, respectively.

For individual initializers, Random Uniform performs poorly, winning only in 25\% of the cases, while Random Normal and Truncated Normal win in 58\%. For shape-dependent initializers, He Normal wins in 50\% of the cases, while others win or tie in 67\% to 75\%. Thus, QRNG benefits a range of initializers, perhaps with the exception of Random Uniform.

\begin{table}[htb]
\caption{Count of the final outcomes grouped by initializer.}
\label{tbl:final_outcomes_count_initializer}
\resizebox{\textwidth}{!}{%
\begin{tabular}{@{}l|rrrrrrrrrr|r@{}}
\toprule
            & \multicolumn{6}{c}{Shape dependent}                                                   & \multicolumn{3}{c}{Shape agnostic}            & \multicolumn{1}{c|}{Orthogonal} & Grand \\
            \cmidrule(lr){2-7} \cmidrule(lr){8-10} \cmidrule(lr){11-11}
Outcome    & Glorot & Glorot & He & He & Lecun & Lecun & Random & Random & Truncated & Orthogonal & Total \\ 
            & normal & uniform & normal & uniform & normal & uniform & normal & uniform & normal &  &  \\             
            \midrule
\textsc{l:l,*,*}    & 3  & 1  & 2  & 1  &    &    & 5  & 6  & 4  &    & 22  \\
\textsc{l:t,l,*}    & 1  & 2  & 4  & 3  & 3  & 3  &    & 3  & 1  & 5  & 25  \\ \midrule
Loss Total       & 4  & 3  & 6  & 4  & 3  & 3  & 5  & 9  & 5  & 5  & 47  \\ \midrule
\textsc{t:t,t,t}    &    &    &    & 1  &    &    &    &    &    &    & 1   \\ \midrule
Tie Total         &    &    &    & 1  &    &    &    &    &    &    & 1   \\ \midrule
\textsc{w:t,w,wt}   & 2  & 1  &    &    & 1  & 1  & 1  & 1  & 1  & 1  & 9   \\
\textsc{w:w,l,*}    & 2  & 2  & 2  & 3  & 2  & 2  &    &    &    & 3  & 16  \\
\textsc{w:w,t,wt}   &    &    &    &    &    &    &    &    &    & 1  & 1   \\
\textsc{w:w,w,l}    & 1  & 1  & 1  &    & 1  & 1  &    & 1  &    & 1  & 7   \\
\textsc{w:w,w,wt}   & 3  & 5  & 3  & 4  & 5  & 5  & 6  & 1  & 6  & 1  & 39  \\ \midrule
Win Total         & 8  & 9  & 6  & 7  & 9  & 9  & 7  & 3  & 7  & 7  & 72  \\ \midrule
Grand Total & 12 & 12 & 12 & 12 & 12 & 12 & 12 & 12 & 12 & 12 & 120 \\ \bottomrule
\end{tabular}%
}
\end{table}

\subsubsection{RQ3. How does the choice of deep neural network architecture affect the model's performance when it is initialized using the QRNG-based weight initialization scheme?}\label{sec:rq_architecture}

For CNN architecture, QRNG-based initializers win in $60\%$ of 40 cases ($\mean{\alpha}\approx 0.08$) without tie, for MLP~--- win in $75\%$ of 40 cases ($\mean{\alpha}\approx 0.07$) without tie, for LSTM~--- win in $35\%$ of 20 cases ($\mean{\alpha}\approx 0.00$) and tie in $5\%$, for Transfomer~--- win in $55\%$ of 20 cases ($\mean{\alpha}\approx 0.01$) without ties.

Thus, the strongest improvements come from MLP and CNN architectures trained on computer vision datasets. 
However, $\mean{E}(A_m) \approx -36\%$ for LSTM and $\mean{E}(A_m) \approx -7\%$ for Transformer, suggesting that QRNG-based initialization may not always yield higher accuracy but can achieve comparable accuracy faster, reducing training time and saving computational resources.

Moreover, in 60\% of the 10 LSTM cases using the Adam optimizer, the result is \mask{l:t,l,*} (i.e., the same accuracy is achieved slower). The comparison of raw data in Appendix~\ref{sec:accuracy_figures} suggests that LSTM reach the value $A_m$ more quickly. However, the seed selection penalty ($\Delta_\qrng=4$) results in a performance loss. Improving the seed selection heuristic could potentially mitigate this issue in the future (see Section~\ref{sec:open_questions} for more details).

Consequently, all deep neural network architectures studied show benefits in either maximum accuracy or training speed.

\subsubsection{RQ4. How does the change in the dataset affect the performance of deep neural networks initialized with the QRNG-based weight initialization scheme?}\label{sec:rq_dataset}

Table~\ref{tbl:final_outcomes_count} indicates that the impact of different datasets varies. For the MNIST dataset, QRNG-based initializers win in 60\% of 40 cases ($\mean{\alpha}\approx 0.03$) without ties. For CIFAR-10, they win in 75\% of 40 cases ($\mean{\alpha}\approx 0.11$) without tie. For IMDB, they win in 45\% of 40 cases ($\mean{\alpha}\approx 0.00$)  and tie in 3\%.

Partitioning data by optimizer reveals more details. For MNIST with SGD, models with QRNG-based initializers win in 90\% of 20 cases ($\mean{\alpha}\approx 0.04$), while with Adam, they win in 30\% of cases ($\mean{\alpha}\approx 0.01$). 

Our analysis of the accuracy distributions given in the Appendix~\ref{sec:accuracy_figures} suggests that the improvement of the accuracy of PRNG and QRNG with epochs is similar. The accuracy distributions in Appendix~\ref{sec:accuracy_figures} suggest similar accuracy improvements with epochs for PRNG- and QRNG-based models, but the QRNG seed selection heuristic incurs a computational penalty $\Delta_\qrng$. This issue may be resolved in the future by improving seed selection heuristics (we will discuss this in Section~\ref{sec:open_questions}).

For CIFAR-10, the results are more balanced: QRNG-based initializers win in 75\% of 20 cases with SGD ($\mean{\alpha}\approx 0.11$) and in 65\% of 20 cases with Adam ($\mean{\alpha}\approx 0.10$). Compared to MNIST, CIFAR-10 is a more challenging dataset, making model learning more difficult. Thus, QRNG-based initialization may help learn more complex data representations.

For IMDB, the results are fairly balanced. Models with QRNG-based initializers win in 45\% of the 20 cases with Adam ($\mean{\alpha}\approx 0.00$) and tie in 5\% of the experiments, while with SGD they win in 45\% of the cases ($\mean{\alpha}\approx 0.00$). 

Based on these data, is QRNG-based initialization more helpful for computer vision datasets than natural language datasets? The short answer is ``not necessarily.'' While, on average, we were unable to improve the accuracy of the LSTM and Transformer models using QRNG-based initialization (trained on the natural language dataset), we saw that the same accuracy can be achieved faster (as shown in Section~\ref{sec:rq_architecture}), which is still advantageous.

\subsubsection{Hypothesis}
The answers to the research questions support the hypothesis that QRNG-based weight initialization can significantly accelerate neural network training. Improvements are evident across data sets, architectures, initializers, and optimization algorithms.

\subsection{Discussion}
\subsubsection{Practical considerations}\label{sec:practical considerations}
We provide a reference implementation of QRNG-based initializers for Keras~v.3.3.3, which can readily use three different backends: JAX~\cite{jax2018github}, PyTorch, or TensorFlow. It can also be easily ported to other popular machine learning frameworks. The changes to the initializer's underlying random number generator do not require any changes downstream: the code for compiling the model and training it remains unchanged. By passing a QRNG-based initializer instance to a neural network layer, practitioners can treat these implementations as ``black boxes''. This process conceals the complexities of the underlying implementation from the practitioner. Those who wish to engage more deeply may draw random values with specific seeds that can be manually applied to the model weights.

In Appendix~\ref{sec:single_layer_study}, we observe that a simple neural network performs better with a number of weights following the pattern $4, 8, 12, \ldots$. Thus, when designing neural network layers, it is advisable to select the number of weights that adhere to this pattern (which is often the default behaviour\footnote{As $2^n$ is divisible by $4$ for $n\ge2$.}).

\subsubsection{Limitations}\label{sec:threats_to_validity}
This section discusses limitations and threats to the validity of our study classified as per \cite{yin2009case}.

Inaccuracies in our implementation or in the code of the underlying packages can affect the results, threatening \textit{internal validity}. To mitigate the first threat, our QRNG-based implementations have undergone code peer review and were tested against Keras/TensorFlow PRNG-based distributions.

The second threat may be related to issues with the implementation of Philox PRNG in TensorFlow backend. By training a simple MLP for a single epoch using Philox and the Mersenne Twister algorithm, we found that the distributions of the results were similar (see Appendix~\ref{sec:single_layer_study_prng} for details). Consequently, the observed behaviour is not unique to a particular PRNG or implementation.

To focus on the effect of random generators, we need to isolate other sources that may affect the model's performance. We identified five main areas affecting model performance that may threaten \textit{conclusion validity}:
\begin{inlinelist}
    \item random selection of training, validation, and testing datasets
    \item stochastic nature of the optimizer
    \item sophisticated constructs in models' archtitecture
    \item speed of the QRNG-based initializers
    \item limited number of epochs.
\end{inlinelist}

To address the first two issues, we ran each experiment \num{100} times. Although data shuffling (done after each epoch) may yield different outcomes, we expect the relative performance differences to remain consistent.  We ran the formal experiments only on CPUs to eliminate the effect of randomness from the GPU software stack (although we ran the code informally on GPUs and observed similar results). 

To address the third issue, we simplified model architectures to avoid the effects of more sophisticated techniques. Although this reduces the overall model efficacy, our primary goal is to assess the effects of random generators, not optimize performance.

QRNG-based initializers are generally slower than PRNG-based initializers. The computational complexity is $O\left(d^2 N_{\max}\right)$, where $N_{\max}$ is the maximum number of elements for any of the $d$ dimensions; see Appendix~\ref{sec:distr_performance} for details and a way to reduce the complexity to $O\left(d N_{\max}\right)$. However, the initialization time for setting the weights before training is negligible compared to the duration of model training. We timed our initialization code to assess the effect of the fourth issue. Appendix~\ref{sec:distr_performance} shows that drawing the values from the distributions takes only a fraction of a second, while training models can take minutes or longer. As a result, QRNG-based initializer can reduce the overall training time and achieve an accuracy comparable to PRNG with fewer epochs.

Finally, we capped training at 30 epochs in our experiments to ensure consistency across all setups. Eventually, if we train our models for a larger number of epochs, PRNG-based models may become more accurate than QRNG-based models, or vice versa (i.e., a different model may reach a higher value of $A_m$ than we have seen in our experiments). The data in Appendix~\ref{sec:accuracy_figures} show that this is unlikely for most cases based on the analysis of the raw accuracy plots and the changes in the median accuracy. However, it is possible, though not likely, that extended training could yield different results. Thus, our findings should be interpreted within the context of a 30-epoch training limit.

The limited generalizability of our findings challenges its \textit{external validity}. Although we examined three datasets (from two modalities), four model architectures, two optimizers, and ten initializers, we cannot guarantee that these results will apply to other datasets, architectures, optimizers, or initializers. However, the promising results warrant further investigation. The same empirical methodology can be applied to other modeling scenarios through rigorously designed experiments.
The following are some open questions for future research.

\subsubsection{Open questions}\label{sec:open_questions}
In our study, we demonstrated that QRNG-based initializers can outperform PRNG-based ones. Many questions remain unanswered; below, we sketch some potential future research avenues.

Why is QRNG-based initialization effective? It is possible that insights from Monte Carlo methods research, where QRNG initialization is commonly used, can be applied here. The dimensional integers used to seed QRNG (see Appendix~\ref{sec:sobol} for details) are optimized  pairwise (by minimizing the number of bad correlations between pairs of variables)~\cite{joe2008constructing}. The creators of these dimensional integers suggest that the success of simple pairwise optimization may be because many practical problems can be reduced to ``low effective dimension''~\cite{caflisch1997valuation}, which ``means either that the integrand depends mostly on the initial handful of variables, or that the integrand can be well approximated by a sum of functions with each depending on only a small number of variables at a time''~\cite{joe2008constructing}, see~\cite{wang2003effective,wang2005why,wang2008low} for further analysis. These relations capture enough information necessary for decision making. If we apply similar logic to neural networks initialization, initializing the weights uniformly at the level of individual layers and adding additional uniformity for each pair of layers may give an optimizer a better initial condition than RPNG-based initialization. 
In order to test whether these conjectures remain valid empirically, more complex models could be tested on larger, more intricate datasets and tasks.

Why Random Uniform initializer benefits the least from QRNG? In Section~\ref{sec:rq_initializers}, we observe that most initializers benefit from QRNG with the exception of the Random Uniform. This initializer mutates starting random numbers the least, simply by rescaling them to the $[-0.05, 0.05]$ range (see Appendix~\ref{sec:initializers} for details). Can this behaviour help us better understand why QRNG-based initialization yields better results in other cases?

How does the performance of QRNG-based initializers vary with the number of weights? As mentioned in Section~\ref{sec:practical considerations}, a simple neural network performs better with a number of weights following the pattern $4,8,12,\ldots$. This raises questions about the underlying factors of this behaviour, perhaps related to balance and QRNGs' low-discrepancy characteristics.

When a sequence length is not a power of two, Sobol' sequences lose balance~\cite{owen2022dropping}. Although our preliminary tests with a simple model (in Appendix~\ref{sec:single_layer_study_qrng}) did not show significant practical effects of this factor, we speculate that these effects may be more pronounced in larger and more complex models. Therefore, it is an open question whether incorporating this factor into the design of neural network layers can further improve QRNG initialization.

Sobol' sequence performance may be degraded~\cite{owen2022dropping} if a first element is removed (as in our implementation). According to our initial analysis, excluding the zero element did not significantly affect neural network training and initialization. However, further investigation is needed to determine whether the QRNG performance can be improved.

The seeds for our QRNG-based initializers are selected sequentially using a heuristic (Appendix~\ref{sec:seed}), which is not optimal (as shown in Appendix~\ref{sec:single_layer_study_qrng}). An alternative approach to seed selection\footnote{For example, selecting a seed at random for each layer, which would set $\Delta_\qrng=0$, and potentially leading to a reduction of \mask{l:t,lt,*} and \mask{w:w,lt,*} outcomes, where \mask{lt} would be replaced with \mask{w}.} or refining the heuristic could improve our results.

We have only explored one version of the Sobol' sequences (with a specific set of direction numbers). Other versions exist (e.g., scrambled sequences~\cite{owen1998scrambling} or those with optimization after sampling~\cite{scipy_sobol}) raising the question of which version is most effective for QRNG applications in initializers. Our initial analysis shows scrambled sequences produce similar results to unscrambled sequences but introduce numerical instability. However, quantitative assessments need more research.

Other quasirandom sequences exist, such as the Faure, Halton, and Niederreiter sequences. Would those sequences perform better than Sobol's sequences if we used them in QRNG-based initializers?

QRNG has a greater impact on shape-dependent and orthogonal initializers than on shape-agnostic ones (as shown in Section~\ref{sec:rq_initializers}). More research is required to determine the reasons for these discrepancies.

Furthermore, while our initial experiments isolated QRNG's effects using simple models and datasets, further research would require scaling up complexity, varying data modalities and volumes, model sizes, and optimizers to fully grasp QRNG's implications.

\section{Related Work}\label{sec:related-work}
The following is a brief summary of related literature; all these works are complementary to our own.

Low-discrepancy sequences were tried for various tasks in machine learning, such as selecting a subset of data for training~\cite{mishra2021enhancing}, generating neural networks~\cite{keller2020artificial} and their surrogates~\cite{longo2021higher}, replacing the backpropagation algorithm~\cite{jordanov1999neural}, improving optimizers~\cite{bangyal2022improved},  updating belief in
Bayesian networks~\cite{cheng2000computational}, and tuning hyperparameters and improving regularization~\cite{bergstra2012random}. To the best of our knowledge, researchers have not yet used QRNGs to initialize neural network weights.

Other random number generators have been explored to initialize neural networks. For example, node dropout regularization was performed using true random number generators~\cite{koivu2022quality}. Finally, machine learning models (including dense neural networks) may perform better when initialized with random numbers generated by quantum computers~\cite{bird2020effects} (although other researchers were unable to replicate these results~\cite{heese2021effects}). 

\section{Summary}\label{sec:summary}
We demonstrated that QRNG-based initializers can achieve higher accuracy or speed up neural network training in 60\% of the 120 experiments conducted. The top 25\% of the improved maximum median accuracies range between $\approx 0.0775$ and $0.3550$. The negative side effects of trying QRNG are minimal, with the top 25\% of observed losses resulting in a drop in accuracy between $\approx 0.0031$ and $0.0402$. This trend is observed across various data types, architectures, optimizers, and initializers.

The extent to which this effect generalizes remains to be seen. We encourage the community to explore and validate our findings.

\section*{Acknowledgements}
This work was partially supported by the Natural Sciences and Engineering Research Council of Canada (grant \# RGPIN-2022-03886). The authors thank the Digital Research Alliance of Canada and the Department of Computer Science at Toronto Metropolitan University for providing computational resources. The authors also express their gratitude to the members of our research group~---~Montgomery Gole, Mohammad Saiful Islam, and Mohamed Sami Rakha~---~for their valuable feedback on the manuscript and insightful discussions.

\clearpage 

\printbibliography

%%%%%%%%%%%%%%%%%%%%%%%%%%%%%%%%%%%%%%%%%%%%%%%%%%%%%%%%%%%%

\clearpage
\appendix

\section{Low Discrepancy Sequences}\label{sec:sobol}
Sobol' sequences, introduced by Sobol' in 1967~\cite{sobol1967distribution}, are quasirandom sequences widely used in numerical analysis and Monte Carlo methods~\cite{jackel2002monte}. They provide a deterministic and evenly distributed set of points in multidimensional space, thereby overcoming some limitations of traditional pseudorandom sequences \cite{sobol1967distribution, jackel2002monte}.

Sobol' sequences can be constructed efficiently~\cite{anotonov1979economic} using the following recursive formula: 
\begin{equation}\label{eq:sobol}
x_{n,k} = x_{n-1,k} \oplus_2 v_{j,k},
\end{equation}
where $x_{n,k}$ is the $n$-th draw of Sobol' integer in dimension $k$, $v_{j,k}$ is a direction integer (needed to initialize the recursion), $\oplus_2$ is a single XOR operation for each dimension. The recursion starts with $x_{0,k}=0$. 

Various methods exist to compute $v_{j,k}$. In our work, we use Sobol' sequences~\cite{sobol1967distribution} implemented\footnote{The implementation seems to be based on the Algorithm 659~\cite{bratley1988algorithm}.} in SciPy v.1.13.1 software~\cite{2020SciPy-NMeth,scipy_sobol}, which is configured to produce Sobol' sequences for up to \num{21200} dimensions using direction integers $v_{j,k}$ computed by~\cite{joe2008constructing}. As we use individual dimensions to initialize the networks' layers, this number of dimensions is more than adequate. A user who needs more than \num{21200} dimensions may, e.g., reuse the dimensions or try scrambled versions~\cite{owen1998scrambling} of Sobol's sequences.

In general, Sobol' sequences have $d$ dimensions and are in the range $[0,1)^d$.  We discard the first element of a sequence (which is always $0$), similar to the TensorFlow v.2.16.1 implementation~\cite{tensorflow2015-whitepaper,tf_sobol}. Due to this, our implementation of Sobol' sequences reduces the range to $(0,1)^d$. It is necessary to discard this element in order to implement QRNG-based normal and truncated normal distributions, which we will discuss in Appendices~\ref{sec:qrng-normal} and~\ref{sec:qrng-truncated-normal}, respectively.

\section{Sampling from distributions}\label{sec:distributions}
In this work, we will need to sample values from univariate uniform, normal, and truncated normal distributions, denoted by $\uniform{(\cdot)}$, $\normal{(\cdot)}$, and $\tnormal{(\cdot)}$, respectively. The subscript $(\cdot)$ indicates the random number generator to use for sampling and takes the value $\prng$ for PRNG and $\qrng$ for QRNG.

\begin{figure}[htb]
    \centering
    \includegraphics[width=\textwidth]{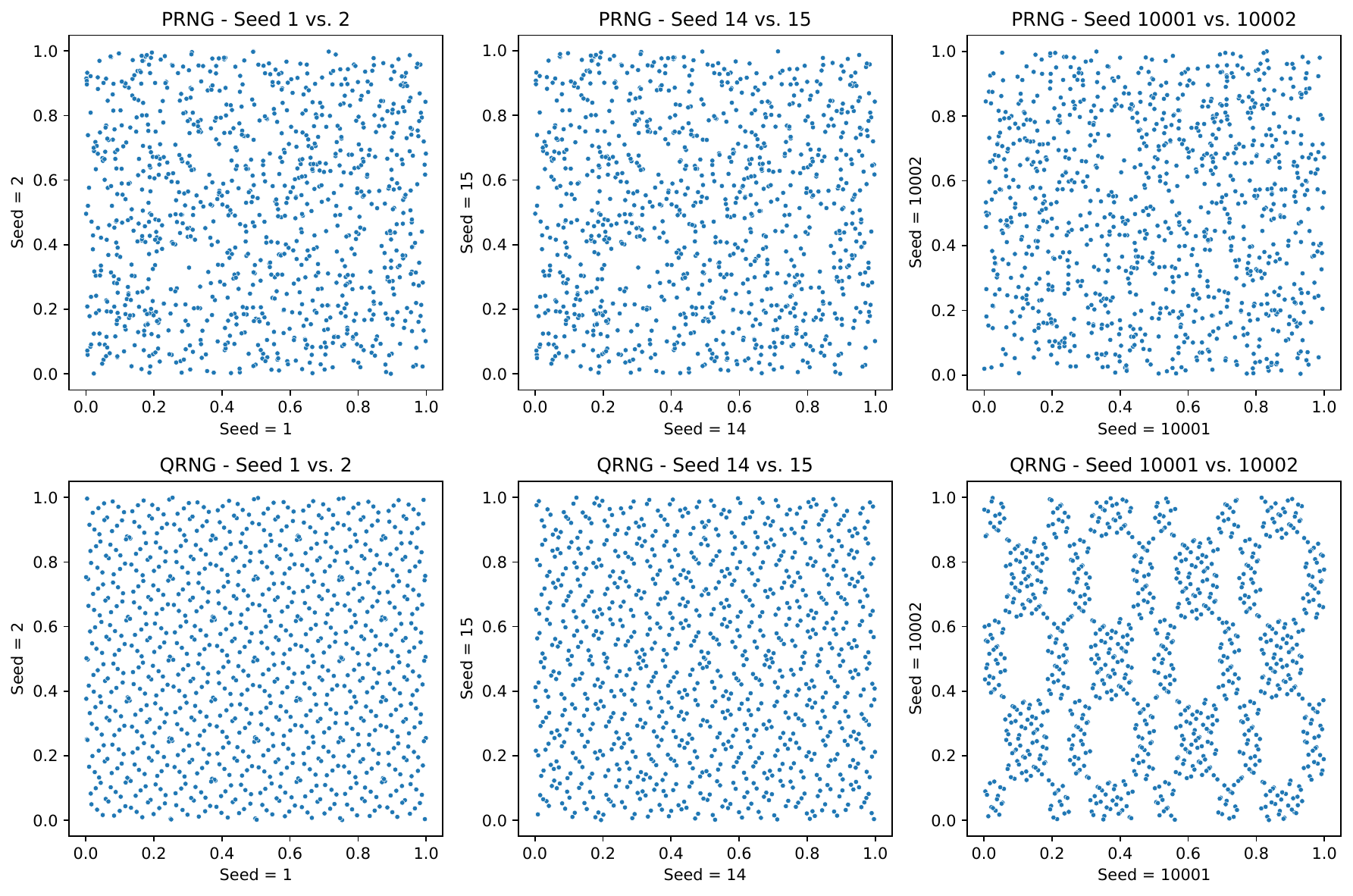}
    \caption{A two-dimensional projection of the first 1024 draws of the Keras/TensorFlow pseudorandom number generator (top pane) and the Sobol' sequences (bottom pane). Axis labels denote seed values of the random generator.}
    \label{fig:prng_vs_qrng}
\end{figure}

Figure~\ref{fig:prng_vs_qrng} shows two-dimensional projections for PRNGs and QRNGs under study. Note that QRNG aims to cover a domain as homogeneously and uniformly as possible.

\begin{figure}[htb]
    \centering
    \includegraphics[width=\textwidth]{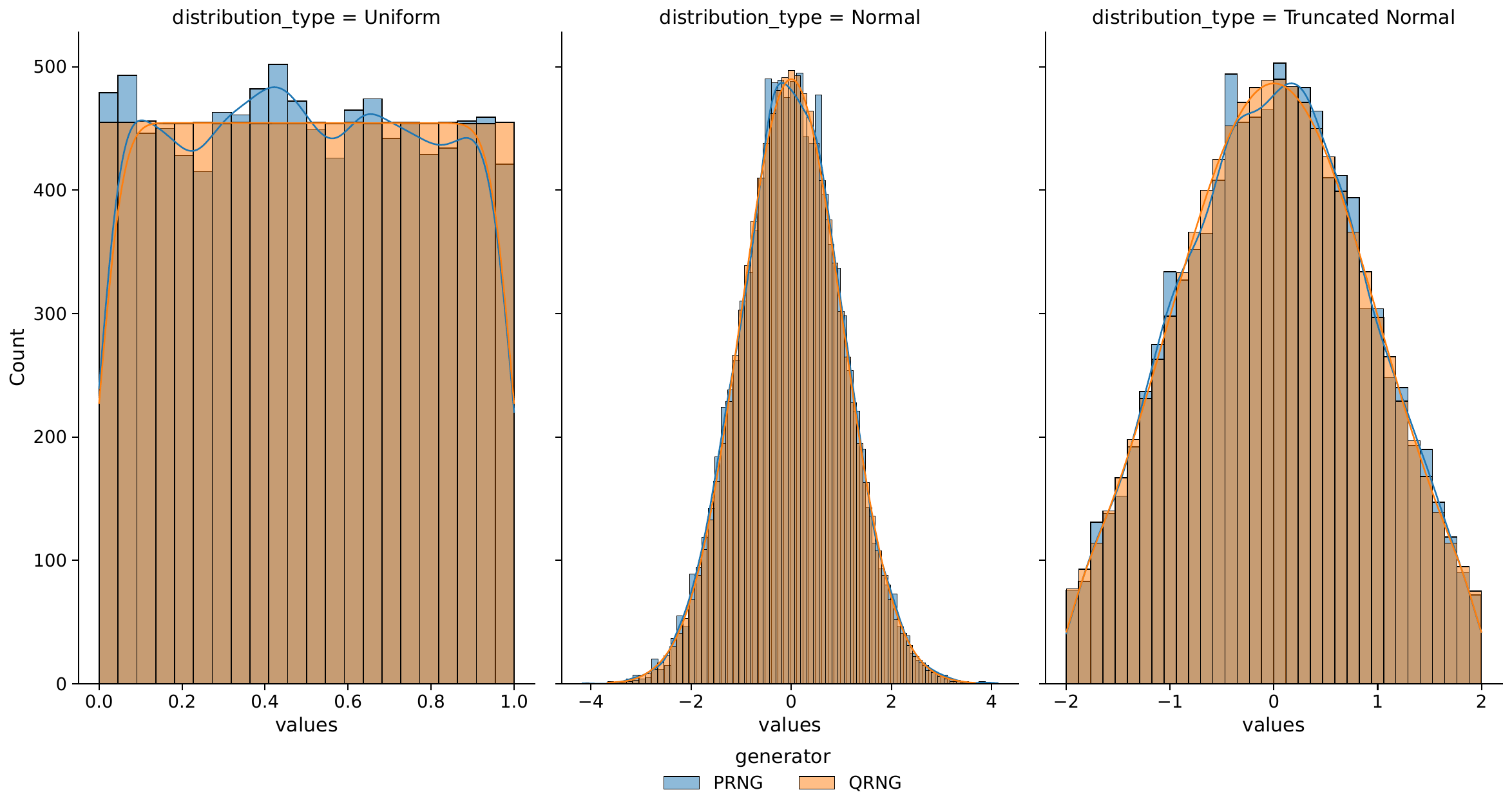}
    \caption{Sample draws from univariate uniform, normal, and truncated normal distributions.}
    \label{fig:prng_vs_qrng_distr}
\end{figure}

The distributions obtained using PRNG and QRNG are compared in Figure~\ref{fig:prng_vs_qrng_distr}. The QRNG version is ``smoother'' than the PRNG version. This is because QRNGs (that use Sobol' sequences discussed in Appendix~\ref{sec:sobol}) have better uniformity properties than PRNGs. Below are details on how distributions are implemented.

\subsection{Using Keras/TensorFlow PRNG}\label{sec:distr_prng}
Keras v.3.3.3~\cite{chollet2015keras} can use three different backends: JAX, PyTorch, or` TensorFlow. We use TensorFlow v.2.16.1~\cite{tensorflow2015-whitepaper}. This version of Keras is designed to use backend implementations of the distributions. TensorFlow v.2.16.1 uses continuous uniform, normal, or truncated normal distributions sampled with PRNGs. These implementations are widely used and well tested. Thus, we treat them as baselines and leverage as-is.  Below are some notes on TensorFlow's implementation of the distributions.

\subsubsection{Uniform distribution} \label{sec:prng_uniform}

TensorFlow draws the values from $\uniform{P}$ in the range $[0, 1)$~\cite{tf_uniform}. 
The PRNG algorithm used by TensorFlow depends on the hardware~\cite{tf_rnd_algorithm}. 
In most cases, it will be either Philox or ThreeFry~\cite{salmon2011parallel}, but it can also select another algorithm from the underlying libraries~\cite{tf_rnd_algorithm}. 
We may not know which specific algorithm is used in advance (although it is typically Philox~\cite{tf_rnd_gen}). However, the selection of the PRNG algorithm is deterministic (i.e., a testbed should select PRNG consistently). 

We empirically validated that on our testbed the sequences produced by automatic selection matched those produced by the Philox algorithm. Thus, our experiments initialize neural network weights using Philox-based PRNGs. Tensorflow may use different PRNG algorithms for other sources of randomness, such as those used by the SGD and Adam optimizers. Nevertheless, they will be consistent for a given testbed.

\subsubsection{Normal distribution} TensorFlow draws the values from $\normal{P}$ in the range $(-\infty, +\infty)$~\cite{tf_normal}. 
The mean $\mu \in \sR$ and standard deviation $\sigma \in \sR_{> 0}$ (governing the distribution) vary based on application. TensorFlow's manual and associated Python code listing do not describe the technical details of sampling from the distribution. However, code analysis indicates that the Box-Muller transform~\cite{box1958note} is used.

\subsubsection{Truncated normal distribution}\label{sec:prng_tnormal}
TensorFlow implements a specific case of a $\tnormal{P}$ distribution: the values are drawing from $\normal{P}$ in the range $[-2\sigma, +2\sigma]$~\cite{tf_tnormal}. Technically, the values are drawn from $\normal{P}$ with a specified $\mu$ and $\sigma$, discarding and re-drawing samples with more than two standard deviations from the mean~\cite{tf_tnormal}.

\subsection{Using QRNG}\label{sec:distr_qrng}
We use the inverse transform sampling approach\footnote{Inverse transformation sampling begins with uniform samples in the range between $0$ and $1$. These samples can be viewed as probabilities, which can be fed into an inverse cumulative distribution function (of a distribution of interest). Details of the method can be found in~\cite[Section 2.2]{devroye1986non_uniform}.}, followed by scaling and shifting. Let us look at the details.

\subsubsection{Implementation notes}\label{sec:implementations}
Our QRNG-based distributions are tested against the baseline Keras v.3.3.3 and TensorFlow v.2.16.1 implementations distributions (described in Appendix~\ref{sec:distr_prng}) to assure correctness of implementation. 

Files \verb|src/distributions_qr.py| and \verb|tests/test_distributions_qr.py| contain the code and associated test cases, respectively~\cite{source_code}. Essentially, we modify Keras~\cite{chollet2015keras} classes that are used to sample random numbers.

Appendix~\ref{sec:sobol} describes how we implement $d$-dimensional Sobol' sequences in the range of $(0,1)^d$. To maintain low discrepancy properties of Sobol' sequences, we should sample from the beginning of the sequence for a given dimension $k$. Informally, we can refer to $k$ as the seed for our QRNG.

\subsubsection{Uniform distribution}

Let us sample a vector of $n$ random numbers $\rvu$  from a uniform distribution (based on Sobol' sequences). It can be achieved by scaling and shifting a Sobol' sequence of length $n$ for $k$-th dimension (which gives us an inverse of the  cumulative distribution function for the uniform distribution):
\begin{equation}\label{eq:uniform}
    \rvu \sim  \uniform{Q}(a,b;k) = (b-a) \rvs_k + a,
\end{equation}
with $\uniform{Q}(a,k)$ denoting a uniform distribution with the range $(a,b)$ and $\rvs_k$ denoting a Sobol' sequence of length $n$ for dimension $k$.

We can now use $\uniform{Q}(0,1,k)$ to get samples for normal and truncated normal distributions via the inverse transform sampling approach, as discussed below.

\subsubsection{Normal distribution}\label{sec:qrng-normal}
To sample  $n$ random numbers $\rvn$  from a normal distribution that uses Sobol'-sequences based QRNG, we use:
\begin{equation}\label{eq:normal}
    \rvn \sim  \normal{Q}(\mu,\sigma;k) = \Phi^{-1}\left[\uniform{Q}(0,1;k) \right]\sigma+\mu = \Phi^{-1}(\rvs_k )\sigma+\mu,
\end{equation}
where $\normal{Q}(\mu,\sigma;k)$ denotes a normal distribution with the mean $\mu$ and standard deviation $\sigma$ and $\Phi^{-1}$ is the inverse cumulative distribution function of the standard normal distribution\footnote{A standard normal distribution has a mean of zero and a standard deviation of one.}. We use an implementation of $\Phi^{-1}$ provided by SciPy v.1.13.1.

As $\uniform{Q}(0,1;k)$ returns values in the range $(0,1)$, theoretically, the output of $\Phi^{-1}$ should always be finite. Potentially, when working with floating points numbers, $\lim_{x \to 0} \Phi^{-1}(x) = -\infty$ and $\lim_{x \to 1} \Phi^{-1}(x) = \infty$. We have not experienced this issue in practice since we skip the first element of the Sobol' sequence. However, this issue will manifest itself if the first element is not skipped (which can be mitigated using scrambled Sobol' sequences~\cite{owen1998scrambling}).

\subsubsection{Truncated normal distribution}\label{sec:qrng-truncated-normal}
To sample  $n$ random numbers $\rvt$  from a truncated normal distribution that uses Sobol'-sequences based QRNG, so that we match the $[-2\sigma, +2\sigma]$ range discussed in Appendix~\ref{sec:prng_tnormal} we do
\begin{equation}\label{eq:tnormal}
    \begin{split}
    \rvt \sim  \tnormal{Q}(\mu,\sigma;k) &= \Phi^{-1} \{ \Phi(\alpha) + \uniform{Q}(0,1;k)  \left[\Phi(\beta) - \Phi(\alpha)\right]\}  \sigma + \mu \\
    &= \Phi^{-1} \{\underbrace{ \Phi(\alpha) + \rvs_k  \left[\Phi(\beta) - \Phi(\alpha)\right]}_{\gamma}\}  \sigma + \mu \\
    &\approx \Phi^{-1} \{ 0.02 +   0.95 \rvs_k \}  \sigma + \mu,            
    \end{split}
\end{equation}
where $\Phi$ is the cumulative distribution function of the standard normal distribution, $\alpha = -2$, and $\beta = 2$. We use an implementation of $\Phi$ and $\Phi^{-1}$ provided by SciPy v.1.13.1. The values of $\alpha$ and $\beta$ are chosen to satisfy the range $[-2\sigma, +2\sigma]$ requirement.

Equation~\ref{eq:tnormal} is slow (since $\Phi$ and $\Phi^{-1}$ are expensive to compute) but numerically stable.  The stability analysis is detailed below. 

As above, $\lim_{\gamma \to 0} \Phi^{-1}(\gamma) = -\infty$. Thus, Equation~\ref{eq:tnormal} may be problematic when an element in $\rvs_k$, denoted $\rs$, yields $\gamma \approx 0$:
\begin{equation*}
    \begin{gathered}
   \gamma = \Phi(\alpha) + \rs  \left[\Phi(\beta) - \Phi(\alpha)\right] = 0 \Rightarrow \\           
     \rs = \frac{\Phi(\alpha)}{\Phi(\alpha) - \Phi(\beta)} \approx -0.02.\\
    \end{gathered}
\end{equation*}
Therefore,  $\Phi^{-1}(\gamma) \neq -\infty$ for all $\rs$, as $\rs > 0$.
Similarly, $\Phi^{-1} (\gamma) \neq \infty$ because $\lim_{\rs \to 1} \gamma \approx 0.98$.

\subsection{Performance analysis}\label{sec:distr_performance}

Suppose that we want to create $d$-dimensional sequence of random values with $N_k$ values for each dimension. Further, suppose that the values for the $k$-th dimension should be drawn independently of the previous dimensions (which is typical for our use case, since we usually want to draw the sequence for a particular value of $k$). 

PRNG generators implement all three distributions (random uniform, random normal, and truncated normal) independently of their seed values. 

QRNGs based on Sobol' sequences are different. Because Equation~\ref{eq:sobol} is recursive, we must draw $N_k$ values for dimensions $1$ through $k$ and then discard the values for dimensions $1$ to $k-1$. In this scenario, the total computation cost would be:
\begin{equation}\label{eq:comp_complexity}
    \sum_{k=1}^{d}{k N_k} \ge \sum_{k=1}^{d}{k N_{\max}} = \frac{d(d+1)}{2} N_{\max} = O\left(d^2 N_{\max}\right),
\end{equation}
where $N_{\max}$ is the maximum number of elements for any of the $d$ dimensions\footnote{Algorithm 659~\cite{bratley1988algorithm} explicitly adds the cost of XOR bit manipulation $O[\log(d)]$. The computational cost of this calculation is negligible on modern computers, so we eliminate this term.}.

\begin{figure}[ht]
    \centering
    \includegraphics[width=\textwidth]{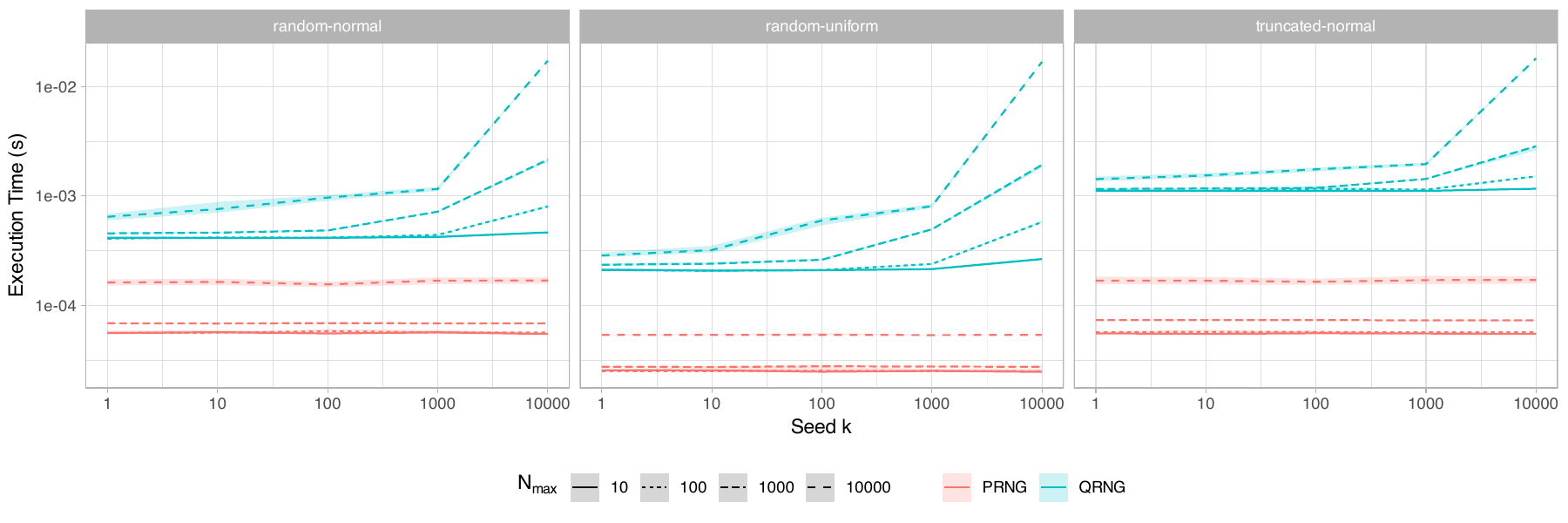}
    \caption{Time needed to draw the $N_{\max} = 10, 100, 1000, 10000$ values from the random normal $\normal{(\cdot)}(0,1; k)$,  random uniform $\uniform{(\cdot)}(0,1; k)$, and truncated normal $\tnormal{(\cdot)}(0,1; k)$ distributions.  $x$-axis shows seed value $k$, $y$-axis shows execution time measured in seconds. The lines represent median accuracy based on 1000 repetitions, while the ribbons represent the range between lower and upper quartiles.}
    \label{fig:timing}
\end{figure}

Figure~\ref{fig:timing} presents the amount of time needed to draw the $N_{\max}$ values from the random uniform $\uniform{(\cdot)}(0,1; k)$, random normal $\normal{(\cdot)}(0,1; k)$, and truncated normal $\tnormal{(\cdot)}(0,1; k)$ distributions. As expected, in the case of PRNG the amount of time required to generate the random numbers is independent of the seed (the lines stay flat). The amount of time required to generate values from the random uniform distribution is the fastest; the time to generate draws from random normal and truncated normal distributions is similar but slower than for the random uniform distribution, since they use a version of the Box-Muller transform~\cite{box1958note}. 

QRNG requires more time to draw from random uniform, random normal, and truncated normal distributions than their PRNG counterparts. Within the QRNG group, random uniform is the fastest, followed by random normal, and truncated normal. This is expected based on the number of operations in Equations~\ref{eq:uniform}, \ref{eq:normal}, and \ref{eq:tnormal}. For all three distributions, execution time increases with $k$ (as expected based on Equation~\ref{eq:comp_complexity}).

As mentioned above, our QRNG-based formulas are slower than Keras/TensorFlow's PRNG-based formulas. However, the time spent initializing the weights before training is immaterial compared to the time spent training the model. As shown in Figure~\ref{fig:timing}, drawing the values from the distributions for a specific $k$ takes only a fraction of a second (even  for $N_{\max} = 10000$). Comparatively, training a model takes seconds to minutes. Thus, we can still save time training the model (since fewer epochs are required to achieve a certain level of accuracy).

Caching Sobol' sequences can improve implementation performance if we need to initialize weights quickly (e.g., because we use very large $k$ values or because every second matters). The sequences can be cached in a $N_{\max} \times d$ table if we know in advance $N_{\max}$ and $d$. In this case, the computations will cost only 
\begin{equation*}
 \sum_{k=1}^{d}{N_{\max}} = d N_{\max} = O\left(d N_{\max}\right).
\end{equation*}
Using this method, the order of computational complexity of QRNG would be reduced to that of PRNG.

\subsection{Using Mersenne Twister PRNG}\label{sec:distr_prng_mt}
We use the inverse transform sampling approach discussed in Appendix~\ref{sec:distr_qrng} to compute random uniform, random normal, and truncated normal distributions based on the Mersenne Twister PRNG. There is only one difference: to replace $\rvs_k$ with $\rvm_k$ in Equations~\ref{eq:uniform}, \ref{eq:normal}, and~\ref{eq:tnormal}, where $\rvm_k$ denotes a vector of draws from the Mersenne Twister PRNG with seed $k$.

\section{Weight Initializers}\label{sec:initializers}
\subsection{Parameters and Implementations}
Keras v.3.3.3 implements ten popular kernel initializers that use random number generators~\cite{keras_initializers}. Table~\ref{tab:initializers} list the initializers under study along with the default parameters (which we use in our study).

Glorot, He and Lecun (Uniform and Normal) adapt the shape of distribution by selecting parameters based on the number of input units, represented as $\Nin$ in the weight tensor, and/or the number of output units, represented as $\Nout$, in the weight tensor.

The creation of the Orthogonal initializer is more involved. The following is a quote from the Keras manual~\cite{keras_initializers_orthogonal}.
\begin{displayquote}
``If the shape of the tensor to initialize is two-dimensional, it is initialized with an orthogonal matrix obtained from the QR decomposition of a matrix of random numbers drawn from a normal distribution. If the matrix has fewer rows than columns then the output will have orthogonal rows. Otherwise, the output will have orthogonal columns.

If the shape of the tensor to initialize is more than two-dimensional, a matrix of shape \texttt{(shape[0] * ... * shape[n - 2], shape[n - 1])} is initialized, where \texttt{n} is the length of the shape vector. The matrix is subsequently reshaped to give a tensor of the desired shape.''
\end{displayquote}
A multiplicative factor denoted by $g$ is applied to the orthogonal matrix.

We use Keras/TensorFlow implementations of the algorithms as-is, but we modify the code to use QRNG rather than PRNG to sample the values from the respective distributions. Files  \verb|src/custom_initializers.py| and \verb|tests/test_custom_initializers.py| contain the code and associated test cases, respectively~\cite{source_code}.

\begin{table}[ht]
    \caption{Parameters of the initializers under study.}
    \label{tab:initializers}
    \centering
    \begin{tabular}{@{}lll@{}}
    \toprule
         Initializer & Distribution & Parameters  \\
    \midrule
         Glorot uniform   & $\uniform{(\cdot)}(a, b; k)$ & $a = - \sqrt{\frac{6}{\Nin+\Nout}}$, $b = \sqrt{\frac{6}{\Nin+\Nout}}$  \\ \\
         Glorot normal    & $\tnormal{(\cdot)}(\mu, \sigma; k)$ & $\mu=0$, $\sigma=\sqrt{\frac{2}{\Nin+\Nout}}$ \\ \\
         He uniform       & $\uniform{(\cdot)}(a, b; k)$ & $a = - \sqrt{\frac{6}{\Nin}}$, $b = \sqrt{\frac{6}{\Nin}}$ \\ \\
         He normal        & $\tnormal{(\cdot)}(\mu, \sigma; k)$ & $\mu=0$, $\sigma=\sqrt{\frac{2}{\Nin}}$ \\ \\
         Lecun uniform    & $\uniform{(\cdot)}(a, b; k)$ & $a = - \sqrt{\frac{3}{\Nin}}$, $b = \sqrt{\frac{3}{\Nin}}$ \\ \\
         Lecun normal     & $\tnormal{(\cdot)}(\mu, \sigma; k)$ & $\mu=0$, $\sigma=\sqrt{\frac{1}{\Nin}}$ \\ \\
         Orthogonal       & $\normal{(\cdot)}(\mu, \sigma; k)$ & $\mu=0$, $\sigma=1$, $g = 1$  \\ \\
         Random normal    & $\normal{(\cdot)}(\mu, \sigma; k)$ & $\mu=0, \sigma=0.05$ \\ \\
         Random uniform   & $\uniform{(\cdot)}(a, b; k)$ & $a=-0.05, b=0.05$ \\ \\
         Truncated normal & $\tnormal{(\cdot)}(\mu, \sigma; k)$ & $\mu=0, \sigma=0.05$\\
    \bottomrule
    \end{tabular}
\end{table}

\subsection{Auto-selection of seed}\label{sec:seed}
In our experiments, the seed $k$ is automatically selected using the function \verb| get_starting_dim_id_auto| in the file \verb|src/train_and_eval.py|. The files  \verb|src/global_config.py| and \verb|tests/test_global_config.py| contain the code and associated test cases, respectively~\cite{source_code}. 

\subsubsection{PRNG seed selection}\label{sec:prng_seed_selection}
For all models, PRNG-based initializers choose $k$ arbitrary for each layer of the model and experiment (which is a default Keras/TensorFlow behaviour). The seed selection for QRNG-based initializers is described below.

\subsubsection{QRNG seed selection}\label{sec:qrng_seed_selection}
In QRNG-based initializers, $k$ is chosen sequentially for every layer. As discussed in Appendix~\ref{sec:implementations}, the seed $k$ is mapped to the $k$-th dimension of the Sobol' sequences. We have observed that the sequence's starting value of $k$ may significantly affect the speed with which the optimizer reaches high accuracy values.

No universal pattern could be applied to all the models under study. However, we have seen that models that reach high levels of accuracy in early epochs usually retain this competitive advantage into the future. In addition, the number of seeds that lead to sub-par results is small.

As a result, we introduce a heuristic for selecting a seed for model training, as shown in Algorithm~\ref{alg:auto_seed_selection}. In essence, we randomly select $X$ seeds from the range $[W, Z]$ and train the model for $Y$ epochs, repeating\footnote{Because optimization is stochastic, different instances of the optimizer may converge to different solutions when starting from the same weights.} the training $R$ times. In order to match the range of dimensions used in the implementation of Sobol' sequences under study, the minimum value for $W, X,$ and $Z$ is \num{1} and the maximum value is \num{21200} (see Appendix~\ref{sec:sobol} for details).

This seed selection process can become expensive, since we must train the models for an additional $\Delta_{\qrng} = XYR - Y = Y(XR-1)$ epoch. The term $-Y$ refers to the fact that we can continue to train the best model, saving $Y$ epochs. Due to possible defects (associated with managing global states) in the underlying libraries, we do not do this in our code to minimize the risk of giving the QRNG-based model an unfair advantage. It is certainly reasonable to do this in a practical setting.

Our paper uses accuracy to measure performance, so we seek seeds that maximize accuracy. The following values yield adequate results for our use cases: $R =1, X =5, Y =1, W =1,$ and $Z =10$. Based on these parameters, QRNG-based initializers have a penalty of $\Delta_{\qrng} = Y(XR-1) =4$ epochs. Readers may adjust these values according to their use cases through empirical evaluation.

\begin{algorithm}[htb]
\SetKwComment{Comment}{/* }{ */}
\SetKwInOut{Input}{Input}
\SetKwInOut{Output}{Output}
\SetKw{KwOr}{or}
\SetKw{KwIn}{in}

\Input{$W \in \{1, 2, \ldots, 21200\}$ \Comment*[r]{The minimum value of the seed to try}} 
\Input{$Z \in \{1, 2, \ldots, 21200\}$ \Comment*[r]{The maximum value of the seed to try}} 
\Input{$X \in \{1, 2, \ldots, 21200\}$ \Comment*[r]{The number of seeds to try}} 
\Input{$Y \in \sZ_{\ge 1}$ \Comment*[r]{The number of epochs to train the model}} 
\Input{$R \in \sZ_{\ge 1}$ \Comment*[r]{The number of times to train a model with a particular seed}} 
\Input{\algvar{model\_cfg} \Comment*[r]{Model configuration}}
\Input{\algvar{train\_data} \Comment*[r]{Dataset to train the model}}
\Input{\algvar{test\_data} \Comment*[r]{Dataset to test the model}}
\Output{Suggested seed / dimension id denoted by $\nu$}
\BlankLine
$\nu \gets \varnothing $  \;
\algvar{best\_metric\_value} $\gets \varnothing$ \;
\algvar{seeds} $\gets$ Sample without replacement $X$ integers in the range $[W, Z]$ \;
\algvar{seeds} $\gets$ Sort \algvar{seeds} \Comment*[r]{Increase reproducibility when metric values are tied}
\ForEach{\algvar{seed} \KwIn \algvar{seeds}}{
    \For{$1$ \KwTo $R$}{
        Set starting seed value to \algvar{seed} \;
        \algvar{model} $\gets$ Initialize the \algvar{model} using \algvar{model\_cfg} \Comment*[r]{The seed value is assigned sequentially beginning at the starting seed value, i.e., \algvar{seed}}
        
        \algvar{trained\_model} $\gets$ Train the \algvar{model} for $Y$ epochs on \algvar{train\_data}  \;
        \algvar{current\_metric\_value} $\gets$  Evaluate \algvar{trained\_model} on \algvar{test\_data} and compute model's performance on \algvar{test\_data} \;
        \Comment{Assume that we want to maximize the metric}
        \If{best\_metric\_value $<$ current\_metric\_value \KwOr best\_metric\_value $= \varnothing $ }{
            \algvar{best\_metric\_value} $\gets$ \algvar{current\_metric\_value} \;
            $\nu \gets$ \algvar{seed} \Comment*[r]{New best seed}
        }
    }
}
\Return{$\nu$} \;
\caption{A heuristic to automatically select a starting seed value for a given model. }\label{alg:auto_seed_selection}
\end{algorithm}

Here are some examples of values of $k$ that we use in our models.

\paragraph{MLP and CNN} MLP and CNN models select a new value of $k$ for every layer. 
In QRNG-based initializers, $k$ is chosen sequentially for every layer starting at $\nu$ (the starting index is provided by the Algorithm~\ref{alg:auto_seed_selection}). As discussed in Appendix~\ref{sec:implementations}, the seed $k$ is mapped to $k$-th dimension of the Sobol' sequences.

For example, our MLP under study (described in Appendix~\ref{sec:mlp}) has three dense layers that require weight initalization. For the first layer, we will seed the initializer with $k=\nu$, the second layer~--- with $k=\nu+1$, and the third layer~--- with $k=\nu+2$. The weights are assigned to specific units of the layer without reshaping.

Our CNN under study (described in Appendix~\ref{sec:cnn}) has three convolutional and one dense layer. Three convolutional layers will be initialized with $k=\nu,\nu+1,\nu+2$, respectively; dense layer will be initialized with $k=\nu+3$. The tensor of the obtained weights will be reshaped to match the dimensionality of a layer (in terms of channels and kernels). Keras/TensorFlow standard code handles this reshaping for us.

\paragraph{LSTM and Transfomers}

LSTM and Transformer models consist of multiple submodules, each submodule is initialized with an incrementing value of $k$. In LSTM models, the four gates (input, forget, cell, and output) are treated as a single layer, and we experiment with different initializers while seeding QRNG-based methods with $k=\nu$; the same initializer is used for the dense layer (used for classification) with $k=\nu+1$. As described in Appendix~\ref{sec:lstm}, the recurrent states in the LSTM are initialized using the QRNG-based orthogonal initializer with $k=\nu$.

In Transformer models, the multi-head attention layer consists of four fully connected layers representing queries, keys, and values (QKV), followed by an output layer. The same initializer type is used for all layers. The QKV multi-head attention and associated output weights are initialized with $k=\nu, \nu+1,\nu+2,\nu+3$, respectively. Two additional fully connected layers follow the multi-head attention layer and are initialized with $k=\nu+4,\nu+5$, respectively.  The dense layer used for classification is initialized with $k=\nu+6$.

\section{Models}\label{sec:models}

The architecture of the models is shown below. As discussed in Section~\ref{sec:exp_design}, we keep the models simple to isolate the effect of random number generators on model performance.

\subsection{MLP}\label{sec:mlp}
\subsubsection{MLP for the performance experiments in Appendix~\ref{sec:single_layer_study}}\label{sec:single_layer_study_architecture}
\begin{figure}[htb]
    \centering
    \includegraphics[width=0.6\textwidth]{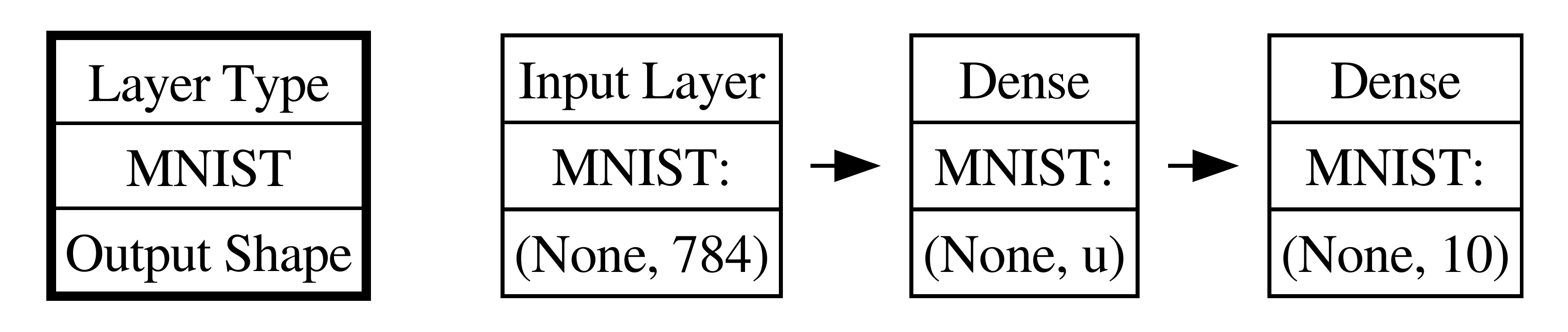}
    \caption{MLP architecture (used in the experiments in Appendix~\ref{sec:single_layer_study}). The number of units (neurons) $u$ varies from $1$ to $70$. The leftmost node represents a legend.}
    \label{fig:arch_motivational_mlp}
\end{figure}

Figure~\ref{fig:arch_motivational_mlp} shows an MLP architecture designed for image processing (input images are flattened). It has one fully connected layer. For a given experiment, the number of neurons in the layer varies from $1$ to $70$. ReLU (Rectified Linear Unit) activation functions~\cite{fukushima1969visual} are used after the layer. We experiment with different weight initialization techniques for kernels, and biases are initialized to zero. Details of the experiments are given in Appendix~\ref{sec:single_layer_study}. The source code for the model can be found in \verb|src/model/baseline_ann_one_layer.py|~\cite{source_code}.

The final output layer consists of a dense layer with a softmax activation function and ten neurons (one per class). The kernel initializer is Glorot Uniform (the default for Keras v~.3.3.3), and the biases are zero.

\subsubsection{MLP for core experiments}\label{sec:baseline_mlp}
\begin{figure}[htb]
    \centering
    \includegraphics[width=\textwidth]{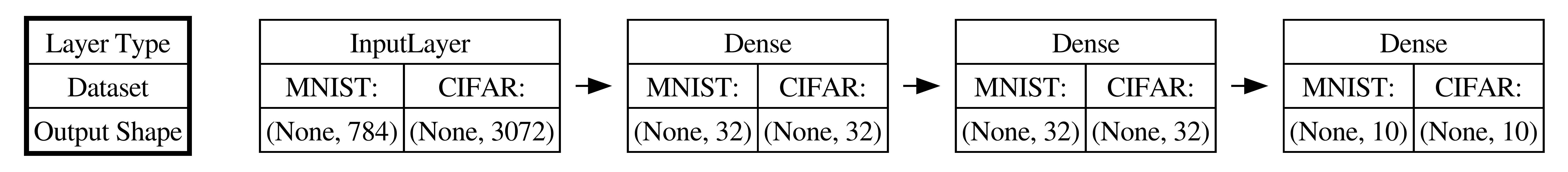}
    \caption{MLP architecture used in the core experiments. The leftmost node represents a legend.}
    \label{fig:arch_baseline_mlp}
\end{figure}

Figure~\ref{fig:arch_baseline_mlp} shows an MLP architecture designed for image processing (input images are flattened). It has two fully connected layers consisting of 32 neurons each. ReLU (Rectified Linear Unit) activation functions~\cite{fukushima1969visual} are employed after each layer. For kernels, we experiment with different weight initialization techniques (listed in Section~\ref{sec:methods}), and biases are initialized to zero. The source code for the model can be found in \verb|src/model/baseline_ann.py|~\cite{source_code}.

The final output layer consists of a dense layer with a softmax activation function and ten neurons (one per class). In each experiment, the kernel initializer is the same as the fully connected layers initializer, and biases are initialized to zero.

\subsection{CNN}\label{sec:cnn}
\begin{figure}[htb]
    \centering
    \includegraphics[width=\textwidth]{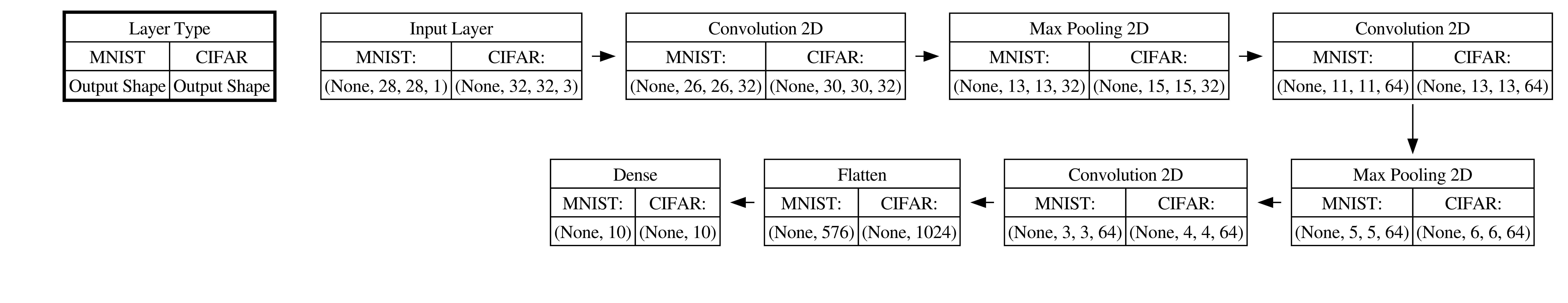}
    \caption{CNN architecture used in the core experiments. Top-left node represents a legend.}
    \label{fig:arch_baseline_cnn}
\end{figure}

Figure~\ref{fig:arch_baseline_cnn} pictures a CNN architecture designed for image processing. It consists of three convolutional layers, each with 32, 64, and 64 filters. The feature maps are downsampled using Max pooling layers, and ReLU activations are applied after each convolution layer. Similarly to MLP architectures, kernel initializers (shwon in Section~\ref{sec:methods}) differ based on experiment, while biases are initialized to zero. The source code for the model can be found in \verb|src/model/baseline_cnn.py|~\cite{source_code}.

The final output layer consists of a dense layer with a softmax activation function and ten neurons (one per class). In each experiment, the kernel initializer is the same as the convolutional layer initializer, and biases are initialized to zero.

\subsection{LSTM}\label{sec:lstm}
\begin{figure}[htb]
    \centering
    \includegraphics[width=\textwidth]{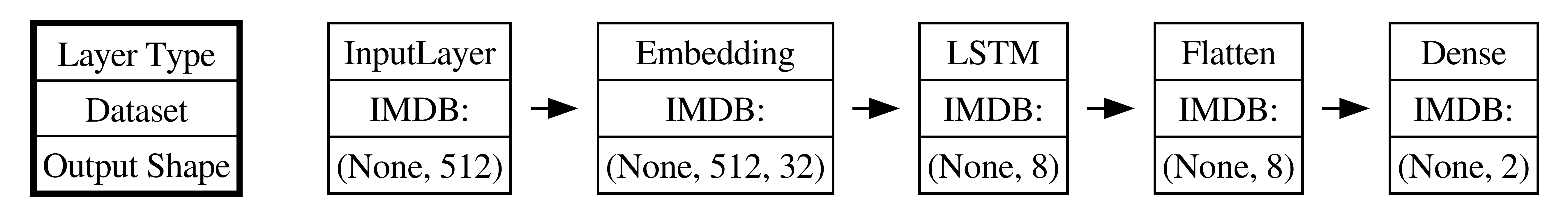}
    \caption{LSTM architecture used in the core experiments. The leftmost node represents a legend.}
    \label{fig:arch_baseline_lstm}
\end{figure}

Figure~\ref{fig:arch_baseline_lstm} depicts an LSTM architecture (a type of recurrent neural network) designed for classification. LSTM cells maintain four gates that constrain recurrent states. A gate consists of 8 neurons whose weights are initialized using an initializer (listed in Section~\ref{sec:methods}) specific to each experiment. Biases are initialized to zero. A recurrent weights matrix is populated using an orthogonal initializer which is initialized with the help of QRNG for QRNG experiments and with PRNG for PRNG experiments. Sigmoid activation function is use for the recurrent step.

The input data are passed through a 32-dimensional embedding layer. Default Keras v~.3.3.3 initialization scheme (PRNG-based) is used to initialize the layer.

The final output layer consists of a dense layer with a softmax activation function and two neurons (one per class). In each experiment, the kernel initializer is the same as the LSTM gates initializer, and biases are initialized to zero.

 The source code for the model can be found in \verb|src/model/baseline_lstm.py|~\cite{source_code}. 

\subsection{Transformer}
\begin{figure}[htb]
    \centering
    \includegraphics[width=\textwidth]{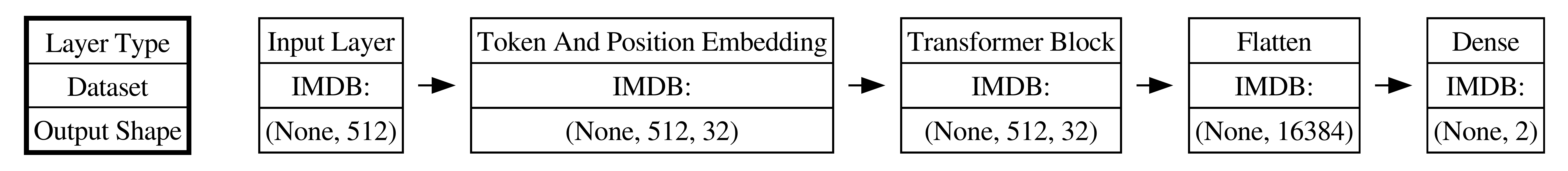}
    \caption{Transformer architecture used in the core experiments. The leftmost node represents a legend.}
    \label{fig:arch_baseline_transformer}
\end{figure}

Figure~\ref{fig:arch_baseline_transformer} displays an encoder-only Transformer architecture designed for classification.  To implement it, we use KerasNLP v.0.12.1 package~\cite{kerasnlp2022}. 

It is a simplified version of the architecture given in the Keras manual~\cite{nandan2020text}. It has two attention heads that take positional embedding vectors of size 32 for each token (the default Keras v~.3.3.3 PRNG-based scheme is used for initialization). Two fully connected layers are used, each consisting of 32 neurons and a dropout rate of 0.1. We experiment with different weight initialization techniques (depicted in Section~\ref{sec:methods}) for attention and fully connected layers while uniformly initializing positional embedding weights. Biases are initialized to zero. The source code for the model can be found in \verb|src/model/baseline_transformer.py|~\cite{source_code}.

The final output layer consists of a dense layer with a softmax activation function and two neurons (one per class). In each experiment, the kernel initializer is the same as the attention heads' initializer, and biases are initialized to zero.

The source code for the model can be found in \verb|src/model/baseline_transformer.py|~\cite{source_code}.

\section{Optimizers}\label{sec:optimizers}
Both Adam and SGD optimizers use the same learning rate of \num{0.0001}. The only exception was SGD-trained LSTM, where we saw no improvement in 30 epochs. For that case, we increased the learning rate to \num{0.01}.

In all experiments, the remaining hyperparameters of the optimizers are left at their default values, as set by Keras v.3.3.3. In particular, SGD momentum is set to \num{0}; Adam hyperparameters are $\beta_1=0.9$, $\beta_2= 0.999$, and $\hat \epsilon = 10^{-7}$.

\section{Datasets}\label{sec:datasets}

The source code for the data pipeline can be found in \verb|src/datapipeline.py|~\cite{source_code}. For all the datasets, labels are converted using one-hot encoding. To eliminate the risk of randomness from the data split process, we use complete training and test datasets provided by Keras unchanged (without splitting the train dataset into train and validation). The following are the details for each of the three datasets.

\subsection{MNIST}
The MNIST handwritten digit dataset~\cite{lecun1998gradient} is widely used in computer vision. It comprises a collection of $28 \times 28$ grayscale images of handwritten digits (0--9); the pixel values  are in the range of 0 to 255. In total, there are \num{60000} training images and \num{10000} testing images. The images are labelled with their corresponding digits, making them suitable for classification tasks.

The MNIST dataset is divided into two subsets: \num{60000} observations for training and \num{10000} for testing. This distribution follows a ratio of $6:1$, respectively.

\subsection{CIFAR-10}
Another popular dataset for image classification tasks is the CIFAR-10~\cite{krizhevsky2009learning}. It consists of \num{60000} $32 \times 32$ colour images (for each of the three channels, the pixel values range from 0 to 255) spread across ten different classes, with \num{6000} images per class. The dataset includes a variety of objects and animals, making it a more challenging dataset than MNIST. The detailed class list includes airplanes, cars, birds, cats, deer, dogs, frogs, horses, ships, and trucks.

The CIFAR dataset is split into \num{50000} observations for training and \num{10000} for testing, maintaining a ratio of $5:1$, respectively.

\subsection{IMDB}
The Large Movie Review Dataset~\cite{maas-EtAl:2011:ACL-HLT2011} also known as the IMDB dataset is a binary sentiment analysis dataset using IMDB reviews. Sequence modelling architectures, such as Recurrent Neural Networks and Transformers, are commonly benchmarked on this dataset. There are \num{50000} English movie reviews tagged with positive or negative sentiments.

The reviews are tokenized by word (as per \cite{maas-EtAl:2011:ACL-HLT2011}) using a vocabulary size of \num{20000} and a maximum review length of \num{512} tokens for all experiments. IMDB reviews with less than \num{512} tokens are zero-padded.  

For the IMDB dataset, the division includes \num{25000} observations for training and \num{25000} for testing, resulting in a ratio of $1:1$, respectively.

\section{Examples of computing measures of performance}\label{sec:example_performance_calculation}
\paragraph{Special case: level of accuracy not reached}\label{sec:special_case} 
When a model in a pair of models fails to reach a predetermined accuracy level $A$ within 30 epochs, we assign an epoch value of $31$ to $E_{(\cdot)}(A)$ of this model in Equation~\ref{eq:rel_epoch_diff}. This assignment offers a conservative approximation of $E(A)$. For example, if $E_{\qrng}(A) = 20$ and $E_{\prng}(A) > 30$, using $E_{\prng}(A) = 31$ yields $E(A) = (20-31)/31 \approx -35\%$, which should be interpreted as $E(A) \le -35\%$. In this scenario, $D(A)$ remains indeterminate.

\begin{figure}[htb]
    \centering
    \includegraphics[width=\textwidth]{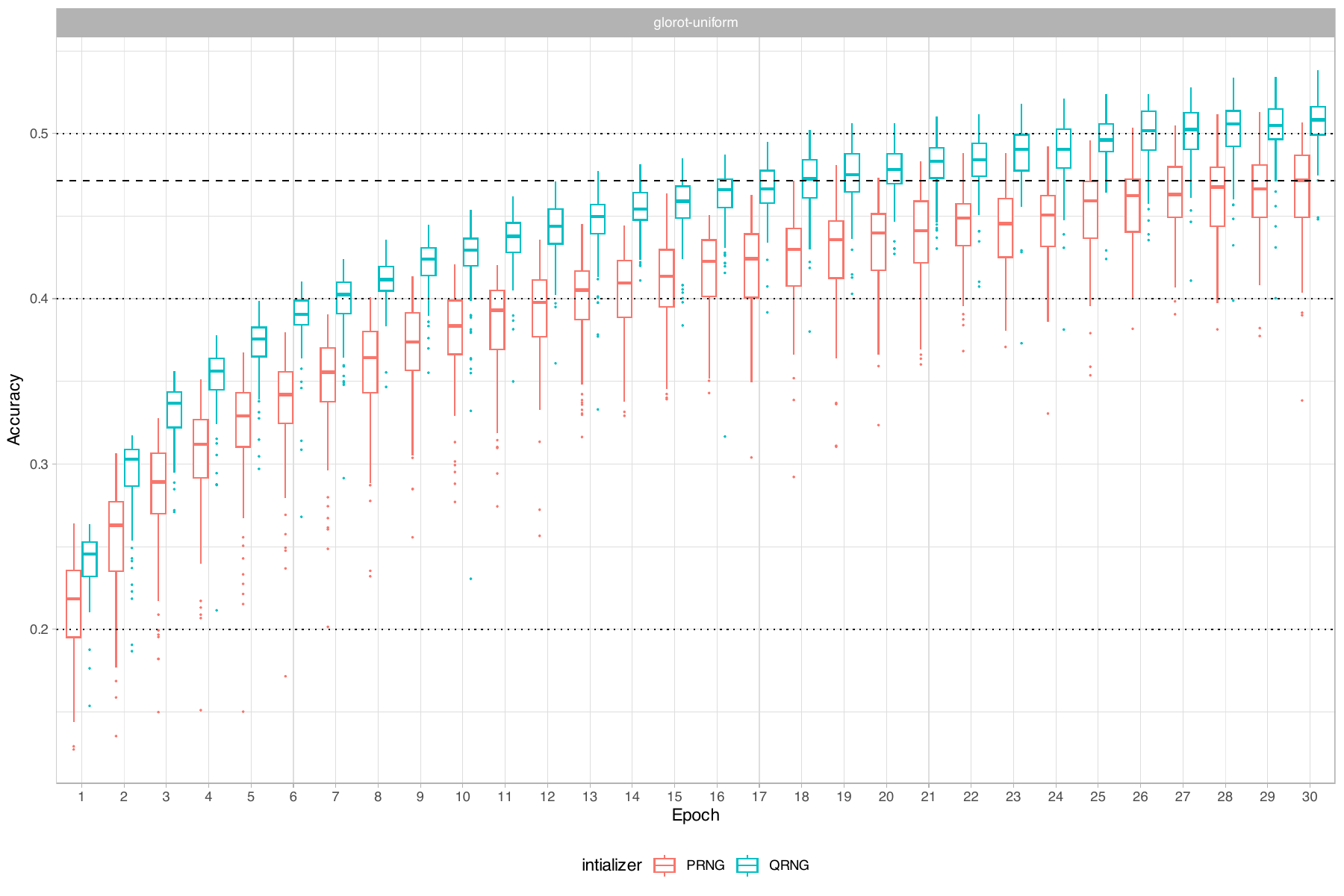}
    \caption{Accuracy distributions for CIFAR-10 test dataset, SGD optimizer, Glorot Uniform optimizer, and CNN model. Horizontal dotted lines indicate accuracy thresholds of $A=0.2, 0.3,$ and $0.4$. Horizontal dashed line indicated $A_m \approx 0.39$.}
    \label{fig:example_performance_calculation}
\end{figure}

\paragraph{Quantitative example}
Consider an example given in Figure~\ref{fig:example_performance_calculation}. Using box-and-whisker plots, we show the distribution of test accuracy for the \num{100} experiments. $x$-axis represents epoch number, $y$-axis represents model accuracy on test dataset.

Note that $E(A)$ and $D(A)$ for $A=0.05, 0.1, 0.15,$ and $0.2$ are identical since we reach $A \approx 0.2$ during the first epoch. Since neither model reaches $A \ge 0.55$, $E$ and $D$ for $A \ge 0.55$ are not computed. 

Let us examine three values of $A=0.2, 0.4, 0.5$. 

Both models reach $A=0.2$ at the first epoch. Based on Equation~\ref{eq:rel_epoch_diff},
\begin{equation*}
    E(0.2) = \frac{E_{\qrng}(0.2) + \Delta_{\qrng} - E_{\prng}(0.2)}{E_{\prng}(0.2)} \times 100 = \frac{1 + 4 -1}{1} \times 100=500\%
\end{equation*}
and, based on Equation~\ref{eq:iqr_diff},
\begin{equation*}
    D(0.2) = D_{\qrng}(0.2) - D_{\prng}(0.2) \approx 0.021 - 0.040 = -0.019. 
\end{equation*}

When $A=0.4$, the QRNG version of the initializer reaches this median accuracy at epoch $7$, while PRNG version at epoch $14$. Thus, 
\begin{equation*}
    E(0.4) = \frac{E_{\qrng}(0.4) + \Delta_{\qrng} - E_{\prng}(0.4)}{E_{\prng}(0.4)} \times 100 = \frac{7 + 4 - 13}{13} \times 100 \approx -15\%
\end{equation*}
and
\begin{equation*}
    D(0.4) = D_{\qrng}(0.4) - D_{\prng}(0.4) \approx 0.019 - 0.029 = -0.010. 
\end{equation*}

Finally, the QRNG version of the initializer reaches $A=0.5$ at epoch 23, while PRNG version does not reach this accuracy in 30 epochs. In this case, we set $E_{\prng}(0.5) = 31$ and the computation becomes:
\begin{equation*}
    E(0.5) \le \frac{E_{\qrng}(0.5) + \Delta_{\qrng} - E_{\prng}(0.5)}{E_{\prng}(0.5)} \times 100 = \frac{26+4-31}{31} \times 100 \approx -3\%
\end{equation*}
and
\begin{equation*}
    D(0.5) = D_{\qrng}(0.5) - D_{\prng}(0.5) \approx 0.023 - \text{indeterminate} = \text{indeterminate}. 
\end{equation*}

Figure~\ref{fig:example_performance_calculation} shows us that the maximum values of $A$ achieved by the model with the PRNG version of the initializer is $\approx 0.47$ and with the QRNG version of the initializer is $\approx 0.51$. Based on Equation~\ref{eq:A_m},
\begin{equation*}
    A_m = \min\left(A_{\prng}^{\max}, A_{\qrng}^{\max}\right) \approx \min\left(0.47, 0.51\right) = 0.47. 
\end{equation*}
This median accuracy is achieved by the PRNG version in epoch $30$ and by the QRNG version in epoch $18$. Thus,
\begin{equation*}
    E(A_M) = E(0.47) = \frac{E_{\qrng}(0.47) + \Delta_{\qrng} - E_{\prng}(0.47)}{E_{\prng}(0.47)} \times 100 = \frac{18+4-30}{30} \times 100 \approx -26\%
\end{equation*}
and
\begin{equation*}
    D(A_M) =D(0.47) = D_{\qrng}(0.47) - D_{\prng}(0.47) \approx 0.023 - 0.037 = -0.014. 
\end{equation*}

Based on Equation~\ref{eq:score_epoch_diff}, 
\begin{equation*}
    S_E[E(A_m)] = S_E[E(A_m)] = S_E[E(0.47)] = S_E[-26\%] = \text{win},
\end{equation*}
The $p$-value of the Fligner-Killeen test is $\approx 3.5 \times 10^{-4}$. Based on Equation~\ref{eq:score_iqr_diff},
\begin{equation*}
    S_D[D(A_m)] = S_D[D(A_m)] = S_D[D(0.47)] = S_D[-0.014] = \text{win}.
\end{equation*}

Thus, we can conclude that for the setup under study (CIFAR-10 test dataset, SGD optimizer, Glorot Uniform initializer, and CNN model), the model with QRNG-based initializer reaches the best accuracy than the model with PRNG-based initializer faster by $\approx 26\%$. Moreover, the QRNG-based model has less variation (in terms of IQRs) by $0.014$.

\section{The impact of QRNG and PRNG on a perceptron performance}\label{sec:single_layer_study}
We have demonstrated the effectiveness of QRNG by answering RQs~1--4. However, the underlying reasons for this effectiveness deserve further investigation. Let us examine a two-layer MLP model (shown in Figure~\ref{fig:arch_motivational_mlp}) to get an idea of how the random number initializer affects the model performance. In order to isolate the impact of a random number generator, the output layer initializer will remain the same (namely, the default choice for Keras and TensorFlow: PRNG-based Glorot Uniform initializer).

Five random number generators are examined for the hidden layer: two PRNGs (Philox and Mersenne Twister\footnote{See Appendix~\ref{sec:distr_prng_mt} for details on how Mersenne Twister is used to draw values from different distributions. }, discussed in Section~\ref{sec:rng_under_study}) and three QRNGs using Sobol' sequences. The number of units in this layer ranges from $1$ to $70$. The weights are initialized using one of the ten initializers under study (listed in Section~\ref{sec:methods}). 

The seeds for the Philox and Mersenne Twister PRNGs are chosen at random for each experiment (see Appendix~\ref{sec:prng_seed_selection}). In the three QRNG experiments, the starting seed is set at a constant value of $1$ or $5$, or it is automatically selected using the algorithm described in the Appendix~\ref{sec:qrng_seed_selection}.

Models are trained on the MNIST dataset for a single epoch using Adam optimizer. We repeat each experiment $100$ times to measure the variance in test accuracy. We use the median as a measure of central tendency and the IQR to assess dispersion, as these are less sensitive to outliers and skewness than the mean and standard deviation.

\subsection{PRNGs}\label{sec:single_layer_study_prng}
Figure~\ref{fig:1-layer-mlp-results} shows the results. For PRNG-based models, increasing the number of hidden layer units increases the test accuracy mainly monotonically. Both PRNG initialization methods produce similar median and IQR accuracy values. The effects of two PRNGs on the optimizer appear similar. Consequently, \textbf{the difference between the PRNG and QRNG results in RQs~1--4 cannot be attributed to Philox algorithm implementation problems.}

\subsection{QRNGs}\label{sec:single_layer_study_qrng}
Figure~\ref{fig:1-layer-mlp-results} suggests that models initialized with QRNGs show a different dynamics of the accuracy values that change with $u$ than models initialized with PRNGs. Specifically, QRNG-based models exhibits a sawtooth pattern, that (based on eyeballing of the figure) can be grouped depending on the odd- and even-numbered count of units $u$, where the even-numbered count of units is further partitioned into two sequences as follows:
\begin{equation}\label{eq:u_sequence}
    u\textrm{-sequence} = 
\begin{cases}
    \text{odd}, & \text{if } (u - 1) \bmod{2} = 0 \\
    \text{even: 2, 6, 10, \ldots}, & \text{if }  (u-2) \bmod{4} = 0\\
    \text{even: 4, 8, 12, \ldots}, & \text{if } u \bmod{4} = 0\\
\end{cases}.
\end{equation}

More research is needed to better understand why these three sequences of unit counts perform differently. Our empirical findings are as follows.

The three sequences yield similar accuracy at high $u$ values (probably because the effect of the QRNG is overpowered by the flexibility provided by the large number of weights at that stage). Among the smaller values of $u$, the odd values yield the lowest levels of accuracy, while $u=2, 6, 10, \ldots$ gives a higher level of accuracy, and $u=4, 8, 12, \ldots$~---~the highest level of accuracy. For example, Table~\ref{tbl:sample_accuracy_iqr_values} illustrates that for $u=33$, QRNG with automatic seed selection scheme for Random Uniform initializer generates a median accuracy of $0.69$, for $u=34$~--- $0.76$, and for $u=32$~--- $0.84$.

\begin{figure}[H]
    \centering
    \includegraphics[width=1\textwidth]{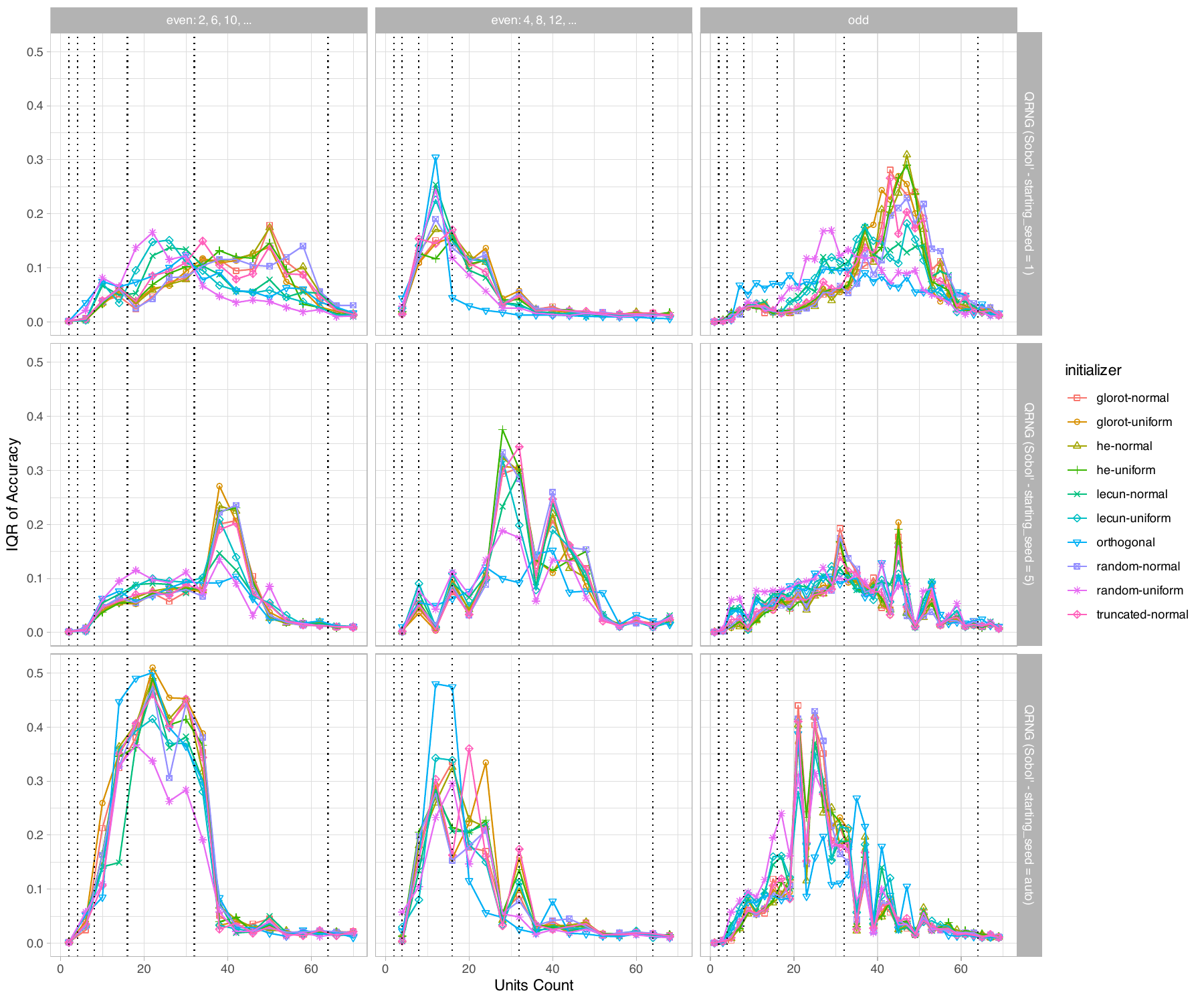}
    \caption{The median test accuracy for the perceptron-based model (given in Figure~\ref{fig:arch_motivational_mlp}) using three different QRNG-based initialization schemes for three sequences of $u$ (shown in Equation~\ref{eq:u_sequence}).  Vertical dotted lines show $u=2, 4, 8, 16, 32, 64$.}
    \label{fig:1-layer-mlp-results-iqr}
\end{figure}

Figures~\ref{fig:1-layer-mlp-results-median} and~\ref{fig:1-layer-mlp-results-iqr} show the median and IQR values of accuracy obtained by the models on a test dataset. We plot the data for the three $u$-sequences separately for better readability. For all sequences of $u$, the median accuracy increases (although non-monotonically), but the pattern of behavior for $u=4, 8, 12, \ldots$ is the least volatile (but IQR values may sometimes be higher). 

\subsubsection{PRNG vs. QRNG}

The difference in performance between models using different random number generators is sometimes striking.  For instance, Table~\ref{tbl:sample_accuracy_iqr_values} shows that the median accuracy using the Random Uniform initializer with 32 units is $0.55$ (IQR of $0.09$) for PRNG Philox and $0.84$ (IQR of $0.05$) for QRNG with automatic seed selection. In this case, QRNG offers $1.5$ times higher median accuracy and significantly less variability (by $0.04$).

\begin{table}[tb]
\centering
\caption{Median accuracy plus-minus IQR values for Random Uniform initializer and $u=31,32,33,34$.}
\label{tbl:sample_accuracy_iqr_values}
%\resizebox{\textwidth}{!}{%
\begin{tabular}{@{}lrrrr@{}}
\toprule
Random number generator                & $u=31$        & $u=32$        & $u=33$        & $u=34$        \\ \midrule
PRNG (Mersenne Twister)              & $0.54\pm0.11$ & $0.55\pm0.10$ & $0.55\pm0.09$ & $0.58\pm0.11$ \\
PRNG (Philox)                        & $0.53\pm0.08$ & $0.55\pm0.09$ & $0.56\pm0.10$ & $0.57\pm0.09$ \\
QRNG (Sobol' --- starting seed = 1)    & $0.40\pm0.12$ & $0.87\pm0.03$ & $0.47\pm0.13$ & $0.70\pm0.07$  \\
QRNG (Sobol' --- starting seed = 5)    & $0.54\pm0.10$ & $0.67\pm0.18$ & $0.57\pm0.10$ & $0.63\pm0.07$ \\
QRNG (Sobol' --- starting seed = auto) & $0.69\pm0.18$ & $0.84\pm0.05$ & $0.69\pm0.17$ & $0.76\pm0.19$ \\ \bottomrule
\end{tabular}%
%}
\end{table}

\subsubsection{QRNG seed selection methods}
When comparing three QRNG initialization methods, the automatic starting seed selection method results in the least variability in precision with $u$ (compare the accuracy values in Table~\ref{tbl:sample_accuracy_iqr_values}). 

However, for $u=4, 8, 12, \ldots$, QRNG-based models with a starting seed value of 1 slightly outperform the automatic method. For instance, Table~\ref{tbl:sample_accuracy_iqr_values} shows that a starting seed of 1 yields an accuracy of $0.87 \pm 0.03$, compared to $0.84 \pm 0.05$ for the automatic method. Conversely, a starting seed of 5 yields an accuracy of only $0.68 \pm 0.18$. 

When $u$ is odd or $u=2, 6, 10, \ldots$, automatic seed selection outperforms the constant seeds 1 and 5. For instance, for $u=33$, QRNG-based models using seed 1 and 5 achieve accuracies of $0.47\pm0.13$ and $0.57\pm0.10$, respectively. These values are comparable to or lower than those produced of PRNG (Philox), which yields an accuracy of $0.56\pm0.10$. In contrast, the QRNG-based model with automatic seed selection achieves an accuracy of $0.69\pm0.17$, which is $1.2$ times better than that of PRNG (Philox).

This suggests that \textbf{seed selection is crucial and that our heuristic for seed selection may need further refinement.}

\subsubsection{QRNG sequence length}
\begin{figure}[H]
    \centering
    \includegraphics[width=1\textwidth]{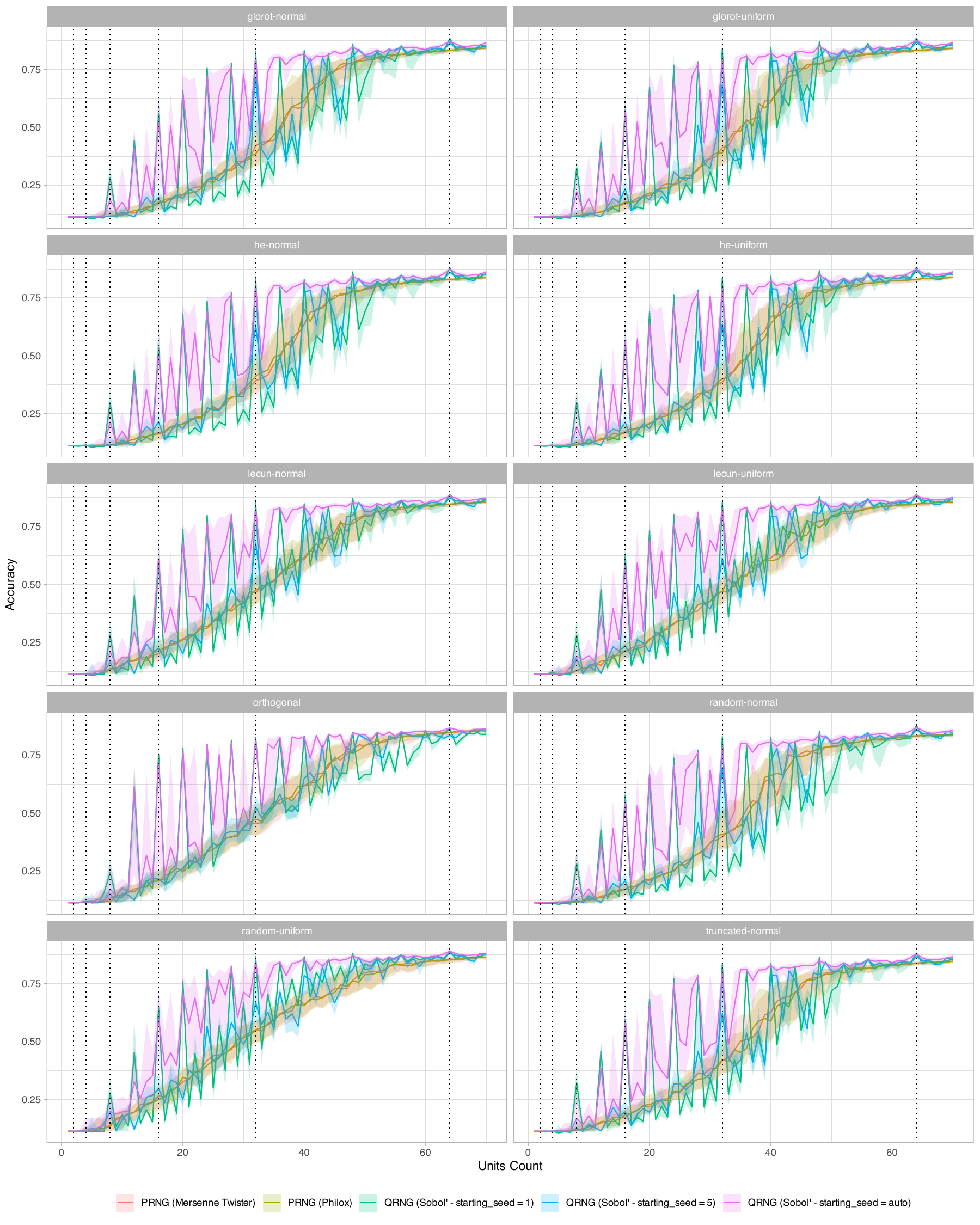}
    \caption{The median test accuracy for the perceptron-based model (given in Figure~\ref{fig:arch_motivational_mlp}). The $x$-axis shows the number of units $u$ in the model. The $y$-axis represents the model's accuracy on the MNIST test dataset. The lines represent median accuracy based on 100 repetitions, while the ribbons indicate the range between the lower and upper quartiles. Vertical dotted lines show $u=2, 4, 8, 16, 32, 64$.}
    \label{fig:1-layer-mlp-results}
\end{figure}

\begin{figure}[H]
    \centering
    \includegraphics[width=1\textwidth]{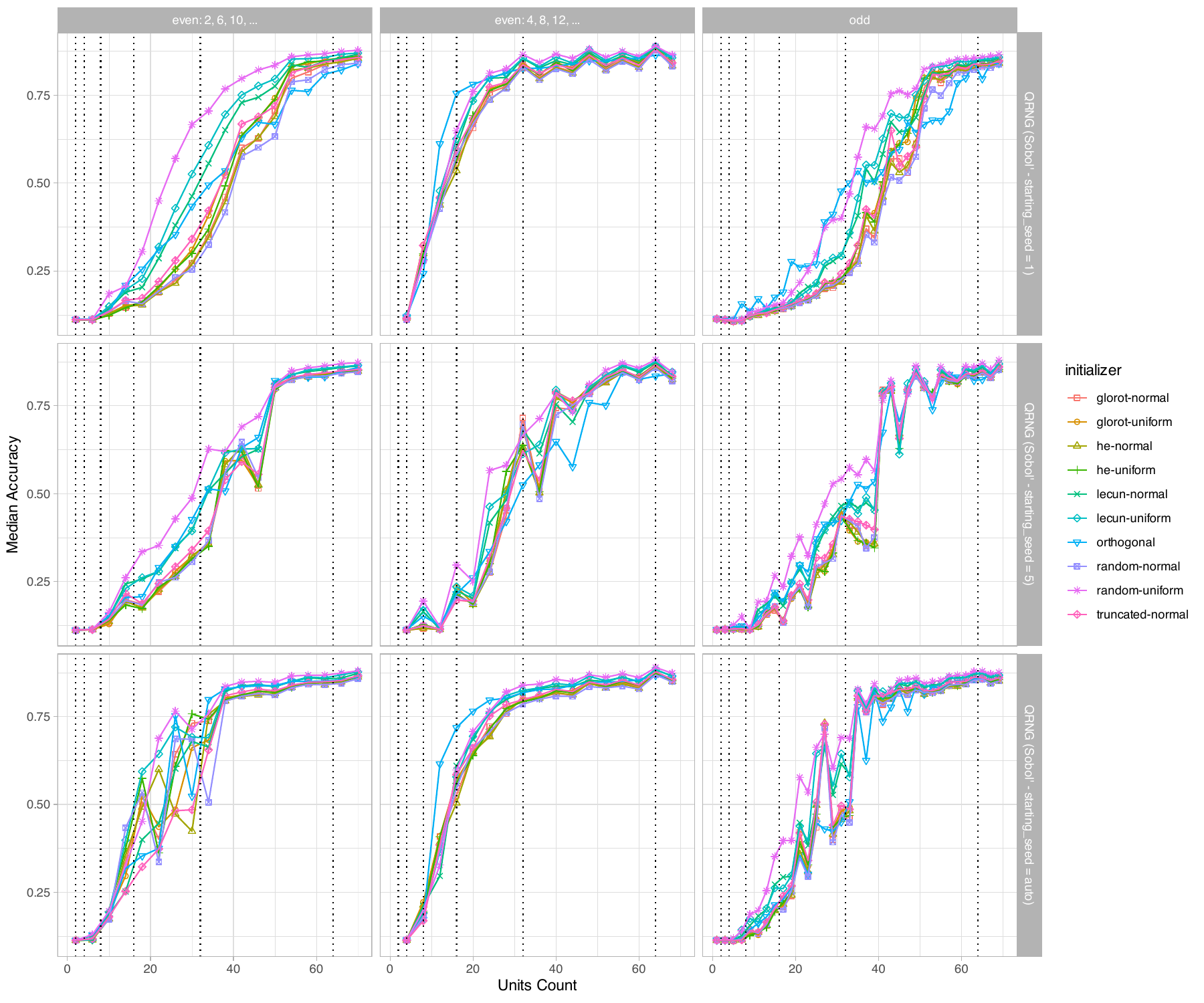}
    \caption{The median test accuracy for the perceptron-based model (given in Figure~\ref{fig:arch_motivational_mlp}) using three different QRNG-based initialization schemes for three sequences of $u$ (shown in Equation~\ref{eq:u_sequence}). Vertical dotted lines show $u=2, 4, 8, 16, 32, 64$.}
    \label{fig:1-layer-mlp-results-median}
\end{figure}

Finally, if the sequence length is not a power of two, Sobol' sequences lose their balance properties, which is detrimental to Monte Carlo-based methods~\cite{owen2022dropping}. However, Figures~\ref{fig:1-layer-mlp-results} and~\ref{fig:1-layer-mlp-results-median}  show no dramatic deviations in accuracy near $u=1, 2, 4, 8, \ldots, 64$ (compared to the closest value $u$ in the corresponding sequence). Although some effect may be present (e.g., at $u=32$), from a practical perspective, this implies that \textbf{we can use the sequences for weight initialization even if the number of samples is not a power of two.}

\section{Summary statistics}\label{sec:summary_stats}

\begin{figure}[H]
    \centering
    \includegraphics[width=\textwidth]{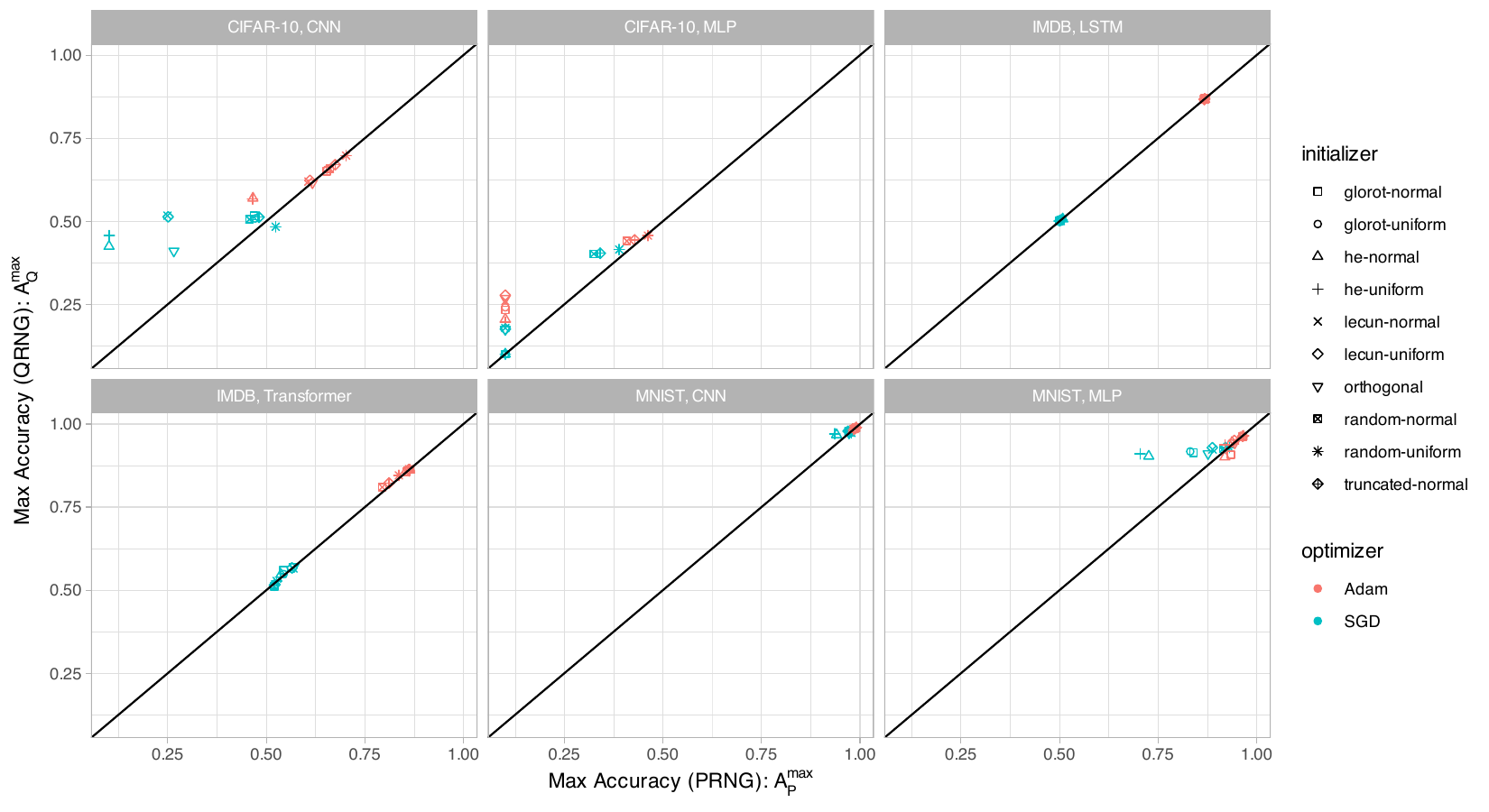}
    \caption{$A_\prng^{\max}$ and $A_\qrng^{\max}$ values needed to compute $S_A$ for all 120 cases. Black line is governed by equation $A_\qrng^{\max} = A_\prng^{\max}$.}
    \label{fig:ap_vs_aq}
\end{figure}

\begin{figure}[H]
    \centering
    \includegraphics[width=\textwidth]{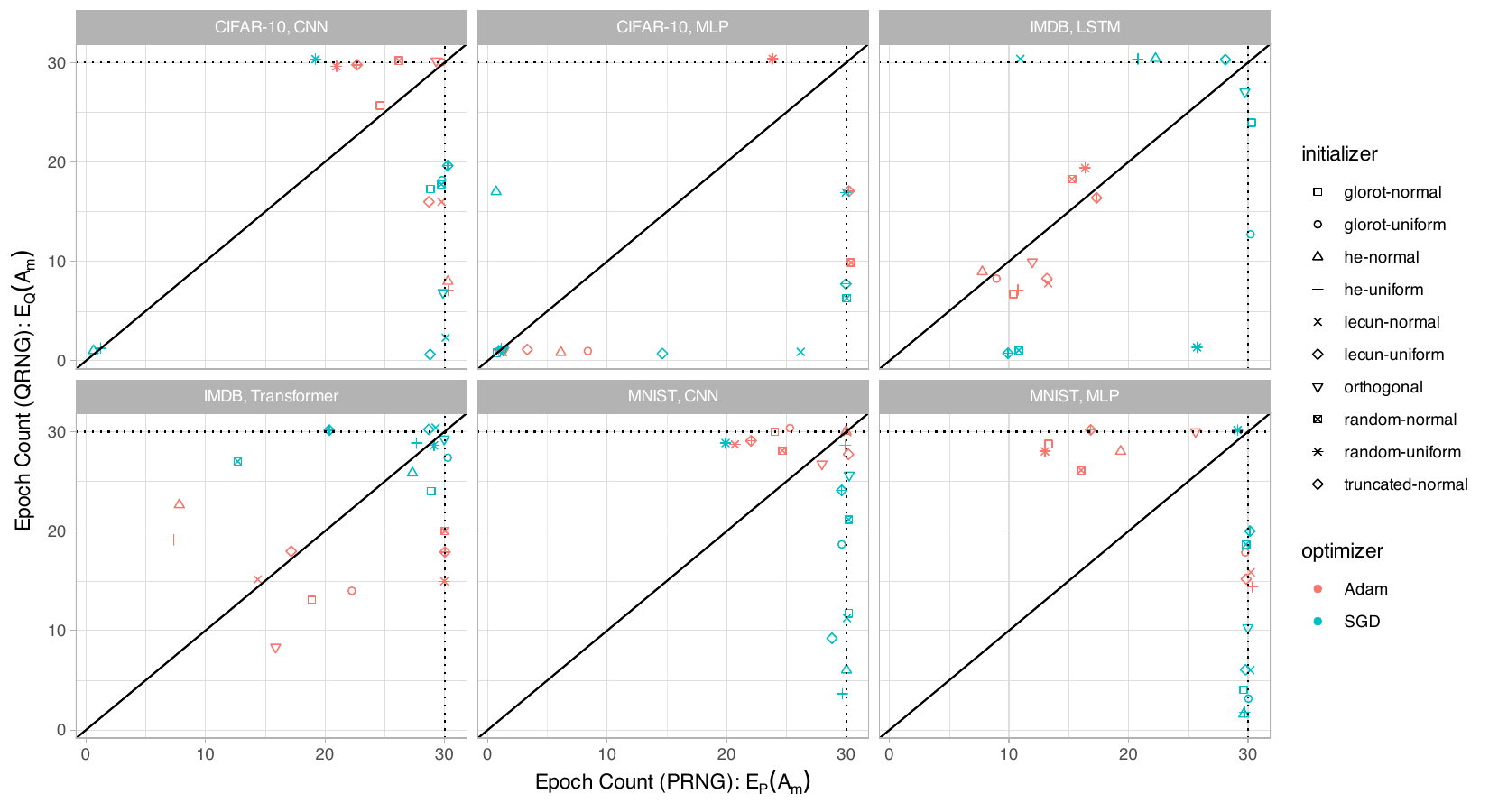}
    \caption{$E_\prng(A_m)$ and $E_\qrng(A_m)$ values needed to compute $E(A_m)$ for all 120 cases. Black line is governed by equation $E_{\qrng}(A) = E_{\prng}(A)$. To improve plot readability, $E_\prng(A_m)$ and $E_\qrng(A_m)$ values are jittered. Horizontal and vertical dotted lines represent 30 epochs.}
    \label{fig:ep_vs_eq}
\end{figure}

\begin{figure}[H]
    \centering
    \includegraphics[width=\textwidth]{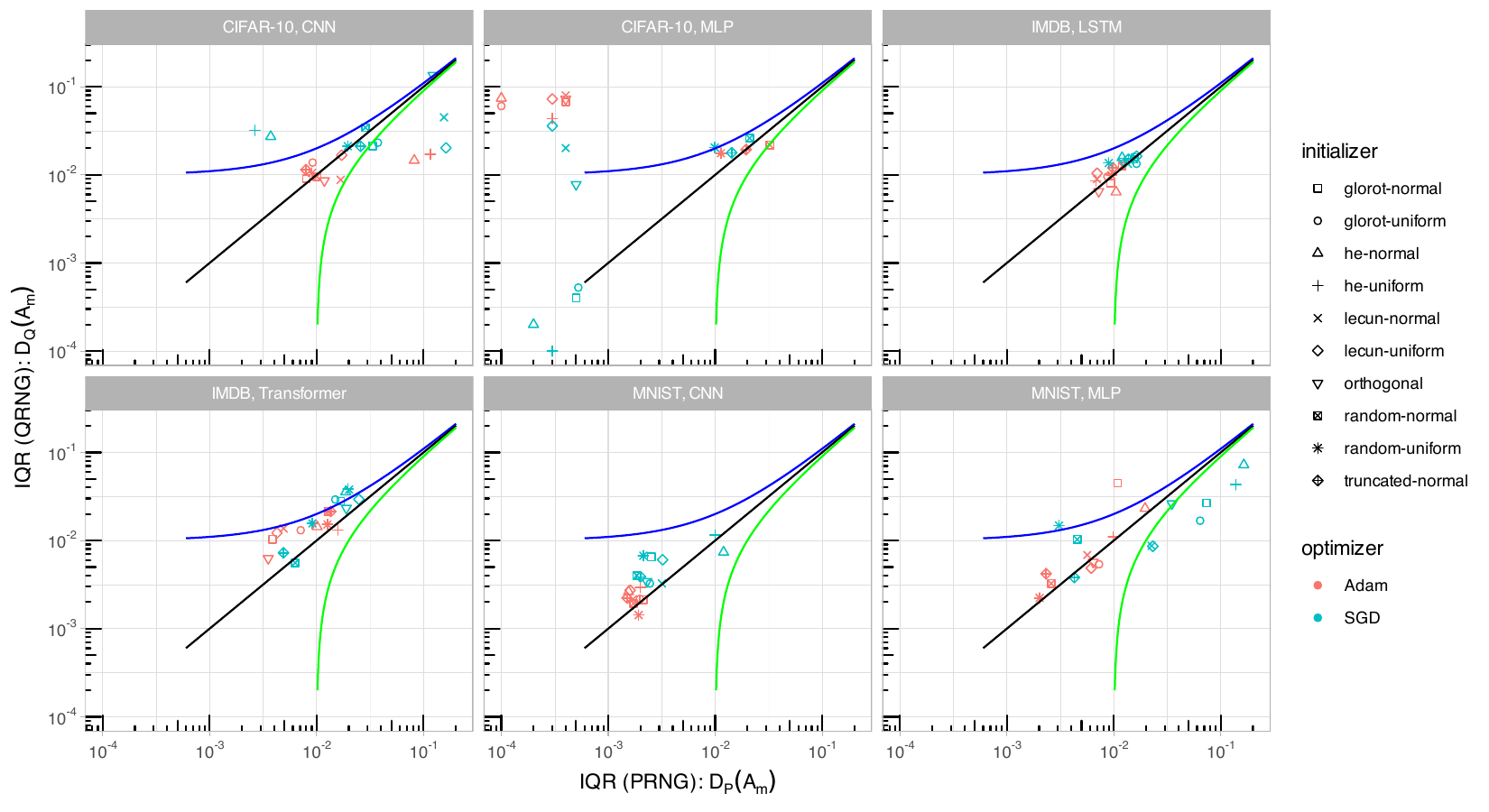}
    \caption{$D_\prng(A_m)$ and $D_\qrng(A_m)$ values needed to compute $D(A_m)$ for all 120 cases. Black line is governed by equation $D(A) = D_{\qrng}(A) - D_{\prng}(A) = 0$; blue line is governed by $D(A) = D_{\qrng}(A) - D_{\prng}(A) = 0.01$; green line is governed by $D(A) = D_{\qrng}(A) - D_{\prng}(A) = -0.01$. }
    \label{fig:dp_vs_dq_all}
\end{figure}

The values of $A_\prng^{\max}$ and $A_\qrng^{\max}$ required to compute $S_A$ for all cases are shown graphically in Figure~\ref{fig:ap_vs_aq}. The values of $E_\prng(A_m)$ and $E_\qrng(A_m)$, necessary for calculating $E(A_m)$, are presented in Figure~\ref{fig:ep_vs_eq}. The values of $D_{\prng}(A_m)$ and $D_{\qrng}(A_m)$, used to determine $D(A_m)$, are illustrated in Figure~\ref{fig:dp_vs_dq_all}.
The average values of $A_{\qrng}^{\max} - A_{\prng}^{\max}$, denoted by $\mean{\alpha}$, grouped by optimizer, dataset, and model, are shown in Table~\ref{tbl:alpha_optmizer_dataset_model}; grouped by initializer~---~in Table~\ref{tbl:alpha_initializer};  grouped by model and optimizer~---~in Table~\ref{tbl:alpha_model_optimizer};  and grouped by dataset and optimizer~---~in Table~\ref{tbl:alpha_dataset_optimizer}  . Finally, the average $E(A_m)$ values, denoted by $\mean{E}(A_m)$, grouped by model and optimizer are given in Table~\ref{tbl:e_avg_model_optimizer}.

\begin{landscape}
\begin{table}[tb]
\caption{Average $A_{\qrng}^{\max} - A_{\prng}^{\max}$, denoted by $\mean{\alpha}$, grouped by optimizer, dataset, and model.}
\label{tbl:alpha_optmizer_dataset_model}
\centering
\resizebox{1.35\textheight}{!}{%
\begin{tabular}{@{}l|rr|r|rr|r|rr|r|r|rr|r|rr|r|rr|r|r|r@{}}
\toprule
          & \multicolumn{9}{c|}{Adam}                                                                                       & Adam  & \multicolumn{9}{c|}{SGD}                                                                                        & SGD   & Grand \\
          \cmidrule(lr){2-10} \cmidrule(lr){12-20} 
Outcome          & \multicolumn{2}{c|}{CIFAR-10} & CIFAR-10 & \multicolumn{2}{c|}{IMDB} & IMDB  & \multicolumn{2}{c|}{MNIST} & MNIST & Total & \multicolumn{2}{c|}{CIFAR-10} & CIFAR-10 & \multicolumn{2}{c|}{IMDB} & IMDB  & \multicolumn{2}{c|}{MNIST} & MNIST & Total & Total \\
          \cmidrule(lr){2-3} \cmidrule(lr){5-6} \cmidrule(lr){8-9} \cmidrule(lr){12-13} \cmidrule(lr){15-16} \cmidrule(lr){18-19}    
          & CNN           & MLP          & Total    & LSTM    & Transf.    & Total & CNN         & MLP         & Total &       & CNN           & MLP          & Total    & LSTM    & Transf.    & Total & CNN         & MLP         & Total &       &       \\ 
\midrule
\textsc{l:l,*,*}  & 0.00          & 0.00         & 0.00     & 0.00    & 0.00           & 0.00  & 0.00        & -0.01       & -0.01 & 0.00  & -0.04         &              & -0.04    &         & -0.01          & -0.01 & 0.00        &             & 0.00  & -0.01 & -0.01 \\
\textsc{l:t,l,*}  & 0.00          &              & 0.00     & 0.00    & 0.00           & 0.00  & 0.00        & 0.00        & 0.00  & 0.00  &               & 0.00         & 0.00     & 0.00    & 0.00           & 0.00  &             & 0.00        & 0.00  & 0.00  & 0.00  \\ \midrule
Loss Total          & 0.00          & 0.00         & 0.00     & 0.00    & 0.00           & 0.00  & 0.00        & -0.01       & 0.00  & 0.00  & -0.04         & 0.00         & -0.01    & 0.00    & 0.00           & 0.00  & 0.00        & 0.00        & 0.00  & 0.00  & 0.00  \\ \midrule
\textsc{t:t,t,t}  &               &              &          & 0.00    &                & 0.00  &             &             &       & 0.00  &               &              &          &         &                &       &             &             &       &       & 0.00  \\ \midrule
Tie Total          &               &              &          & 0.00    &                & 0.00  &             &             &       & 0.00  &               &              &          &         &                &       &             &             &       &       & 0.00  \\ \midrule
\textsc{w:t,w,wt} &               &              &          & 0.00    & 0.00           & 0.00  &             &             &       & 0.00  &               &              &          & 0.00    &                & 0.00  &             &             &       & 0.00  & 0.00  \\
\textsc{w:w,l,*}  &               & 0.15         & 0.15     &         & 0.00           & 0.00  & 0.00        &             & 0.00  & 0.09  & 0.34          & 0.02         & 0.15     &         & 0.01           & 0.01  &             &             &       & 0.10  & 0.09  \\
\textsc{w:w,t,wt} &               &              &          &         &                &       &             &             &       &       &               &              &          &         &                &       & 0.00        &             & 0.00  & 0.00  & 0.00  \\
\textsc{w:w,w,l}  &               & 0.12         & 0.12     &         &                &       &             &             &       & 0.12  & 0.14          & 0.06         & 0.08     &         & 0.02           & 0.02  &             &             &       & 0.07  & 0.08  \\
\textsc{w:w,w,wt} & 0.06          & 0.02         & 0.05     &         & 0.01           & 0.01  &             & 0.01        & 0.01  & 0.03  & 0.12          & 0.07         & 0.10     &         &                &       & 0.01        & 0.07        & 0.05  & 0.06  & 0.05  \\ \midrule
Win Total          & 0.06          & 0.11         & 0.10     & 0.00    & 0.01           & 0.00  & 0.00        & 0.01        & 0.01  & 0.05  & 0.17          & 0.05         & 0.11     & 0.00    & 0.01           & 0.00  & 0.01        & 0.07        & 0.04  & 0.06  & 0.06  \\ \midrule
Grand Total          & 0.02          & 0.10         & 0.06     & 0.00    & 0.00           & 0.00  & 0.00        & 0.00        & 0.00  & 0.02  & 0.15          & 0.04         & 0.09     & 0.00    & 0.00           & 0.00  & 0.01        & 0.07        & 0.04  & 0.04  & 0.03 \\
\bottomrule
\end{tabular}%
}
\end{table}

\begin{table}[tb]
\caption{Average $A_{\qrng}^{\max} - A_{\prng}^{\max}$, denoted by $\mean{\alpha}$, grouped by initializer.}
\label{tbl:alpha_initializer}
\centering
\resizebox{1.2\textheight}{!}{%
\begin{tabular}{@{}l|rrrrrr|r|rrr|r|r|r@{}}
\toprule
            & \multicolumn{6}{c|}{Shape dependent}                                                   & Shape & \multicolumn{3}{c|}{Shape agnostic}            & Shape & Orthogonal & Grand \\
            \cmidrule(lr){2-7} \cmidrule(lr){9-11} 
Outcome            & Glorot & Glorot & He & He & Lecun & Lecun & dependent            & Random & Random & Truncated & agnostic              & Total & Total \\
            & normal & uniform & normal & uniform & normal & uniform & Total            & normal & uniform & normal & Total              &  &  \\    
\midrule
\textsc{l:l,*,*}    & -0.01         & 0.00           & -0.01     & 0.00       &              &               & -0.01            & 0.00          & -0.01          & 0.00             & -0.01              &            & -0.01 \\
\textsc{l:t,l,*}    & 0.00          & 0.00           & 0.00      & 0.00       & 0.00         & 0.00          & 0.00             &               & 0.00           & 0.00             & 0.00               & 0.00       & 0.00  \\
\midrule
Loss Total  & -0.01         & 0.00           & 0.00      & 0.00       & 0.00         & 0.00          & 0.00             & 0.00          & -0.01          & 0.00             & 0.00               & 0.00       & 0.00  \\
\midrule
\textsc{t:t,t,t}    &               &                &           & 0.00       &              &               & 0.00             &               &                &                  &                    &            & 0.00  \\
\midrule
Tie Total   &               &                &           & 0.00       &              &               & 0.00             &               &                &                  &                    &            & 0.00  \\
\midrule
\textsc{w:t,w,wt}   & 0.00          & 0.00           &           &            & 0.00         & 0.00          & 0.00             & 0.00          & 0.00           & 0.00             & 0.00               & 0.00       & 0.00  \\
\textsc{w:w,l,*}    & 0.07          & 0.00           & 0.17      & 0.15       & 0.08         & 0.09          & 0.10             &               &                &                  &                    & 0.08       & 0.09  \\
\textsc{w:w,t,wt}   &               &                &           &            &              &               &                  &               &                &                  &                    & 0.00       & 0.00  \\
\textsc{w:w,w,l}    & 0.02          & 0.14           & 0.11      &            & 0.08         & 0.08          & 0.08             &               & 0.03           &                  & 0.03               & 0.14       & 0.08  \\
\textsc{w:w,w,wt}   & 0.04          & 0.03           & 0.10      & 0.09       & 0.07         & 0.07          & 0.06             & 0.03          & 0.01           & 0.02             & 0.02               & 0.03       & 0.05  \\ \midrule
Win Total   & 0.04          & 0.03           & 0.12      & 0.12       & 0.06         & 0.06          & 0.07             & 0.03          & 0.01           & 0.02             & 0.02               & 0.06       & 0.06  \\ \midrule
Grand Total & 0.02          & 0.02           & 0.06      & 0.07       & 0.05         & 0.05          & 0.04             & 0.01          & 0.00           & 0.01             & 0.01               & 0.04       & 0.03 \\
\bottomrule
\end{tabular}%
}
\end{table}

\end{landscape}

\begin{table}[tb]
\caption{Average $A_{\qrng}^{\max} - A_{\prng}^{\max}$, denoted by $\mean{\alpha}$, grouped by model and optimizer.}
\label{tbl:alpha_model_optimizer}
\resizebox{\textwidth}{!}{%
\begin{tabular}{@{}l|rr|r|rr|r|rr|r|rr|r|r@{}}
\toprule
Outcome               & \multicolumn{2}{c|}{CNN}    & CNN   & \multicolumn{2}{c|}{LSTM} & LSTM  & \multicolumn{2}{c|}{MLP} & MLP   & \multicolumn{2}{c|}{Transformer} & Transformer & Grand \\
\cmidrule(lr){2-3}\cmidrule(lr){5-6}\cmidrule(lr){8-9}\cmidrule(lr){11-12}
                      & Adam          & SGD        & Total & Adam        & SGD        & Total & Adam        & SGD       & Total & Adam           & SGD            & Total       & Total \\
\midrule
\textsc{l:l,*,*}      & 0.00          & -0.02      & -0.01 & 0.00        &            & 0.00  & -0.01       &           & -0.01 & 0.00           & -0.01          & -0.01       & -0.01 \\
\textsc{l:t,l,*}      & 0.00          &            & 0.00  & 0.00        & 0.00       & 0.00  & 0.00        & 0.00      & 0.00  & 0.00           & 0.00           & 0.00        & 0.00  \\
\midrule
Loss Total            & 0.00          & -0.02      & 0.00  & 0.00        & 0.00       & 0.00  & -0.01       & 0.00      & -0.01 & 0.00           & 0.00           & 0.00        & 0.00  \\
\midrule
\textsc{t:t,t,t} &  &            &       & 0.00        &            & 0.00  &             &           &       &                &                &             & 0.00  \\
\midrule
Tie Total             &               &            &       & 0.00        &            & 0.00  &             &           &       &                &                &             & 0.00  \\
\midrule
\textsc{w:t,w,wt} &&            &       & 0.00        & 0.00       & 0.00  &             &           &       & 0.00           &                & 0.00        & 0.00  \\
\textsc{w:w,l,*}      & 0.00          & 0.34       & 0.17  &             &            &       & 0.15        & 0.02      & 0.10  & 0.00           & 0.01           & 0.01        & 0.09  \\
\textsc{w:w,t,wt} & & 0.00       & 0.00  &             &            &       &             &           &       &                &                &             & 0.00  \\
\textsc{w:w,w,l} &  & 0.14       & 0.14  &             &            &       & 0.12        & 0.06      & 0.09  &                & 0.02           & 0.02        & 0.08  \\
\textsc{w:w,w,wt}     & 0.06          & 0.06       & 0.06  &             &            &       & 0.01        & 0.07      & 0.05  & 0.01           &                & 0.01        & 0.05  \\
\midrule
Win Total             & 0.04          & 0.09       & 0.08  & 0.00        & 0.00       & 0.00  & 0.08        & 0.06      & 0.07  & 0.01           & 0.01           & 0.01        & 0.06  \\
\midrule
Grand Total           & 0.01          & 0.08       & 0.04  & 0.00        & 0.00       & 0.00  & 0.05        & 0.05      & 0.05  & 0.00           & 0.00           & 0.00        & 0.03 \\
\bottomrule
\end{tabular}%
}
\end{table}

\begin{table}[tb]
\caption{Average $A_{\qrng}^{\max} - A_{\prng}^{\max}$, denoted by $\mean{\alpha}$, grouped by dataset and optimizer.}
\label{tbl:alpha_dataset_optimizer}
\resizebox{\textwidth}{!}{%
\begin{tabular}{@{}l|rr|r|rr|r|rr|r|r@{}}
\toprule
Outcome           & \multicolumn{2}{c|}{CIFAR-10} & CIFAR-10 & \multicolumn{2}{c|}{IMDB} & IMDB  & \multicolumn{2}{c|}{MNIST} & MNIST & Grand  \\
\cmidrule(lr){2-3}\cmidrule(lr){5-6}\cmidrule(lr){8-9}
                  & Adam         & SGD           & Total    & Adam       & SGD         & Total & Adam         & SGD        & Total &  Total           \\
                  \midrule
\textsc{l:l,*,*}  & 0.00         & -0.04         & -0.01    & 0.00       & -0.01       & 0.00  & -0.01        & 0.00       & -0.01 & -0.01       \\
\textsc{l:t,l,*}  & 0.00         & 0.00          & 0.00     & 0.00       & 0.00        & 0.00  & 0.00         & 0.00       & 0.00  & 0.00        \\ \midrule
Loss Total        & 0.00         & -0.01         & -0.01    & 0.00       & 0.00        & 0.00  & 0.00         & 0.00       & 0.00  & 0.00        \\ \midrule
\textsc{t:t,t,t}  &              &               &          & 0.00       &             & 0.00  &              &            &       & 0.00        \\ \midrule
Tie Total         &              &               &          & 0.00       &             & 0.00  &              &            &       & 0.00        \\ \midrule
\textsc{w:t,w,wt} &              &               &          & 0.00       & 0.00        & 0.00  &              &            &       & 0.00        \\
\textsc{w:w,l,*}  & 0.15         & 0.15          & 0.15     & 0.00       & 0.01        & 0.01  & 0.00         &            & 0.00  & 0.09        \\
\textsc{w:w,t,wt} &              &               &          &            &             &       &              & 0.00       & 0.00  & 0.00        \\
\textsc{w:w,w,l}  & 0.12         & 0.08          & 0.10     &            & 0.02        & 0.02  &              &            &       & 0.08        \\ 
\textsc{w:w,w,wt} & 0.05         & 0.10          & 0.08     & 0.01       &             & 0.01  & 0.01         & 0.05       & 0.04  & 0.05        \\\midrule
Win Total         & 0.10         & 0.11          & 0.11     & 0.00       & 0.00        & 0.00  & 0.01         & 0.04       & 0.03  & 0.06        \\ \midrule
Grand Total       & 0.06         & 0.09          & 0.08     & 0.00       & 0.00        & 0.00  & 0.00         & 0.04       & 0.02  & 0.03     \\
\bottomrule
\end{tabular}%
}
\end{table}

\begin{table}[tb]
\caption{Average $E(A_m)$ values, denoted by $\mean{E}(A_m)$, grouped by model and optimizer.}
\label{tbl:e_avg_model_optimizer}
\resizebox{\textwidth}{!}{%
\begin{tabular}{@{}l|rr|r|rr|r|rr|r|rr|r|r@{}}
\toprule
Outcome               & \multicolumn{2}{c|}{CNN}    & CNN   & \multicolumn{2}{c|}{LSTM} & LSTM  & \multicolumn{2}{c|}{MLP} & MLP   & \multicolumn{2}{c|}{Transformer} & Transformer & Grand \\
\cmidrule(lr){2-3}\cmidrule(lr){5-6}\cmidrule(lr){8-9}\cmidrule(lr){11-12}
                      & Adam          & SGD        & Total & Adam        & SGD        & Total & Adam        & SGD       & Total & Adam           & SGD            & Total       & Total \\
\midrule
\textsc{l:l,*,*}  & 41   & 72  & 47        & 47   &     & 47         & 100  &     & 100       & 233         & 104 & 169               & 83    \\
\textsc{l:t,l,*}  & 13   &     & 13        & 31   & 70  & 49         & 31   & 806 & 612       & 29          & 15  & 18                & 125   \\ \midrule
Loss Total        & 31   & 72  & 37        & 33   & 70  & 48         & 90   & 806 & 305       & 165         & 44  & 85                & 106   \\ \midrule
\textsc{t:t,t,t}  &      &     &           & 0    &     & 0          &      &     &           &             &     &                   & 0     \\ \midrule
Tie Total         &      &     &           & 0    &     & 0          &      &     &           &             &     &                   & 0     \\ \midrule
\textsc{w:t,w,wt} &      &     &           & -8   & -47 & -36        &      &     &           & -18         &     & -18               & -32   \\
\textsc{w:w,l,*}  & 9    & 400 & 204       &      &     &            & 333  & 400 & 358       & 36          & 11  & 17                & 235   \\
\textsc{w:w,t,wt} &      & 0   & 0         &      &     &            &      &     &           &             &     &                   & 0     \\
\textsc{w:w,w,l}  &      & -63 & -63       &      &     &            & -27  & -59 & -46       &             & -3  & -3                & -43   \\
\textsc{w:w,w,wt} & -47  & -43 & -44       &      &     &            & -37  & -61 & -52       & -25         &     & -25               & -46   \\ \midrule
Win Total         & -28  & 7   & -1        & -8   & -47 & -36        & 107  & 21  & 58        & -14         & 7   & -7                & 19    \\ \midrule
Grand Total       & 14   & 14  & 14        & 22   & 11  & 17         & 101  & 139 & 120       & 39          & 30  & 34                & 53   \\
\bottomrule
\end{tabular}%
}
\end{table}

\clearpage
\newpage

\section{Details of Models' Performance}\label{sec:accuracy_figures}
Using box-and-whisker plots, we show the distribution of test accuracy for the \num{100} experiments. Each subplot in a figure shows the accuracy for a given PRNG- and QRNG-based initializer type. $x$-axis represents epoch number, $y$-axis represents model accuracy on test dataset. In order to make it easier for the reader (since there are 12 figures), we created a table of figures as shown in Table~\ref{tbl:fig_toc}.

\begin{figure}[ht]
    \centering
    \includegraphics[width=\textwidth]{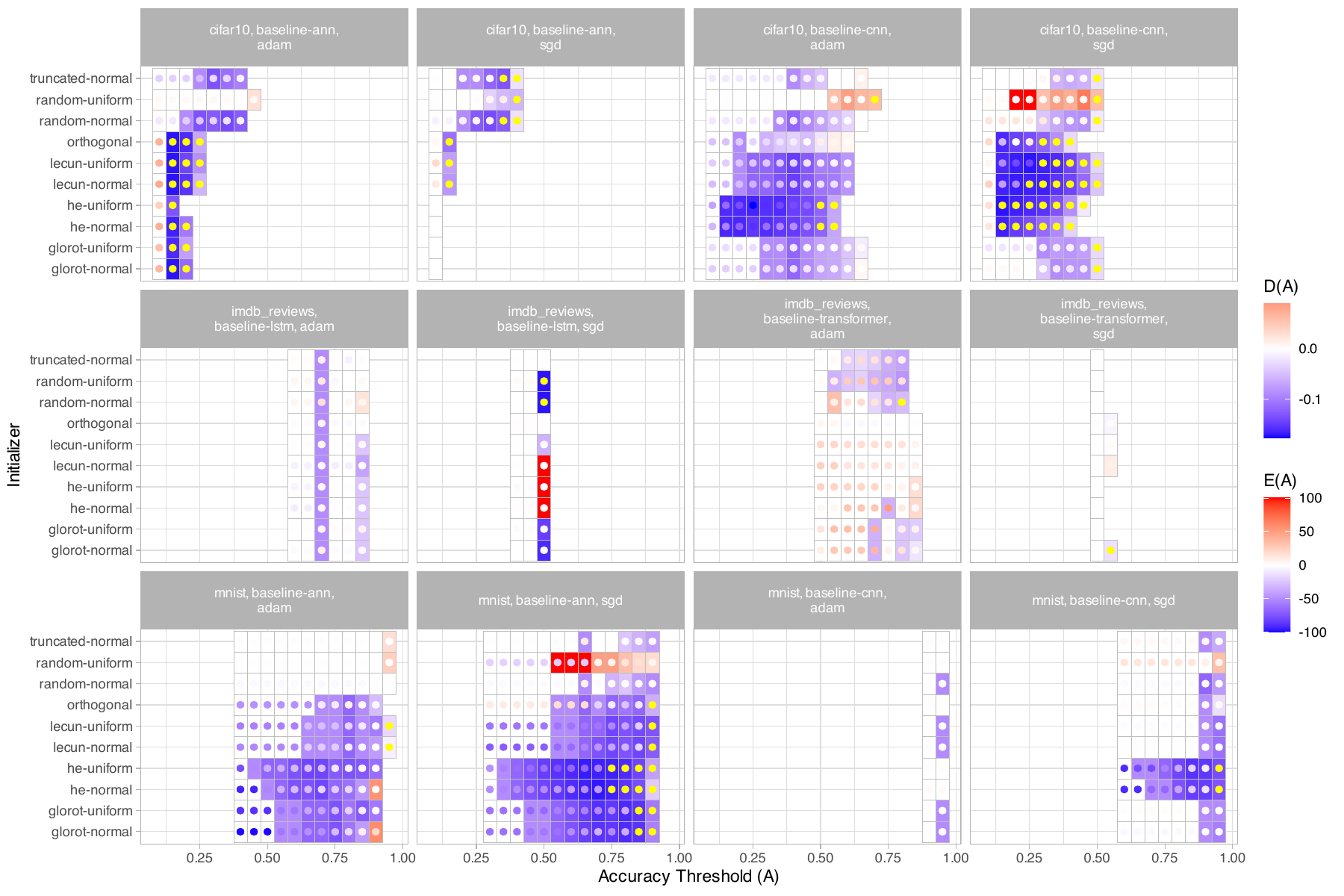}
    \caption{A summary plot showing the values of $E(A)$ in the main colour of a tile and $D(A)$ in the circle in the middle of a given tile. A yellow circle indicates that $D(A)$ is indeterminate (i.e., pseudorandom or quasirandom experiments did not achieve a given level of accuracy in 30 epochs). \\[5 pt]
    The values of the median accuracy thresholds $A$ shown are $0.10, 0.15, 0.20, \ldots, 0.95$. A $0.05$ increment is chosen for $A$ to ensure the plots are readable. \\[5 pt]
    Among all the tiles, 17 have $E(A) > 100\%$. To improve the readability of the figures' colours, we cap these cases at $100\%$ and list them in Table~\ref{tbl:e_outliers}. \\[5 pt] 
    A negative (blue) value of $E(A)$ indicates that the QRNG-based version of initializer achieves $A$ faster than the PRNG-based version. A negative (blue) value of $D(A)$ indicates that QRNG-based version of the initializers exhibit less variation in $A$ than PRNG-based version. The values of $A_m$ are given by the rightmost tile in each row of the heatmaps. }
    \label{fig:summary}
\end{figure}

Additionally, we create a heatmap summary plot, shown in Figure~\ref{fig:summary}, which summarizes the dynamics depicted in the box-and-whisker plots on one page. Table~\ref{tbl:e_outliers} lists 17 values of $E(A)$ from Figure~\ref{fig:summary} that are capped at $100\%$.

\begin{table}[H]
\caption{A table of figures showing the distributions of test accuracy based on \num{100} experiments.}
\label{tbl:fig_toc}
\centering
\begin{tabular}{@{}llrrr@{}}
\toprule
Model                        & Optimizer      & MNIST & CIFAR-10 & IMDB \\ 
\midrule
\multirow{2}{*}{MLP}         & SGD  & Fig.~\ref{fig:MNIST_SGD_MLP_all}      & Fig.~\ref{fig:CIFAR-10_SGD_MLP_all}         & --      \\
                             & Adam & Fig.~\ref{fig:MNIST_Adam_MLP_all}      & Fig.~\ref{fig:CIFAR-10_Adam_MLP_all}         & --      \\
\multirow{2}{*}{CNN}         & SGD  & Fig.~\ref{fig:MNIST_SGD_CNN_all}      & Fig.~\ref{fig:CIFAR-10_SGD_CNN_all}         & --      \\
                             & Adam & Fig.~\ref{fig:MNIST_Adam_CNN_all}      & Fig.~\ref{fig:CIFAR-10_Adam_CNN_all}         & --      \\
\multirow{2}{*}{LSTM}        & SGD  & --      & --         & Fig.~\ref{fig:IMDB_SGD_LSTM_all}      \\
                             & Adam & --      & --         & Fig.~\ref{fig:IMDB_Adam_LSTM_all}     \\
\multirow{2}{*}{Transformer} & SGD  & --      & --         & Fig.~\ref{fig:IMDB_SGD_Transformer_all}    \\
                             & Adam & --      & --         &  Fig.~\ref{fig:IMDB_Adam_Transformer_all}    \\ 
\bottomrule                             
\end{tabular}%

\end{table}

\begin{table}[H]
\caption{The values of $E(A) > 100\%$ (see Figure~\ref{fig:summary} for details).}\label{tbl:e_outliers}
\centering
\begin{tabular}{@{}llllr@{}}
\toprule
Dataset  & Architecture & Initializer    & Optimizer    & $E(A)$   \\ \midrule
CIFAR-10 & CNN         & He Normal      & SGD       & 400   \\
CIFAR-10 & CNN         & He Uniform     & SGD       & 400   \\
CIFAR-10 & MLP         & Glorot normal  & Adam      & 400   \\
CIFAR-10 & MLP         & Glorot normal  & SGD       & 400   \\
CIFAR-10 & MLP         & Glorot uniform & SGD       & 400   \\
CIFAR-10 & MLP         & He normal      & SGD       & 2000  \\
CIFAR-10 & MLP         & He uniform     & Adam      & 400   \\
CIFAR-10 & MLP         & He uniform     & SGD       & 400   \\
CIFAR-10 & MLP         & Lecun normal   & Adam      & 400   \\
CIFAR-10 & MLP         & Orthogonal     & Adam      & 400   \\
CIFAR-10 & MLP         & Orthogonal     & SGD       & 400   \\
IMDB     & LSTM        & Lecun normal   & SGD       & 209   \\
IMDB     & Transformer & He normal      & Adam      & 238   \\
IMDB     & Transformer & He uniform     & Adam      & 229   \\
IMDB     & Transformer & Random normal  & SGD       & 138   \\
MNIST    & MLP         & Glorot normal  & Adam      & 154   \\
MNIST    & MLP         & Random uniform & Adam      & 146   \\ \bottomrule
\end{tabular}
\end{table}

\begin{figure}[H]
    \centering
    \includegraphics[width=\textwidth]{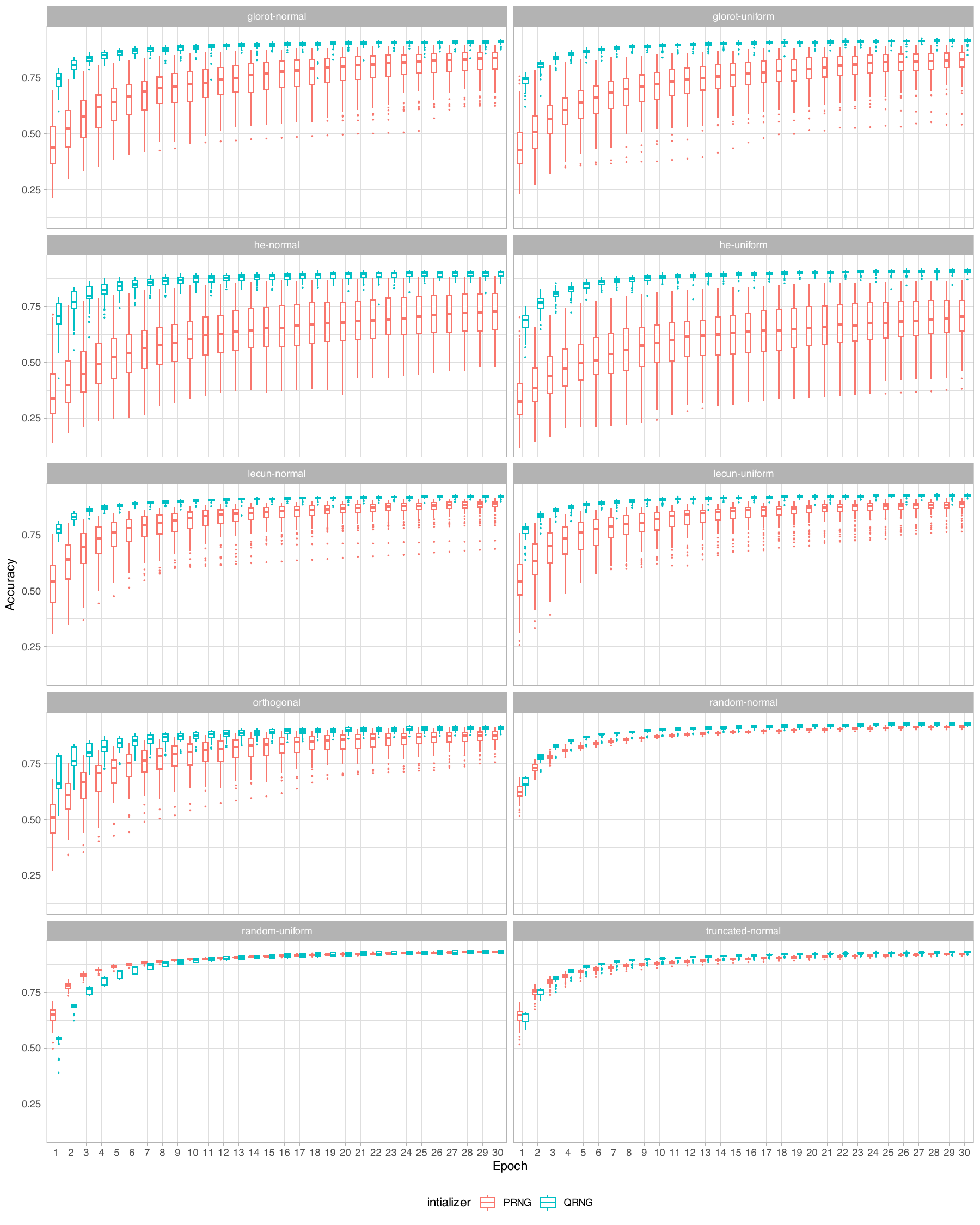}
    \caption{Accuracy distributions for MNIST test dataset, SGD optimizer and MLP model.}
    \label{fig:MNIST_SGD_MLP_all}
\end{figure}

\begin{figure}[H]
    \centering
    \includegraphics[width=\textwidth]{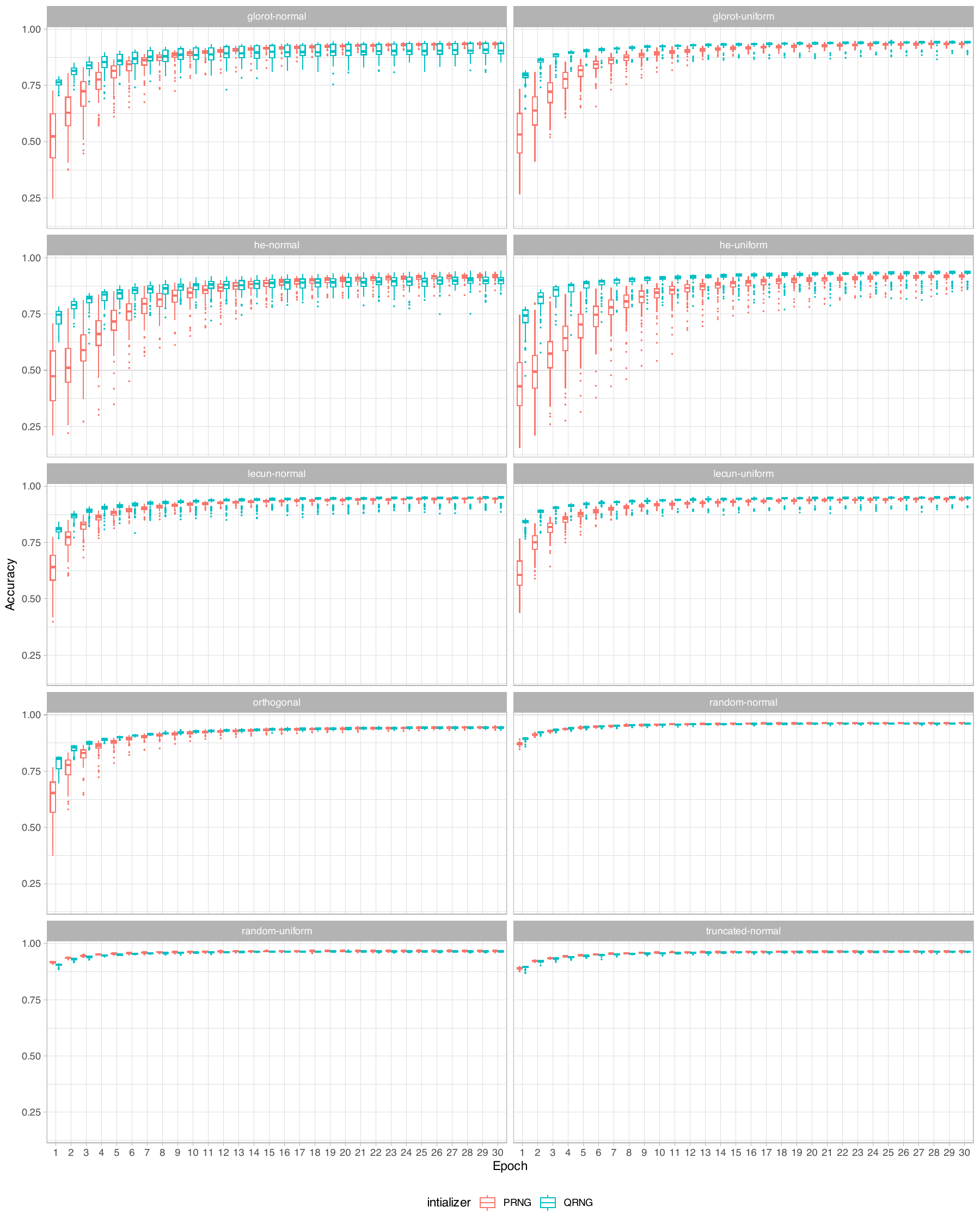}
    \caption{Accuracy distributions for MNIST test dataset, Adam optimizer and MLP model.}
    \label{fig:MNIST_Adam_MLP_all}
\end{figure}

\begin{figure}[H]
    \centering
    \includegraphics[width=\textwidth]{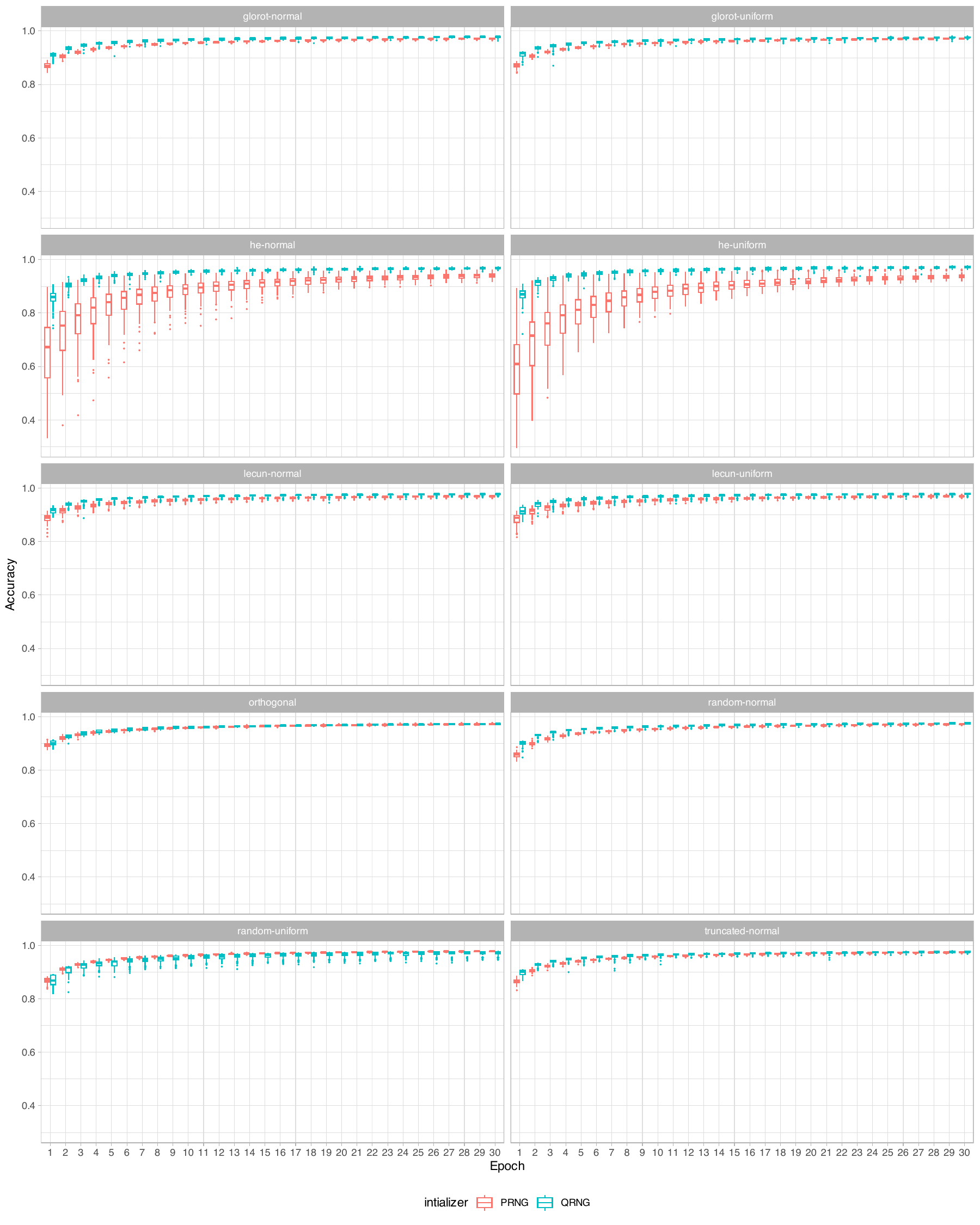}
    \caption{Accuracy distributions for MNIST test dataset, SGD optimizer and CNN model.}
    \label{fig:MNIST_SGD_CNN_all}
\end{figure}

\begin{figure}[H]
    \centering
    \includegraphics[width=\textwidth]{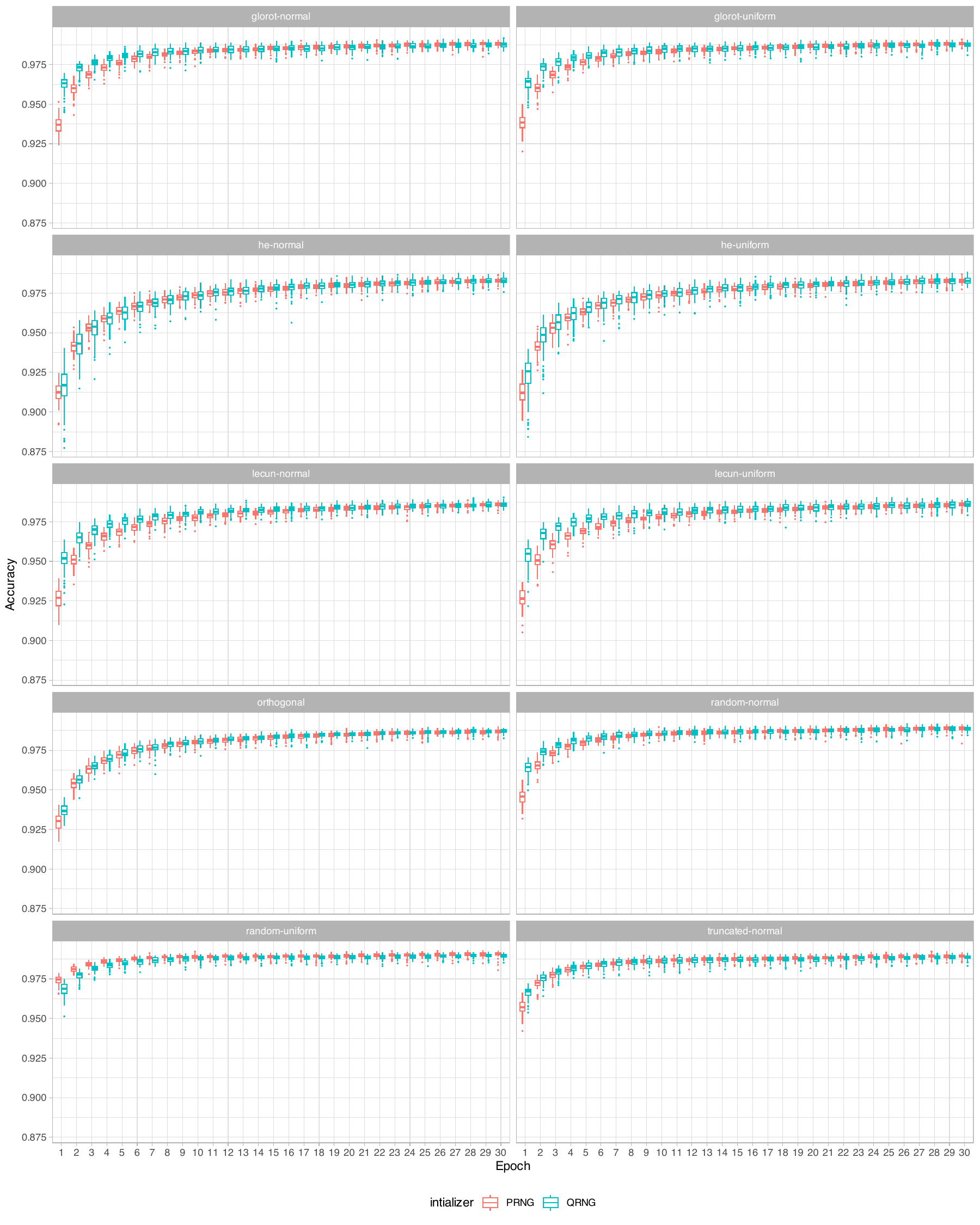}
    \caption{Accuracy distributions for MNIST test dataset, Adam optimizer and CNN model.}
    \label{fig:MNIST_Adam_CNN_all}
\end{figure}

\begin{figure}[H]
    \centering
    \includegraphics[width=\textwidth]{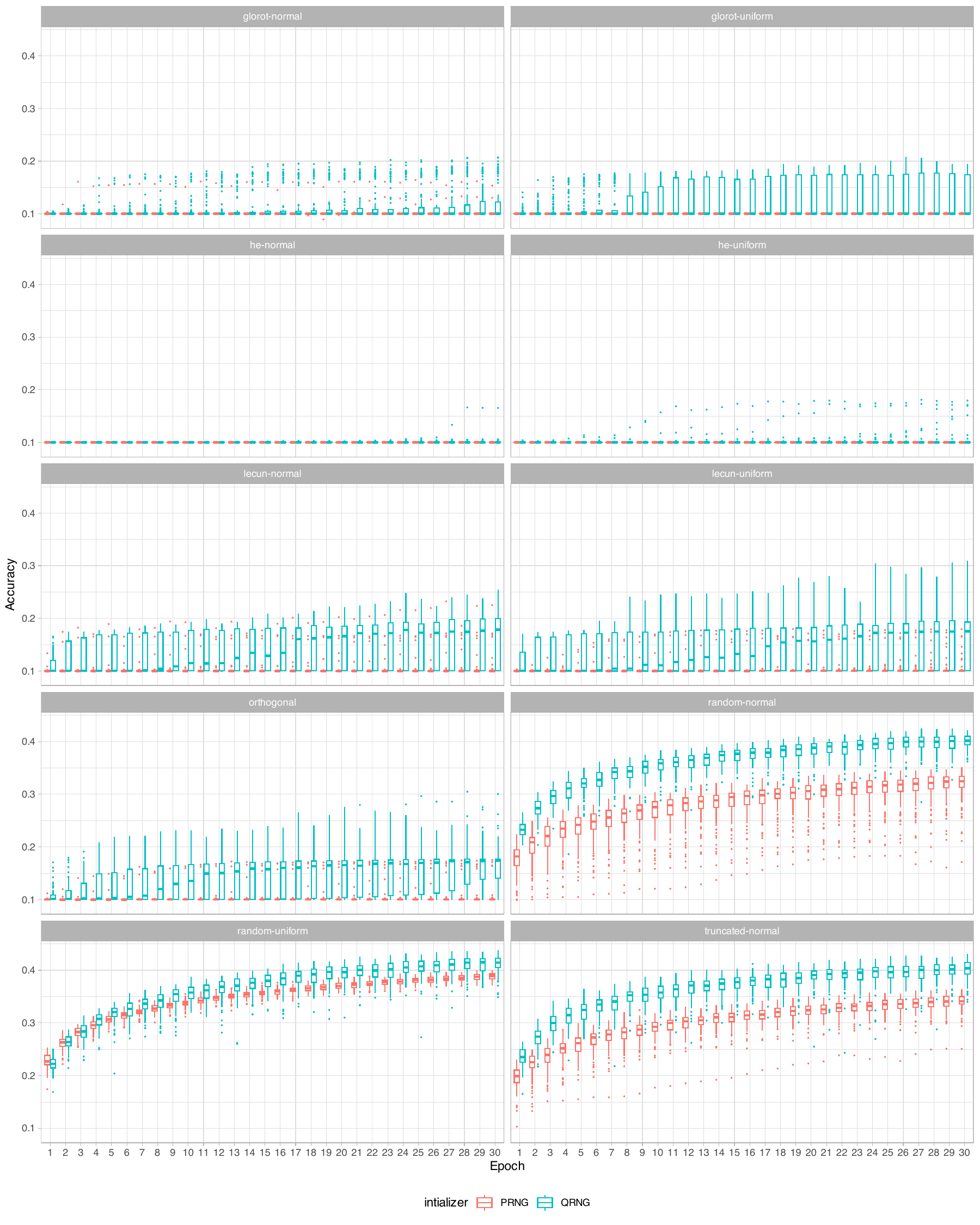}
    \caption{Accuracy distributions for CIFAR-10 test dataset, SGD optimizer and MLP model.}
    \label{fig:CIFAR-10_SGD_MLP_all}
\end{figure}

\begin{figure}[H]
    \centering
    \includegraphics[width=\textwidth]{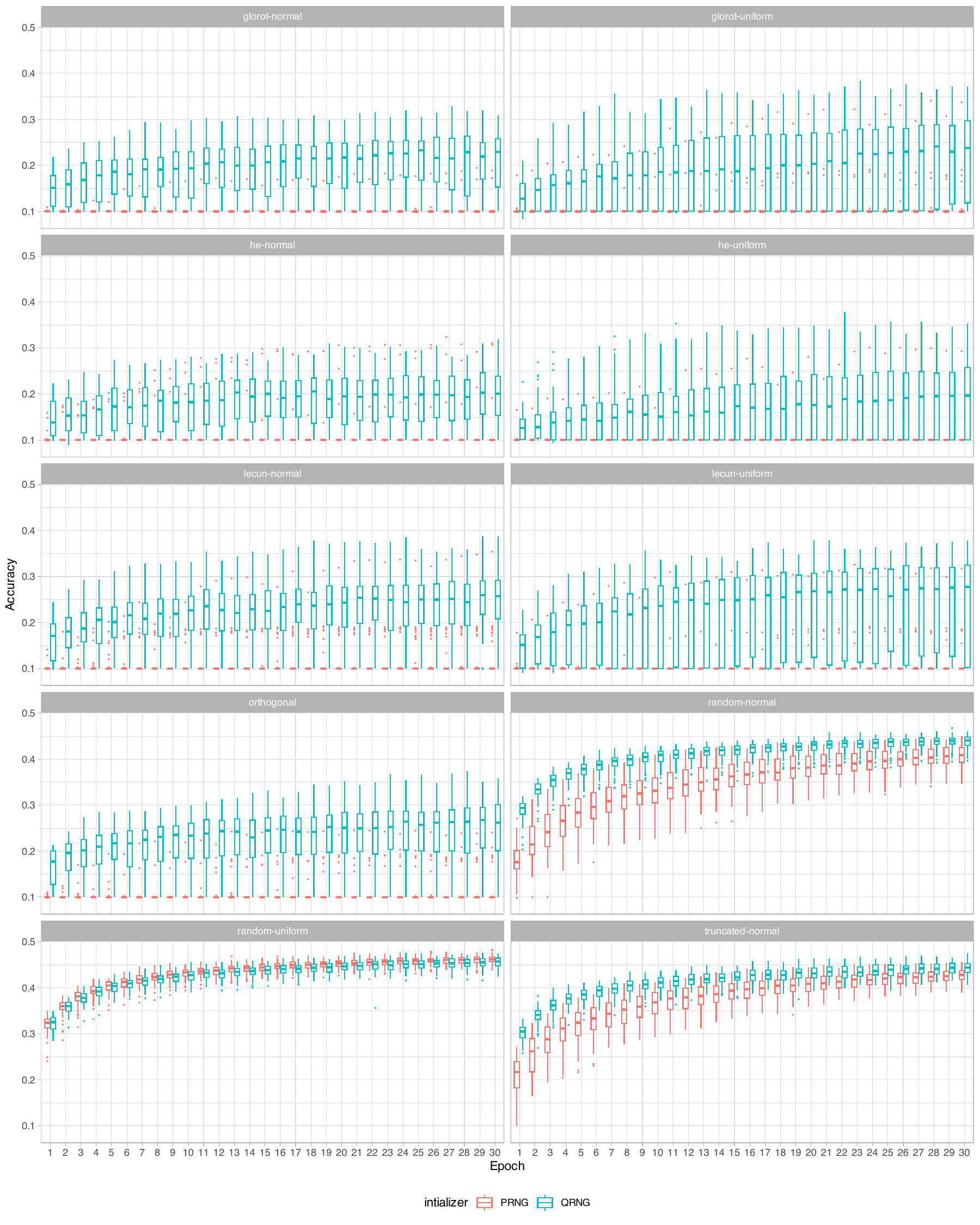}
    \caption{Accuracy distributions for CIFAR-10 test dataset, Adam optimizer and MLP model.}
    \label{fig:CIFAR-10_Adam_MLP_all}
\end{figure}

\begin{figure}[H]
    \centering
    \includegraphics[width=\textwidth]{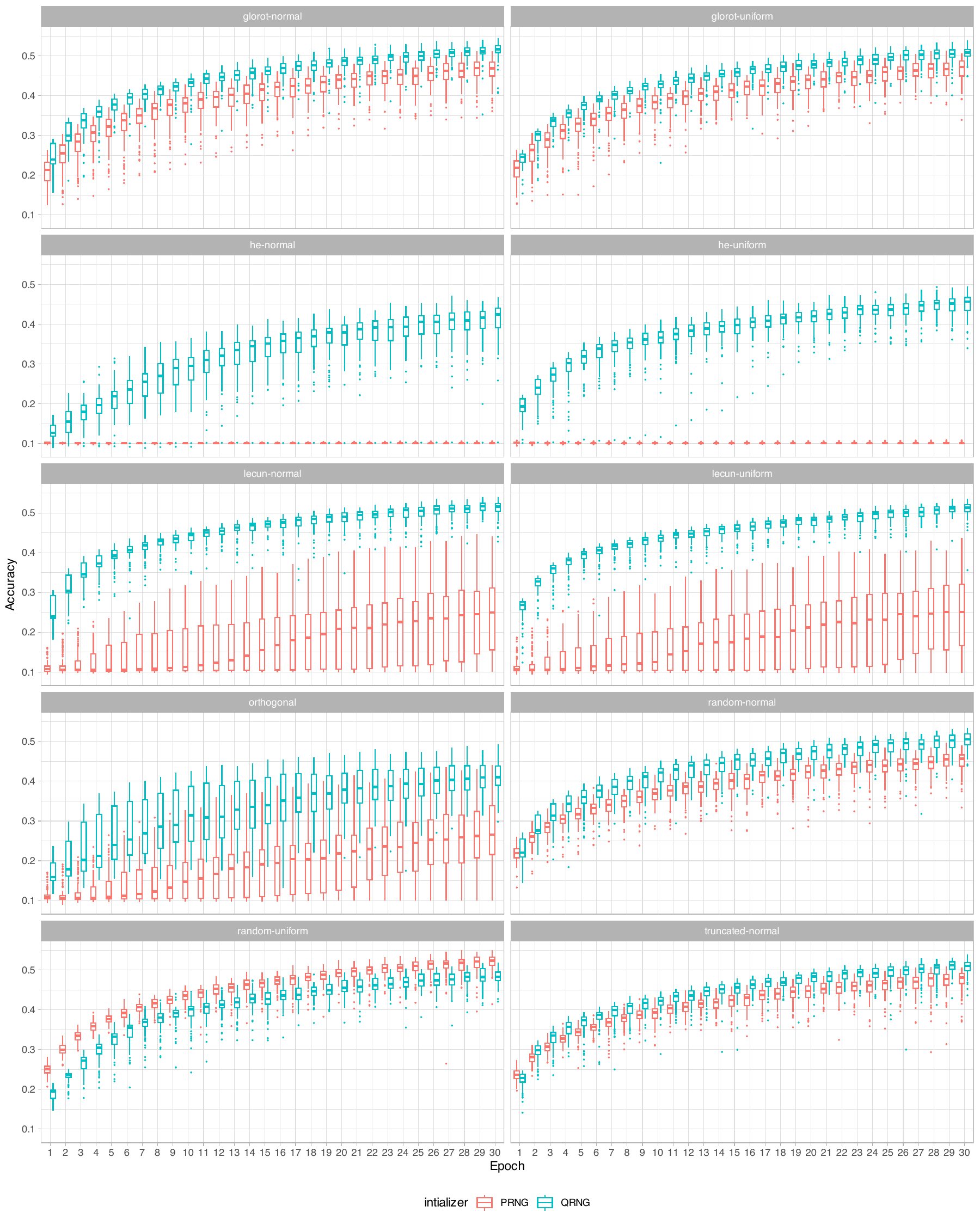}
    \caption{Accuracy distributions for CIFAR-10 test dataset, SGD optimizer and CNN model.}
    \label{fig:CIFAR-10_SGD_CNN_all}
\end{figure}

\begin{figure}[H]
    \centering
    \includegraphics[width=\textwidth]{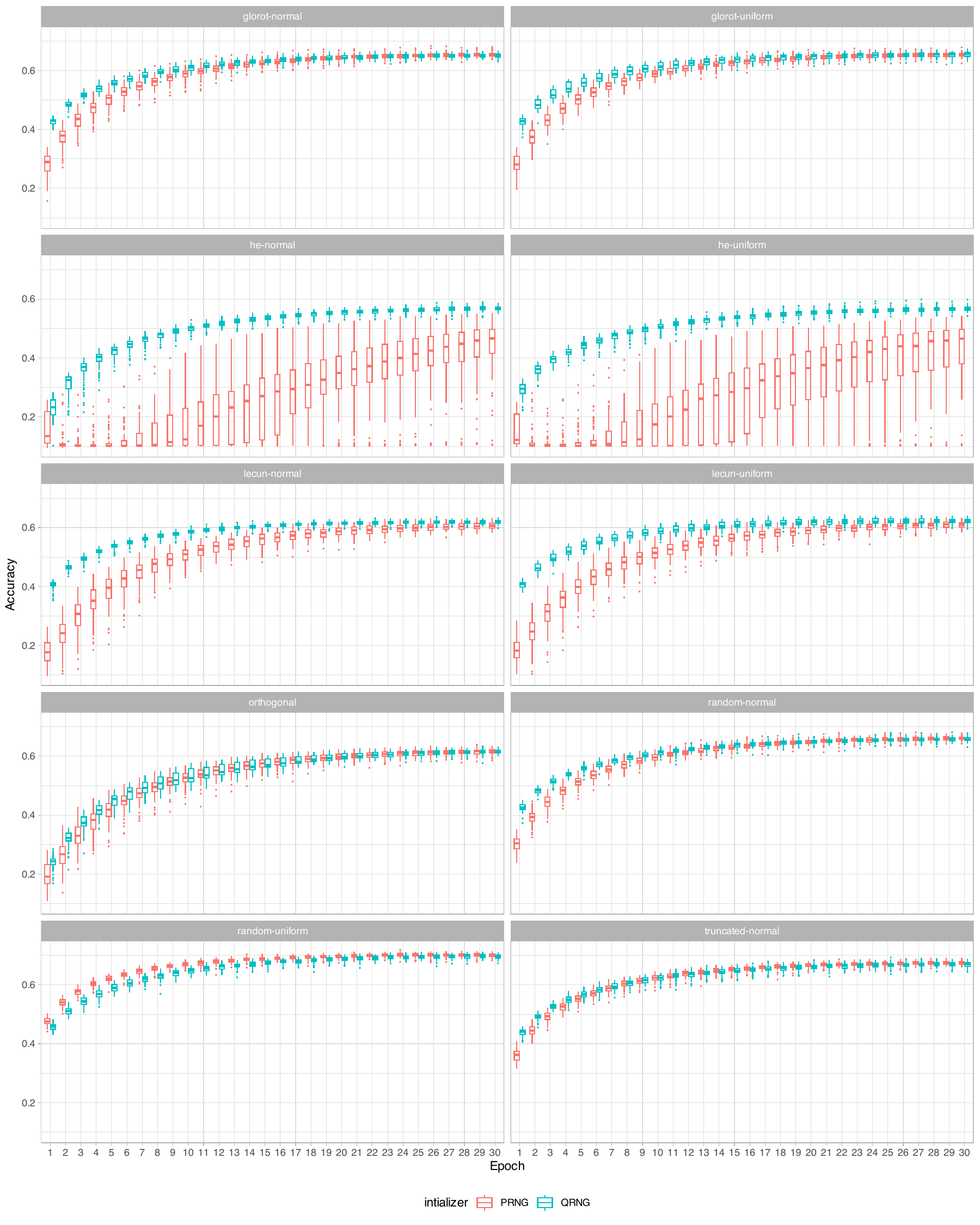}
    \caption{Accuracy distributions for CIFAR-10 test dataset, Adam optimizer and CNN model.}
    \label{fig:CIFAR-10_Adam_CNN_all}
\end{figure}

\begin{figure}[H]
    \centering
    \includegraphics[width=\textwidth]{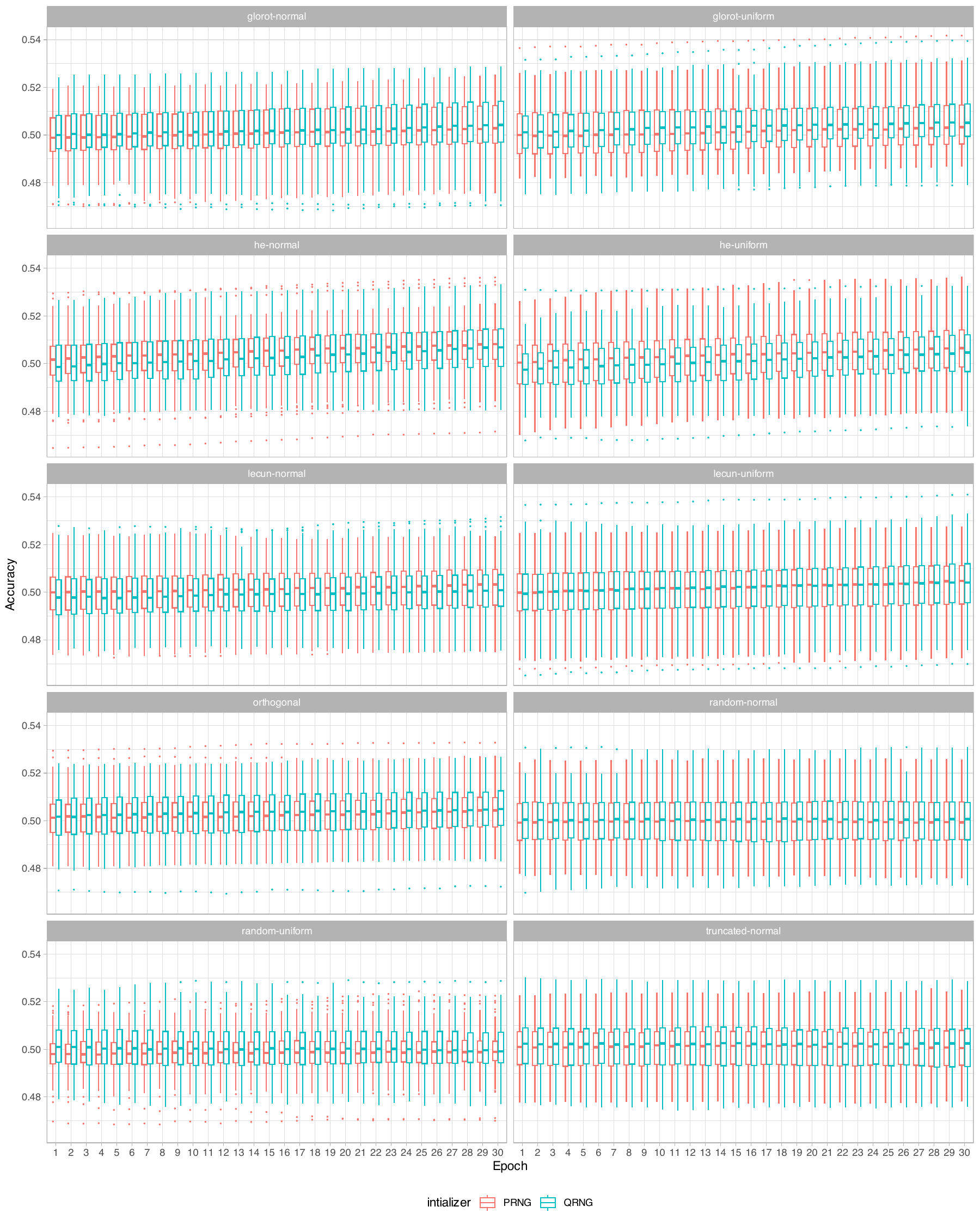}
    \caption{Accuracy distributions for IMDB test dataset, SGD optimizer and LSTM model.}
    \label{fig:IMDB_SGD_LSTM_all}
\end{figure}

\begin{figure}[H]
    \centering
    \includegraphics[width=\textwidth]{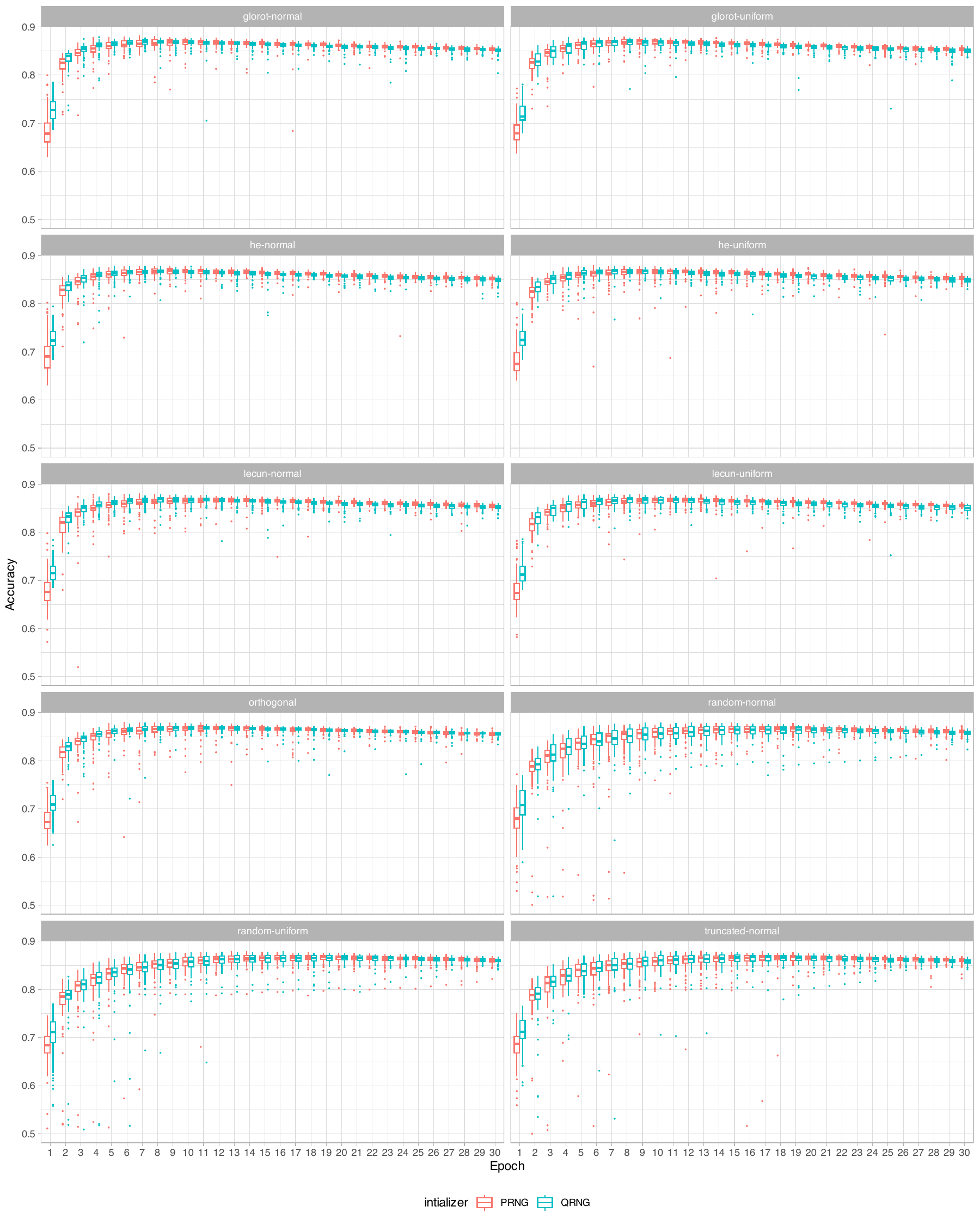}
    \caption{Accuracy distributions for IMDB test dataset, Adam optimizer and LSTM model.}
    \label{fig:IMDB_Adam_LSTM_all}
\end{figure}

\begin{figure}[H]
    \centering
    \includegraphics[width=\textwidth]{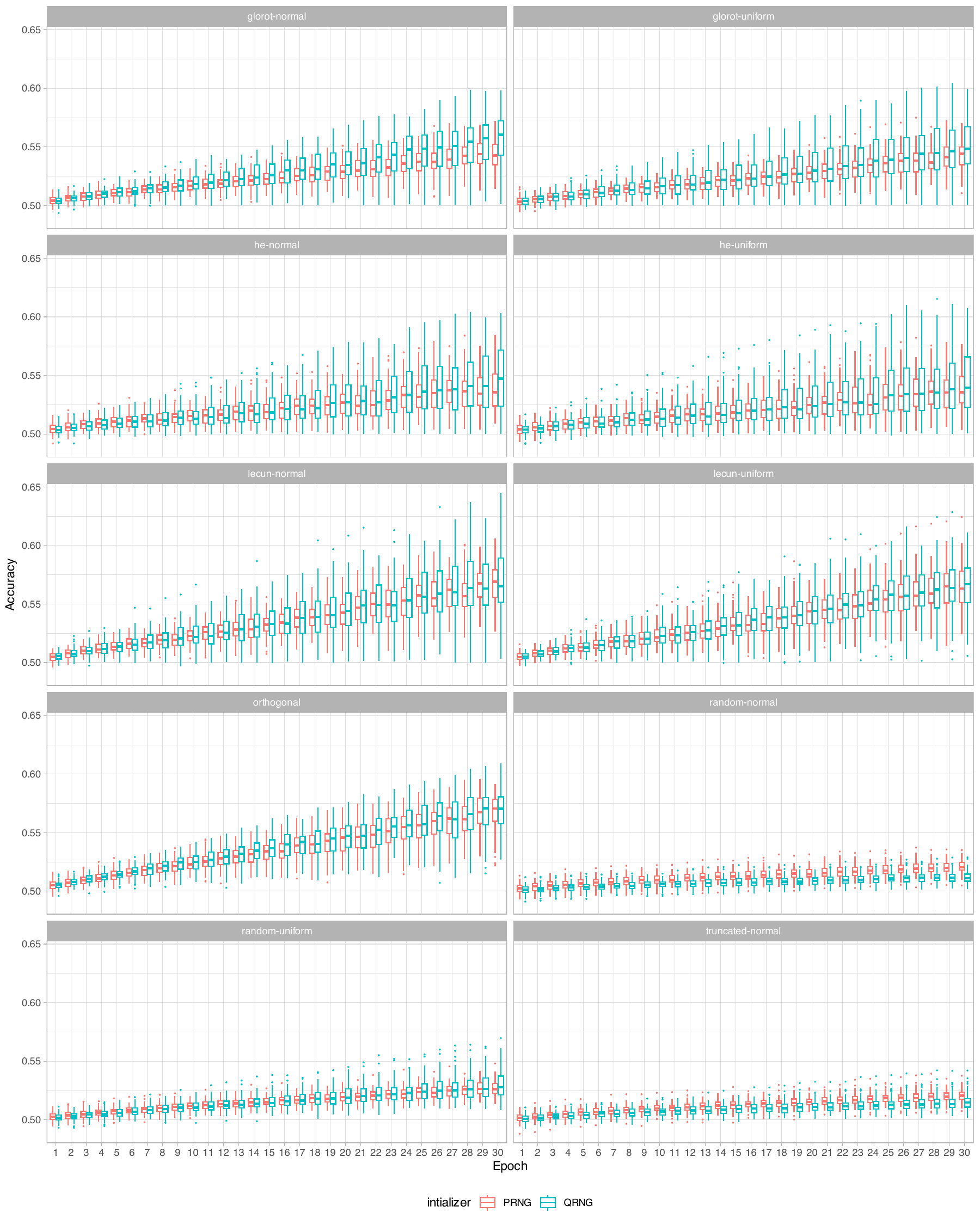}
    \caption{Accuracy distributions for IMDB test dataset, SGD optimizer and Transformer model.}
    \label{fig:IMDB_SGD_Transformer_all}
\end{figure}

\begin{figure}[H]
    \centering
    \includegraphics[width=\textwidth]{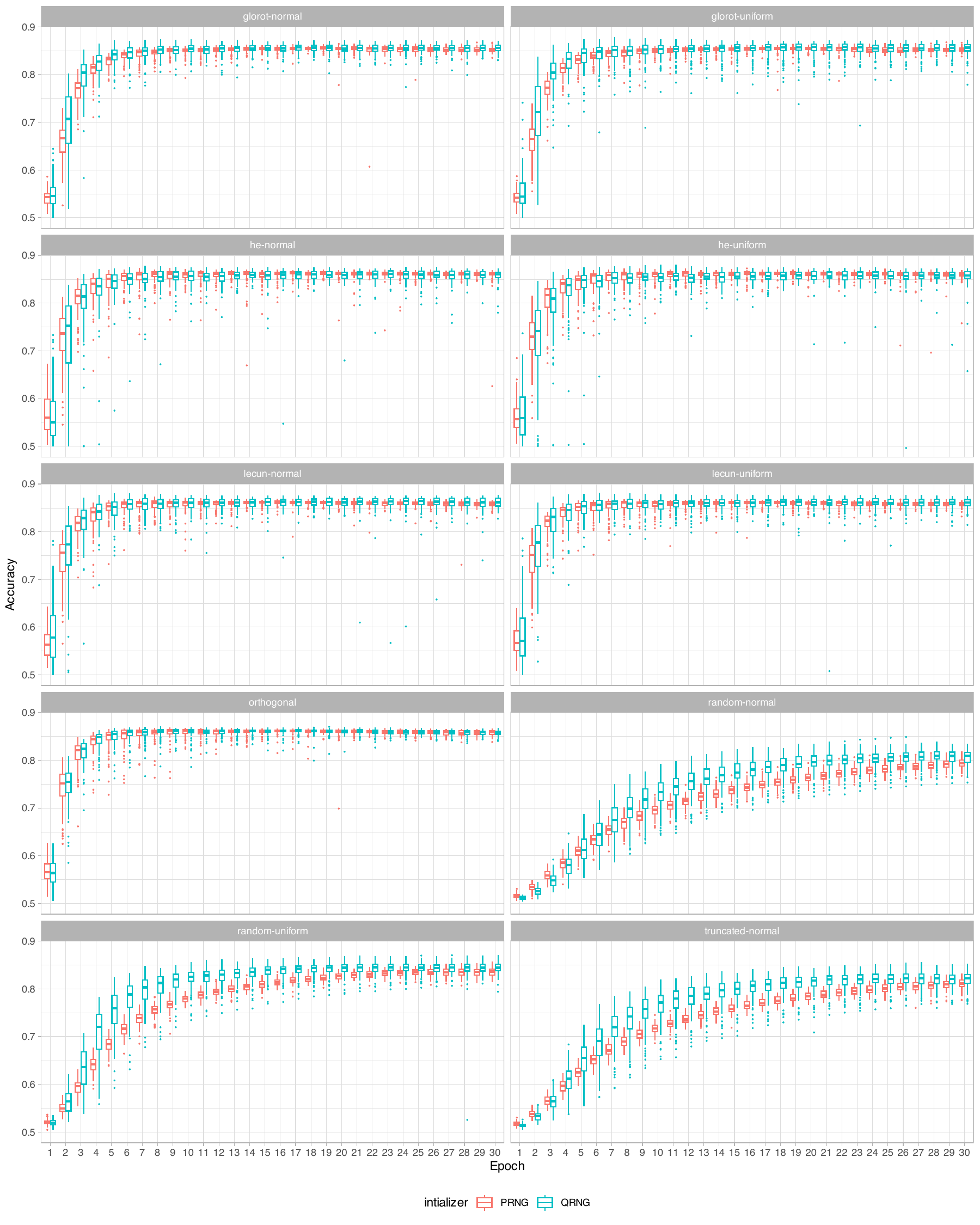}
    \caption{Accuracy distributions for IMDB test dataset, Adam optimizer and Transformer model.}
    \label{fig:IMDB_Adam_Transformer_all}
\end{figure}

\end{document}